\documentclass[10pt,twocolumn,letterpaper]{article}

\usepackage[pagenumbers,datasets]{wacv}      

\usepackage{placeins}

\definecolor{wacvblue}{rgb}{0.21,0.49,0.74}
\usepackage{xr-hyper}
\usepackage[pagebackref,breaklinks,colorlinks,allcolors=wacvblue]{hyperref}
\usepackage{adjustbox}

\usepackage{adjustbox}
\def\wacvPaperID{509} 
\def\confName{WACV}
\def\confYear{2027}

\title{What Makes a Good Medical Image Tokenizer? \\Rethinking Reconstruction and Generation in Medical Image Tokenization}

\author{
Niklas Bubeck$^{1,2,3}$ \quad
Yundi Zhang$^{1,3}$ \quad
Vasiliki Sideri-Lampretsa$^{1,3}$ \quad
Julian McGinnis$^{1,3}$ \\
Jiancheng Yang$^{4,5}$ \quad
Daniel Rueckert$^{1,2,3,6}$ \quad
Jiazhen Pan$^{1,2,3}$ \\[0.6em]
{\small $^{1}$School of Computation, Information and Technology, Technical University of Munich, Germany} \\
{\small $^{2}$Munich Center for Machine Learning, Technical University of Munich, Germany} \\
{\small $^{3}$School of Medicine, Klinikum rechts der Isar, Technical University of Munich, Germany} \\
{\small $^{4}$ELLIS Institute Finland, Finland \quad $^{5}$School of Electrical Engineering, Aalto University, Finland} \\
{\small $^{6}$Department of Computing, Imperial College London, UK} \\[0.3em]
{\tt\small niklas.bubeck@tum.de}
}

\newcommand\blfootnote[1]{%
  \begingroup
  \renewcommand\thefootnote{}\footnote{#1}%
  \addtocounter{footnote}{-1}%
  \endgroup
}

\begin{document}
\maketitle

\begin{abstract}
Latent diffusion models now dominate medical image generation, and every such pipeline rests on a \emph{tokenizer} that compresses images into the latent codes for image generation to operate on. Thereby, the tokenizer choice bounds every downstream task from reconstruction fidelity and generation quality to the representations available for downstream analysis. Yet, medical imaging pipelines routinely utilize tokenizers from natural imaging on the hypothesis that their behavior carries over. However, this is an assumption never tested in the medical imaging regime, where datasets are orders of magnitude smaller and images exhibit far lower inter-sample variance. We present a systematic evaluation of medical image tokenizers evaluating thirty configurations across ten model families on twelve datasets at three compression factors, spanning reconstruction, generation, latent geometry, downstream classification, and memorization. We find that (1)~performance on image reconstruction and generation strongly correlate, unlike prior reports on natural images; (2)~modern tokenizers use nearly all of their codebook entries, but still leave most of the latent space unused; (3)~training-set memorization is mild and is further suppressed by stronger latent space compression; and (4)~discrete quantization can largely preserve downstream classification, with lookup-free schemes being the main exception.\blfootnote{To foster research in this direction, we will open-source the framework and all models upon acceptance.}
\end{abstract}

\section{Introduction}
\label{sec:intro}

\begin{figure}[t]
    \centering
    \includegraphics[width=\linewidth]{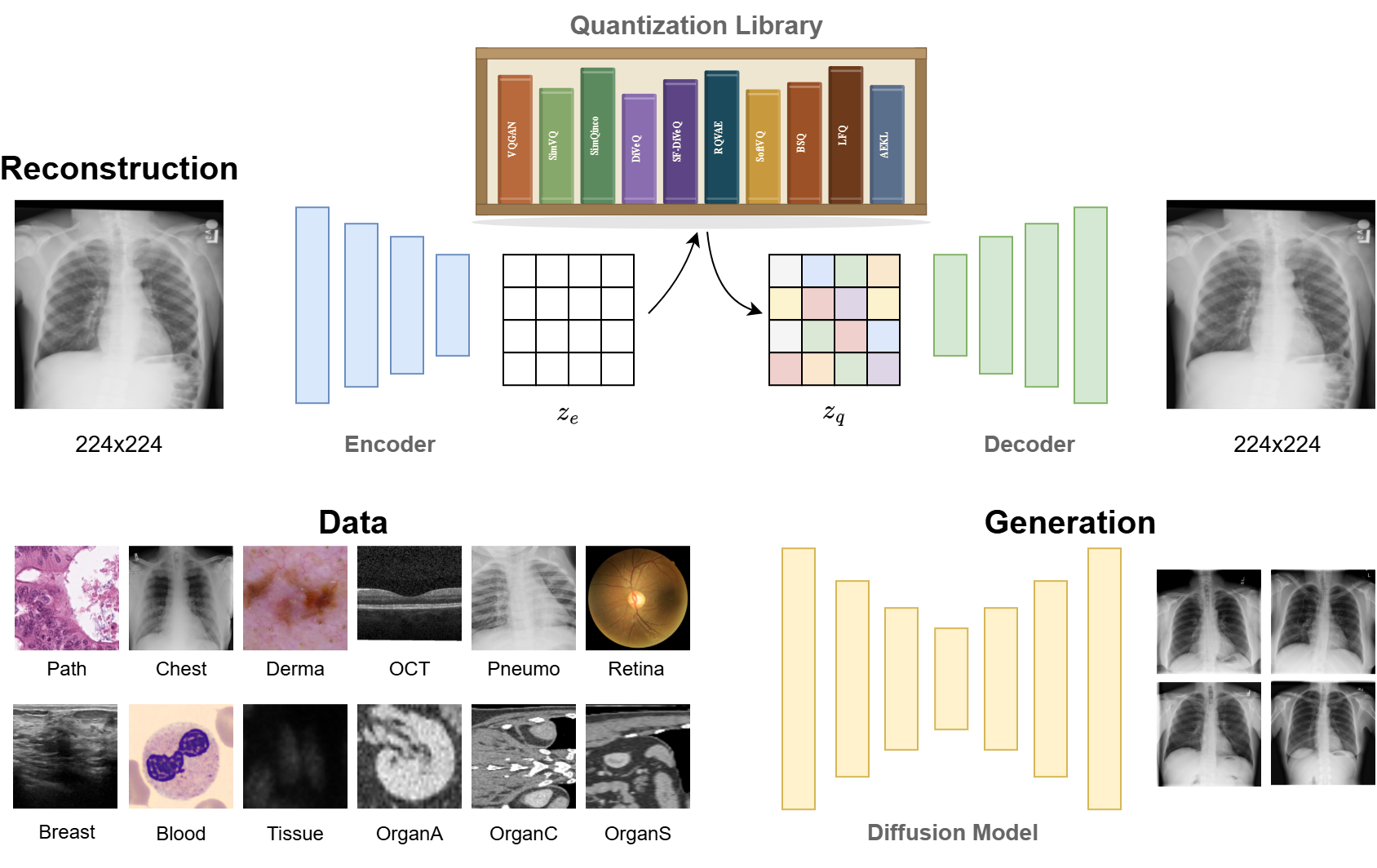}
    \caption{\textbf{Method Overview.} We take quantizers out of the library and evaluate them on diverse medical imaging datasets to study whether prevalent assumptions from natural imaging transfer to the medical domain.}
    \label{fig:teaser_overview}
\end{figure}

Medical imaging faces persistent challenges: limited data hinders generalization~\cite{data-scarcity-medical}, expert annotations are costly~\cite{gen-models-medical}, and strict privacy regulations constrain data sharing~\cite{privacy-medical-ai,medicaldiffusion-3d}. Generative AI offers a compelling path forward, enabling synthetic augmentation, report-to-image synthesis, and latent compression for efficient analysis. Generative models have become dominant across a wide range of medical imaging applications~\cite{kazerouni2023survey,wang2025genai}, including MRI reconstruction~\cite{diffuserecon,calid}, 3D CT and MRI generation~\cite{medicaldiffusion-3d,maisi,tumorflow}, chest X-ray synthesis from radiology reports~\cite{chexray-diffusion,roentgen}, controllable lesion editing~\cite{mam-e}, pathology and dermatology synthesis~\cite{pathldm,pathogen,lesiongenderm,derm-t2im}, retinal imaging~\cite{retinalogos,fundus2oct,retidiff}, and cell morphology modeling~\cite{cytodiffusion,cytodiff}. Across all of these fields, generation is built almost exclusively on diffusion over a \emph{continuous} latent, while autoregressive and discrete-token models remain comparatively rare and task-specific~\cite{medtok,meditok,medvar,spectrum}. That latent is produced by a \textbf{\emph{tokenizer}}, which compresses high-dimensional images into the compact space on which the generator operates. Its choice bounds everything from downstream task performance to generation quality, yet most medical pipelines adopt off-the-shelf quantizers without systematic justification, transferring conclusions from natural imaging. 

In this paper, we study tokenizers as substrates for latent diffusion, which underpins most medical generative pipelines, and challenge the prevailing approach, which carries over assumptions from natural images, and question this transfer, which remains understudied. The first is that reconstruction quality is a poor proxy for generation quality, with recent work instead attributing generation performance to global semantic information and spatial structure in the latent space~\cite{recon-vs-gen,xu2026makingrfid,rfid-gfid,irepa2025}. The second is that a \textit{healthy} codebook, having high utilization and no collapse, signals an expressive latent space~\cite{yu2022vitvqgan,fsq,zhu2024vqganlc,simvq} leading to improved generation performance. Both were established on ImageNet-scale natural images, but medical data occupy a fundamentally different statistical regime: collections are small and highly repetitive, images of the same organ share a global layout and differ only in subtle, localized detail, and the diagnostically meaningful signal often lives in fine-grained, high-frequency texture~\cite{favae}, such as a faint lesion, that a tokenizer must preserve exactly. With \textbf{less data to learn from, lower variation between samples, and a narrower margin for error}, conclusions drawn from natural images may not transfer cleanly to medical imaging. In consequence, we challenge the current practice and call for a careful re-examination before they are relied upon in this domain.
 
Medical imaging raises two additional concerns that natural image tokenizers were never designed to address. First, synthetic medical data is valuable mainly because it \textbf{protects patient privacy}, but that benefit disappears if the model memorizes and reproduces real patients~\cite{privacy-medical-ai, posada2026identity}. Whether the tokenizer's compression bottleneck affects this leakage has not yet been studied. Moreover, the tokens are rarely an end in itself: the same compressed representation is routinely reused for downstream clinical tasks such as classification, detection, and retrieval~\cite{ldae,histovae_retrieval}, and increasingly serves as the input to multimodal and medical foundation models~\cite{llmcxr,medtok,meditok,btb3d}, so whether quantization erases class-discriminative structure becomes a central criterion for selecting a tokenizer. Thus, compared to natural image tokenizers, the medical domain prioritizes two distinct goals over perfect pixel reconstruction: the latent representation must preserve the critical diagnostic features used by clinicians, and the subsequent generator must not compromise privacy by memorizing and reproducing exact patient scans. In this work we aim to close this gap. Our contributions are:

\begin{itemize}
\item \textbf{The first medical image tokenization benchmark}: a controlled holistic evaluation of over sixty trained models.
\item \textbf{A multi-axis evaluation protocol} (reconstruction, generation, latent geometry, memorization, and downstream probing) revealing that natural image assumptions about reconstruction--generation correlation and latent space usage do not hold within our medical benchmark.
\item \textbf{Reproducibility}: open-source framework, all implementations, and trained tokenizer and generator weights, publicly released.
\end{itemize}


\begin{figure*}[t]
    \centering
    \includegraphics[width=\linewidth]{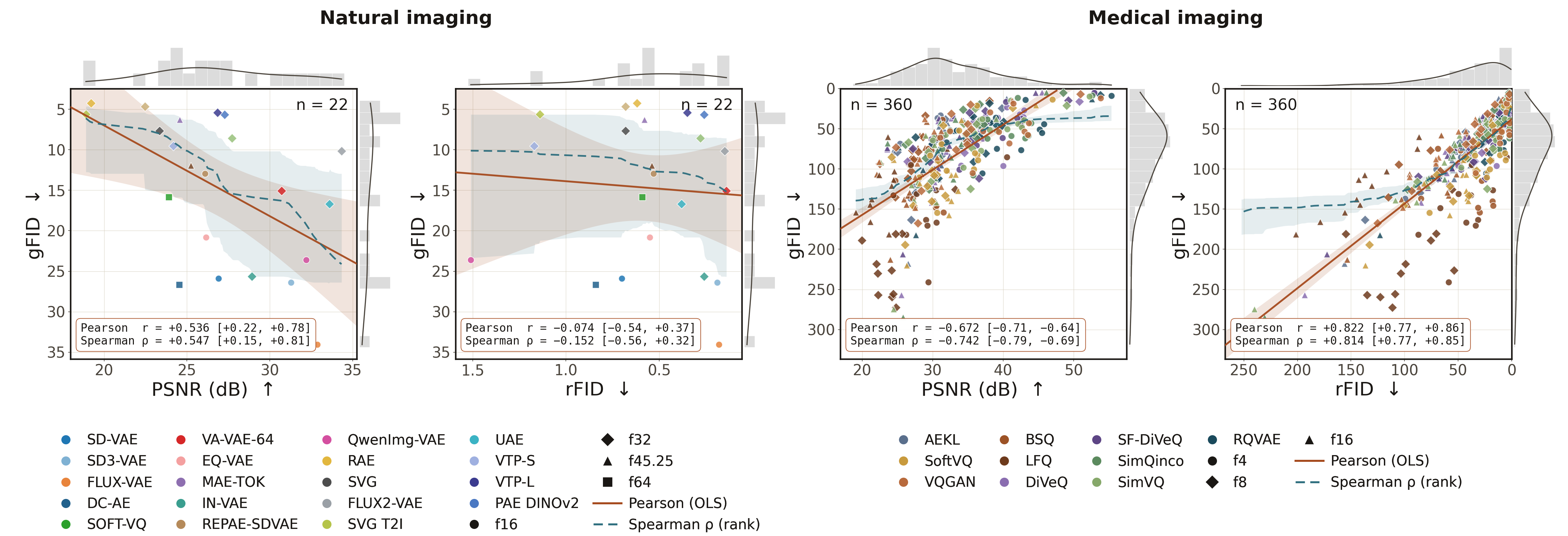}
    \caption{\textbf{\textit{Reconstruction predicts generation performance in medical imaging.}} Reconstruction versus generation performance of different tokenizers across compression factors and datasets, for natural images (left) and medical images (right); natural-image metrics are taken from prior work~\cite{xu2026makingrfid}, exploiting different methods and backbones. Pearson and Spearman fits are overlaid. Results disentangled by dataset and compression factor are provided in the appendix (\cref{sec:supp-correlation}), where the correlation holds within each stratum. Within our benchmark, reconstruction strongly predicts generation, in contrast to the natural-image trend.}
    \label{fig:correlations}
\end{figure*}

\section{Background and Related Work}
\label{sec:related}

The variational autoencoder (VAE)~\cite{vae} is the foundational framework for learned image compression: an encoder parameterizes a posterior $q_\phi(z|x)$, a prior $p(z)$ regularizes the latent space, and a decoder $p_\theta(x|z)$
reconstructs the input. Expressive decoders, however, can bypass the latent entirely (\emph{posterior collapse}) by driving the Kullback-Leibler (KL) term to zero. VQ-VAE~\cite{vqvae} avoids this by discretizing the bottleneck: an encoder $E$ maps $x$ to a continuous latent $z_e$, quantized to $z_q$ via nearest-neighbor lookup in a learned codebook $\mathcal{C} = \{e_k\}_{k=1}^{K}$, and a decoder $D$ reconstructs from the result:
\begin{equation}\label{eq:1}
\begin{aligned}
  z_e &= E(x), \quad
  z_q = e_{k^*}, \\
  k^* &= \arg\min_k \|z_e - e_k\|_2, \quad
  \hat{x} = D(z_q).
\end{aligned}
\end{equation}
The KL term is replaced by a commitment loss $\|z_e - \operatorname{sg}[z_q]\|_2^2$ and a codebook loss $\|\operatorname{sg}[z_e] - z_q\|_2^2$, with $\operatorname{sg}[\cdot]$ the stop-gradient; as $\arg\min$ is non-differentiable, gradients reach the encoder via the straight-through estimator (STE). VQ-GAN~\cite{vqgan} adds adversarial and perceptual losses for sharper reconstructions. Vanilla VQ nonetheless suffers \emph{codebook collapse}, where only a fraction of codes are used, as well as biased STE gradients. Subsequent work addresses these along several axes.

\noindent\textbf{\textit{Soft Quantization.}}
Replacing the hard $\arg\min$ with a differentiable weighted combination:
\begin{equation}
  z_q = \sum_{k=1}^{K} w_k\, e_k,
  \quad w_k \geq 0,\;\; \sum_{k} w_k = 1,
\end{equation}
bypasses the STE entirely. Methods differ in how $w_k$ is set: Wu and Flierl~\cite{wu2020quantization} use Bayesian posterior-mean estimation over centroids, SCQ~\cite{scq} solves for the optimal combination via a differentiable
optimization layer, and SoftVQ-VAE~\cite{softvq} uses a temperature-scaled softmax over codebook similarities, $w_k \propto \exp(-\|z_e - e_k\|^2 / \tau)$, recovering hard VQ as $\tau \to 0$. This generalizes VQ-VAE's $K$-means latent to soft $K$-means and, with data-dependent weights, to Gaussian mixtures~\cite{dilokthanakul2016gmvae}, rendering codebook and commitment losses unnecessary. \emph{Space-filling (SF) methods}~\cite{sfvq,diveq} instead quantize onto line segments between adjacent codes, $z_q = (1-\gamma)\,e_j + \gamma\,e_{j+1}$ with $\gamma \in [0,1)$, so neighboring codes encode similar content.

\noindent\textbf{\textit{Differentiable Quantization.}}
An orthogonal line keeps the hard $\arg\min$ in the forward pass but reparameterizes the output for richer encoder gradients. The STE, $z_q = z_e + \operatorname{sg}[e_{k^*} - z_e]$, copies the decoder gradient unchanged, discarding the quantization step. NSVQ~\cite{nsvq} instead substitutes magnitude-matched random noise for the quantization error. The rotation trick~\cite{rotationtrick} builds a rotation-rescaling matrix $\mathbf{M}$ with $\mathbf{M}\,z_e = e_{k^*}$ and sets $z_q = \operatorname{sg}[\mathbf{M}]\,z_e$, so the gradient $\partial z_q / \partial z_e = \operatorname{sg}[\mathbf{M}]$ encodes both angular and magnitude misalignment. DiVeQ~\cite{diveq} adds a simulated error vector via reparameterization, $z_q = z_e + \|z_e - e_{k^*}\|_2 \cdot u / \|u\|_2$, with direction $u$ aimed at the nearest codeword, keeping the forward pass identical to hard VQ.

\noindent\textbf{\textit{Lookup-Free Quantization.}}
Lookup-free methods drop the explicit codebook, quantizing each latent dimension
independently via per-dimension rounding $z_q = [q_1(z_1), \dots, q_d(z_d)]$ and eliminating collapse by construction. Finite scalar quantization (FSQ)~\cite{fsq} bounds each dimension via a $\tanh$ nonlinearity and rounds to one of $L_i$ uniformly spaced integer levels, $\hat{z}_i = \lfloor \tfrac{L_i-1}{2}\,\tanh(z_i) - o_i \rceil$, with $o_i = \tfrac{1}{2}$ for even $L_i$ and $0$ otherwise, yielding an implicit codebook of size $\prod_{i=1}^{d} L_i$ with no auxiliary losses. LFQ~\cite{lfq} takes $L_i = 2$, binarizing via $z_{q,i} = \operatorname{sign}(z_i) \in \{-1,+1\}$ for $2^d$ codes plus an entropy loss. Binary spherical quantization (BSQ)~\cite{bsq} refines LFQ by $\ell_2$-normalizing the latent onto the unit hypersphere before binarization, which bounds the quantization error and yields a better-conditioned entropy objective.

\noindent\textbf{\textit{Codebook Reparameterization.}}
These methods define codes through learned transformations so all entries receive gradients. SimVQ~\cite{simvq} factorizes the codebook as $\mathcal{C} = \mathbf{W}\mathbf{B}$ via a learnable linear layer over a latent basis. QINCo~\cite{qinco} replaces it with an implicit neural codebook: at each residual stage an MLP $f_\ell$ generates the entries from the current
reconstruction, $\mathcal{C}_\ell = f_\ell(\hat{z}_{\ell-1})$, so codes adapt per-sample and capture inter-stage dependencies. 

\noindent\textbf{\textit{Residual and Product Quantization.}}
These \emph{meta-schemes} wrap any base quantizer to expand its capacity. Residual quantization (RQ)~\cite{rqvae} quantizes the residual error over $L$ stages: from $r_0 = z_e$, each stage picks $e_{k_\ell} = \arg\min_{e \in \mathcal{C}_\ell}
\|r_{\ell-1} - e\|_2$ and updates $r_\ell = r_{\ell-1} - e_{k_\ell}$, giving $z_q = \sum_{\ell=1}^{L} e_{k_\ell}$ and up to $K^L$ vectors. Product quantization (PQ)~\cite{pq} partitions the latent into $m$ subvectors and quantizes each independently, yielding $(K')^m$ codes from $m \times K'$ centroids. Both are agnostic to the base quantizer, acting as composable building blocks.


\begin{table*}[t]
  \centering
  \caption{\textbf{Holistic reconstruction and generation overview.} Cross-dataset central tendency over the 12 datasets, grouped by compression $f$. Cells report the mean, except rFMD/gFMD, which use the more robust \emph{median} since Fr\'echet distances are sensitive to per-dataset sample counts (a few hundred to tens of thousands).$\uparrow$/$\downarrow$ indicate better direction; FMD~$=$~Fr\'echet Medical Distance (each dataset uses its own feature extractor). Statistical and per-dataset specific results can be found in \cref{sec:supp-recon,sec:supp-gen}.}
  \label{tab:recongen}
  \setlength{\tabcolsep}{4pt}
  \resizebox{\textwidth}{!}{%
  \begin{tabular}{lcccccccccccccccccc}
  \toprule
  & \multicolumn{6}{c}{$f{=}4$} & \multicolumn{6}{c}{$f{=}8$} & \multicolumn{6}{c}{$f{=}16$} \\
  \cmidrule(lr){2-7}\cmidrule(lr){8-13}\cmidrule(lr){14-19}
  Method & PSNR$\uparrow$ & SSIM$\uparrow$ & rFID$\downarrow$ & rFMD$\downarrow$ & gFID$\downarrow$ & gFMD$\downarrow$ & PSNR$\uparrow$ & SSIM$\uparrow$ & rFID$\downarrow$ & rFMD$\downarrow$ & gFID$\downarrow$ & gFMD$\downarrow$ & PSNR$\uparrow$ & SSIM$\uparrow$ & rFID$\downarrow$ & rFMD$\downarrow$ & gFID$\downarrow$ & gFMD$\downarrow$ \\
  \midrule
  AEKL~\cite{ldm} & 30.2 & 0.823 & 47.4 & 1.67 & 76.2 & 22.4 & 30.8 & 0.826 & 47.5 & 1.43 & 73.9 & 18.5 & 30.6 & 0.809 & 55.9 & 2.01 & 88.8 & 21.0 \\
  SoftVQ~\cite{softvq} & 34.4 & 0.911 & 21.9 & 0.23 & 79.9 & 20.2 & 30.2 & 0.817 & 50.4 & 1.27 & 107.9 & 38.9 & \underline{32.2} & \underline{0.820} & \underline{48.1} & \underline{0.93} & 113.6 & \textbf{10.1} \\
  VQGAN~\cite{vqgan} & 40.0 & \underline{0.953} & \underline{9.4} & \underline{0.05} & 50.5 & 26.4 & 34.0 & 0.847 & 43.7 & 0.44 & 65.5 & 18.0 & 29.3 & 0.756 & 79.2 & 2.22 & 99.5 & 15.0 \\
  SimVQ~\cite{simvq} & 39.1 & 0.946 & 11.7 & 0.09 & 49.4 & \underline{14.9} & 34.8 & 0.885 & 24.0 & \underline{0.19} & 63.2 & 31.5 & 28.8 & 0.762 & 82.2 & 2.86 & 111.8 & 34.0 \\
  SimQINCo~\cite{qinco} & \underline{40.1} & \underline{0.953} & 9.5 & 0.10 & 52.3 & 28.0 & 35.0 & 0.877 & 26.6 & 0.41 & \textbf{52.8} & 19.2 & 31.6 & 0.797 & 64.5 & 0.99 & 91.5 & 33.2 \\
  RQVAE~\cite{rqvae} & \textbf{41.3} & \textbf{0.963} & \textbf{7.0} & \textbf{0.04} & 52.2 & 25.9 & \textbf{37.1} & \textbf{0.907} & \textbf{19.0} & \textbf{0.08} & \underline{55.5} & \textbf{10.0} & \textbf{34.2} & \textbf{0.842} & \textbf{38.2} & \textbf{0.44} & \textbf{70.5} & 15.7 \\
  DiVeQ~\cite{diveq} & 39.2 & 0.940 & 12.7 & 0.10 & \textbf{48.1} & \textbf{11.5} & 35.3 & 0.884 & 24.2 & 0.20 & 58.4 & \underline{13.2} & 31.0 & 0.794 & 61.6 & 1.33 & 91.2 & 21.2 \\
  SF-DiVeQ~\cite{diveq} & 38.8 & 0.943 & 12.4 & \textbf{0.04} & \underline{48.7} & 26.1 & \underline{35.8} & \underline{0.892} & \underline{22.7} & \underline{0.19} & 58.8 & 17.1 & 31.3 & 0.793 & 57.0 & 1.18 & \underline{83.7} & \underline{11.2} \\
  BSQ~\cite{bsq} & 30.5 & 0.860 & 33.5 & 0.82 & 104.5 & 35.6 & 27.7 & 0.808 & 54.5 & 2.48 & 81.5 & 26.4 & 27.7 & 0.773 & 67.9 & 2.45 & 102.5 & 21.2 \\
  LFQ~\cite{lfq} & 29.4 & 0.860 & 40.1 & 0.94 & 141.9 & 56.0 & 24.0 & 0.708 & 103.8 & 13.37 & 226.1 & 38.7 & 23.2 & 0.655 & 148.0 & 19.24 & 138.5 & 31.9 \\
  \bottomrule
  \end{tabular}%
  }
\end{table*}

\section{Method}
\label{sec:method}

We benchmark all tokenizers within a single autoencoding architecture~\cite{vae}, the de facto autoencoder underlying latent diffusion, ensuring our findings remain representative of deployed medical pipelines. We vary only the quantization module, holding the encoder, decoder, training budget, and data fixed. We first describe this shared autoencoder and the quantizers we plug into it (\cref{sec:autoencoder}), then the generative model trained on the resulting latents (\cref{sec:generation}).

\subsection{Autoencoder and Tokenization}
\label{sec:autoencoder}

The encoder and decoder follow the convolutional VQ-GAN backbone~\cite{vqgan} and realize Eq.~\eqref{eq:1}: the encoder maps an image to a continuous latent $z_e$, the quantizer discretizes it to $z_q$, and the decoder reconstructs the input. At each setting we fix the spatial compression factor $f\in\{4,8,16\}$ and the latent channel dimension $d\in\{3,4,8\}$, so every quantizer operates on latents of identical shape. We remove all self-attention, yielding a purely convolutional encoder-decoder. With backbone and optimization fixed, per-$f$ differences are attributable to the quantizer alone. We omit self-attention, which the backbone instantiates only at certain resolutions (a subset of $f$), since its resolution-dependent global mixing would confound the cross-compression latent-geometry analysis (\S\ref{sec:geometry}).

\medskip
\noindent\textbf{\textit{Quantization Adaptations.}}
Most quantizers are either designed for images or directly applicable. The exception is QINCo~\cite{qinco}, built around an \emph{implicit neural codebook} where a small MLP generates the codebook entries rather than storing them. As QINCo's reconstruction-conditioned residual structure is computationally infeasible at image-tokenizer resolutions, we isolate the implicit codebook in a single-codebook variant named \emph{SimQINCo}: the table $\mathcal{C}\in\mathbb{R}^{K\times d}$ of a standard VQ layer is replaced by 
\begin{equation}
  \mathcal{C} = g_\phi(\mathbf{V}),
\end{equation}
a small shared MLP $g_\phi$ applied to a learnable coordinate matrix $\mathbf{V}\in\mathbb{R}^{K\times d}$, thereby evaluating the implicit neural codebook in isolation.

\medskip
\noindent\textbf{\textit{Training Objective.}}
All tokenizers share an identical training objective combining a pixel-space Mean Squared Error (MSE) reconstruction term, a VGG-based LPIPS perceptual loss~\cite{lpips}, a hinge adversarial loss~\cite{vqgan} against a PatchGAN discriminator~\cite{patchgan}, and the quantizer-specific auxiliary loss $\mathcal{L}_Q$ (commitment, entropy, or codebook regularization) prescribed by each individual quantization scheme:
\begin{equation}
  \mathcal{L}
  = \mathcal{L}_{\text{MSE}}(x,\hat{x})
  + \lambda_{\text{p}}\,\mathcal{L}_{\text{LPIPS}}(x,\hat{x})
  + \lambda_{\text{adv}}\,\mathcal{L}_{\text{adv}}(\hat{x})
  + \mathcal{L}_{Q},
  \label{eq:loss}
\end{equation}
with $\lambda_{\text{p}} = \lambda_{\text{adv}} = 0.01$; the continuous AEKL baseline additionally minimizes a small KL term ($\lambda_{\text{KL}} = 10^{-6}$). The perceptual term is linearly warmed up over the first 50k iterations. The adversarial term is held inactive for the first 100k steps and then linearly warmed up over a further 50k; thereafter its weight is adaptively rescaled by the gradient-norm ratio between $\mathcal{L}_{\text{MSE}}+\lambda_{\text{p}}\mathcal{L}_{\text{LPIPS}}$ and $\mathcal{L}_{\text{adv}}$~\cite{vqgan}. Additionally, to stabilize discriminator training  under the limited data scale typical of medical datasets, we additionally (i)~apply differentiable augmentation (color, translation, cutout) to real and fake inputs~\cite{diffaug}, (ii)~include a consistency-regularization term ($\lambda_{\text{CR}}=4$) penalizing inconsistent logits under augmentation~\cite{crgan}, and (iii)~add a LeCAM regularizer ($\lambda_{\text{LeCAM}}=10^{-3}$)~\cite{lecam} on the discriminator's real/fake logit EMAs.

\begin{figure*}[t]
    \centering
    \includegraphics[width=\linewidth]{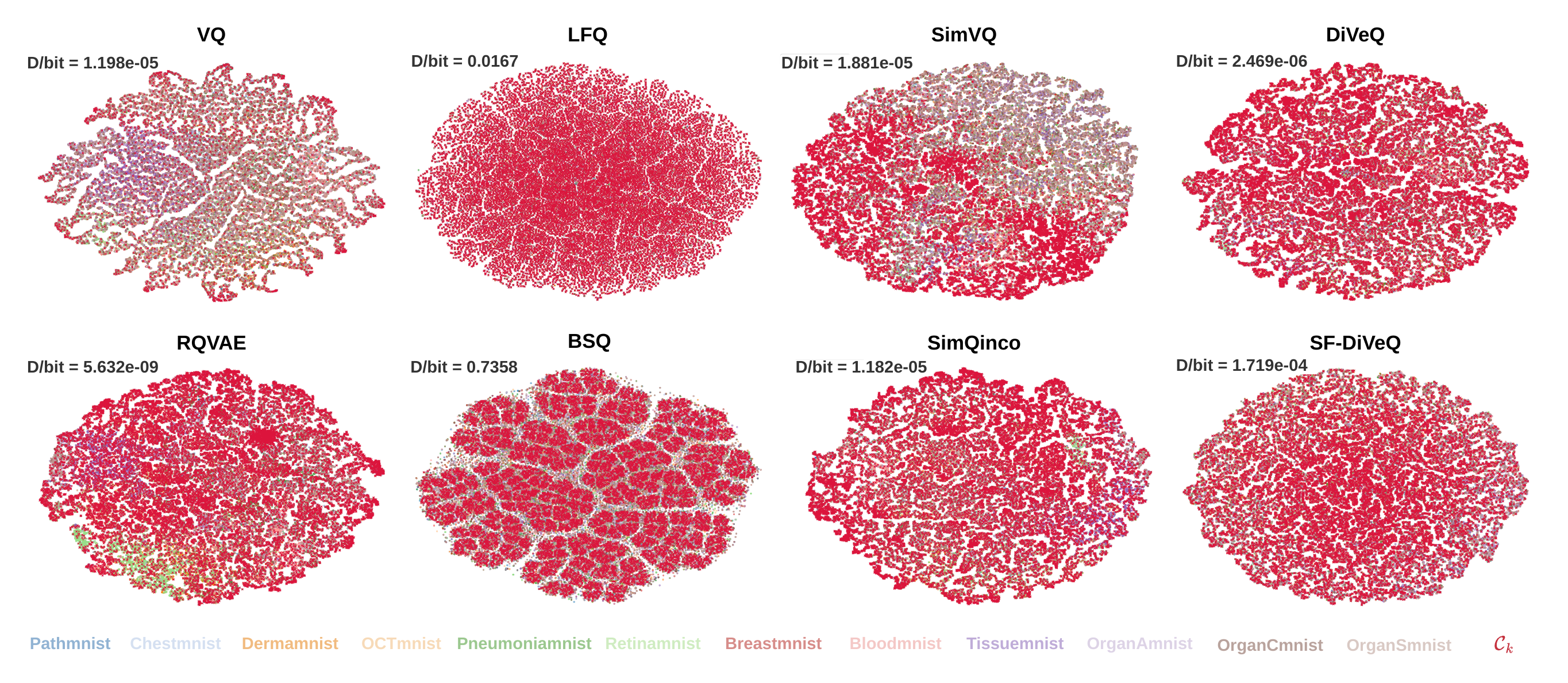}
    \caption{\textbf{\textit{Codebook misalignment across quantization families.}}
    Joint t-SNE of the pre-quant latents $\mathcal{P}_z$ (colored by source
    dataset) and the selected codewords $\mathcal{C}_z$ (red crosses), pooled
    over the test split of all twelve datasets, one panel per family at
    $f{=}8$; distortion-per-bit
    $\mathcal{D}_{\text{per-bit}}{=}\mathcal{D}/H(\mathcal{C})$ is
    annotated top-left. Codewords lying \emph{on} the latent cloud indicate good
    alignment; drift off it is the \emph{internal codebook covariate shift} of
    \cref{sec:exp-misalignment}.}
    \label{fig:misalignment}
\end{figure*}

\subsection{Generation}
\label{sec:generation}

To assess each tokenizer as a generative substrate, we freeze the trained autoencoder and fit a single diffusion model in its continuous pre-quantization latent $z_e$, rather than training an autoregressive transformer over the discrete $z_q$. The latter would entangle generation quality with codebook cardinality (the AR softmax scales as $K$, $K^L$, or $2^d$ across families), penalizing tokenizers for reasons unrelated to latent fidelity; the continuous $z_e$ is well-defined across all tokenizer families and admits a single shared generator, while also yielding a more robust inference pipeline since imperfectly denoised samples still snap to the correct codebook entry under the frozen quantizer, a property leveraged
by state-of-the-art medical latent-diffusion systems~\cite{medicaldiffusion-3d,meddiffusion3d}. Following standard latent-diffusion practice~\cite{ldm}, we therefore rescale every tokenizer's $z_e$ by a single scalar (the reciprocal of its standard deviation, estimated once on the training split) so that all generators operate on unit-variance latents. Then, following the latent-diffusion and DDPM paradigm~\cite{ldm,ddpm}, the forward process progressively corrupts a clean latent $z_0 = E(x)$ with Gaussian noise according to a variance schedule $\{\beta_t\}_{t=1}^{T}$,
\begin{equation}
\begin{aligned}
  q(z_t \mid z_0) &= \mathcal{N}\!\left(z_t;\, \sqrt{\bar{\alpha}_t}\, z_0,\,
  (1-\bar{\alpha}_t)\, \mathbf{I}\right), \\
  \bar{\alpha}_t &= \prod_{s=1}^{t}(1-\beta_s),
\end{aligned}
\label{eq:diffusion-forward}
\end{equation}
and a network $\epsilon_\theta$ is trained to predict the injected
noise~\cite{ddpm},
\begin{equation}
  \mathcal{L}_{\text{DM}}
  = \mathbb{E}_{z_0,\,t,\,\epsilon}
  \!\left[\,\lVert \epsilon - \epsilon_\theta(z_t, t) \rVert_2^2\,\right],
  \quad \epsilon \sim \mathcal{N}(0,\mathbf{I}).
  \label{eq:diffusion-loss}
\end{equation}
Sampling iteratively denoises $z_T \sim \mathcal{N}(0,\mathbf{I})$ back to
$z_0$, which the frozen quantizer snaps to $z_q$ before the decoder maps it
to pixel space.

\noindent\textbf{\textit{Conditioning.}} We instantiate $\epsilon_\theta$ as a DiT-B/2 transformer~\cite{dit}. Beyond the diffusion timestep $t$, the network is conditioned on two discrete signals: the class label $y$ and the source-dataset index $s$, the latter through a learned \emph{dataset embedding} that augments DiT's standard timestep and class embeddings, so a single network is shared across all twelve datasets. Each signal is mapped to a $d_{\text{emb}}$-dimensional vector ($h_t,h_y,h_s$); unlike vanilla DiT, which \emph{sums} the timestep and class embeddings, we \emph{concatenate} the three into a single conditioning vector $c=[h_t;h_y;h_s]\in\mathbb{R}^{3d_{\text{emb}}}$, preventing the class and dataset factors from interfering in a shared additive code. Conditioning enters every block via AdaLN-Zero~\cite{dit}: $c$ regresses the shift, scale, and gate applied to the self-attention and MLP sublayers, with the modulation zero-initialized so each block starts at identity.

\section{Experiments}
\label{sec:experiments}

\subsection{Setup}
\label{sec:setup}
We benchmark eight discrete quantization families: VQGAN~\cite{vqgan}, SimVQ~\cite{simvq}, SimQINCo~\cite{qinco}, RQVAE~\cite{rqvae}, DiVeQ~\cite{diveq},  SF-DiVeQ~\cite{diveq}, LFQ~\cite{lfq}, and BSQ~\cite{bsq}, together with continuous AEKL~\cite{ldm} and soft SoftVQ~\cite{softvq} baselines. Each method is trained at three spatial compression factors
$f\in\{4,8,16\}$ on the twelve 2D datasets of the MedMNIST+ collection~\cite{medmnist} at their native $224\times224$ resolution resulting in over 700{,}000 expert-labeled clinical images spanning pathology, radiology, dermatology,  ophthalmology, and microscopy, with multi-class, multi-label, and ordinal classification tasks. 

\noindent Following \cref{sec:method}, every tokenizer uses the identical encoder-decoder backbone, loss weights, and training
schedule; only the quantizer and compression factor vary, so all observed differences are attributable to the quantization scheme. Both stages, the tokenizer and its DiT-B/2 generator, are trained on 2$\times$ NVIDIA H100 GPUs for two days each. For image generation, we use 100 sampling steps on a linear schedule, following~\cite{ddpm}.

\subsection{Reconstruction and Generation}
\label{sec:exp-recongen}
We evaluate each tokenizer in two regimes. For \emph{reconstruction}, we encode and decode every test image and report PSNR, SSIM, LPIPS~\cite{lpips}, and reconstruction-FID (rFID) against the ground truth using \texttt{torch-fidelity}~\cite{torch-fidelity}. For \emph{generation}, we freeze the autoencoder, train a single DiT-B/2 diffusion model~\cite{dit} on its continuous latents (\cref{sec:generation}), sample per dataset as many images as its test split (conditioned on the dataset and, where applicable, class embedding to match the test-split class distribution) and report
generation-FID (gFID). Since both FIDs use an ImageNet-trained Inception-V3 backbone that is poorly calibrated to medical images, we additionally report \emph{FMD}, which swaps this backbone for the benchmark's official per-dataset ResNet-18 baseline classifier~\cite{medmnist} (head removed) so the distance is measured in a feature space tuned to each dataset. We compute it in both regimes as a domain-aware counterpart to rFID and gFID.

\subsection{Codebook Metrics}
\label{sec:exp-codebook}
We characterize the discrete bottleneck from the per-code selection counts $c_k$ accumulated over the entire test split, with empirical probabilities $p_k = c_k / \sum_j c_j$ over a codebook of size $K$. \emph{Usage} $U = \tfrac{1}{K}\sum_{k=1}^{K}\mathbb{1}[c_k>0]$ is the fraction of codes ever selected, and \emph{normalized entropy}
$\tilde{H} = H(\mathcal{C})/\log_2 K$, with $H(\mathcal{C}) = -\sum_{k=1}^{K} p_k \log_2 p_k$, measures how evenly
information is spread across the codebook. To probe how the latent space is used, we additionally report the \emph{stable rank} of the centered pre-quant latents $\mathbf{Z}\in\mathbb{R}^{N\times d}$, computed from their singular values $\sigma_1\ge\dots\ge\sigma_d$ as $\mathrm{sr}(\mathbf{Z}) = (\sum_{i}\sigma_i^2)/\sigma_1^2 \in [1,d]$, which
counts the effective number of dimensions used. 

\subsection{Codebook Misalignment}
\label{sec:exp-misalignment}
To visualize the relationship between the codebook and the latent distribution it quantizes, we encode the test images of all twelve datasets to their pre-quant latents $\mathcal{P}_z=\{z_n\}$, pool them, and collect the codewords actually selected, $\mathcal{C}_z$. We embed $\mathcal{P}_z \cup \mathcal{C}_z$ jointly with t-SNE and plot both in the shared 2D space, which exposes two failure modes: \emph{codebook collapse}, where only a small fraction of codewords is ever selected (a sparse $\mathcal{C}_z$), and \emph{internal codebook covariate shift}, where the used codes drift off the latent cloud they are meant to cover. We quantify the associated rate--distortion efficiency with the distortion-per-bit $\mathcal{D}_{\text{per-bit}} = \mathcal{D}/H(\mathcal{C})$, where $\mathcal{D} = \tfrac{1}{N}\sum_{n=1}^{N}\lVert z_n - z_{q,n}\rVert_2^2$ is the mean quantization error between the pre-quant latent $z_n$ and its quantized value $z_{q,n}$ (the cumulative sum over residual stages, where applicable) and $H(\mathcal{C})=\sum_{\ell=1}^{L}H(\mathcal{C}_\ell)$ is the empirical code-usage entropy. Charging distortion against the bits actually used rather than the allocated budget $B=\sum_{\ell=1}^{L}\log_2 K_\ell$ (\cref{tab:bitbudget}) penalizes collapse directly and prevents a larger codebook from appearing efficient, following~\cite{diveq}.

\subsection{Correlation}
\label{sec:exp-correlation}
To test whether reconstruction predicts generation, we compute both Pearson and Spearman correlations across all model configurations. The Pearson coefficient $r = \mathrm{Cov}(X,Y)/(\sigma_X \sigma_Y)$ captures linear association, while the Spearman coefficient $\rho = 1 - 6\sum_i \delta_i^2 / [\,n(n^2-1)\,]$, with $\delta_i = \mathrm{rank}(x_i) - \mathrm{rank}(y_i)$, captures monotonic rank agreement and is robust to FID's heavy-tailed scale; reporting both guards against a single outlier family driving the trend. We attach 95\% confidence intervals via the non-parametric bootstrap ($B{=}10{,}000$ resamples with replacement, $2.5$th/$97.5$th percentiles), which is rank- and distribution-free and thus well-suited to FID's heavy-tailed scale and our moderate sample size of a few hundred \emph{(tokenizer,
dataset)} pairs.

\begin{figure}[t]
    \centering
    \includegraphics[width=\linewidth]{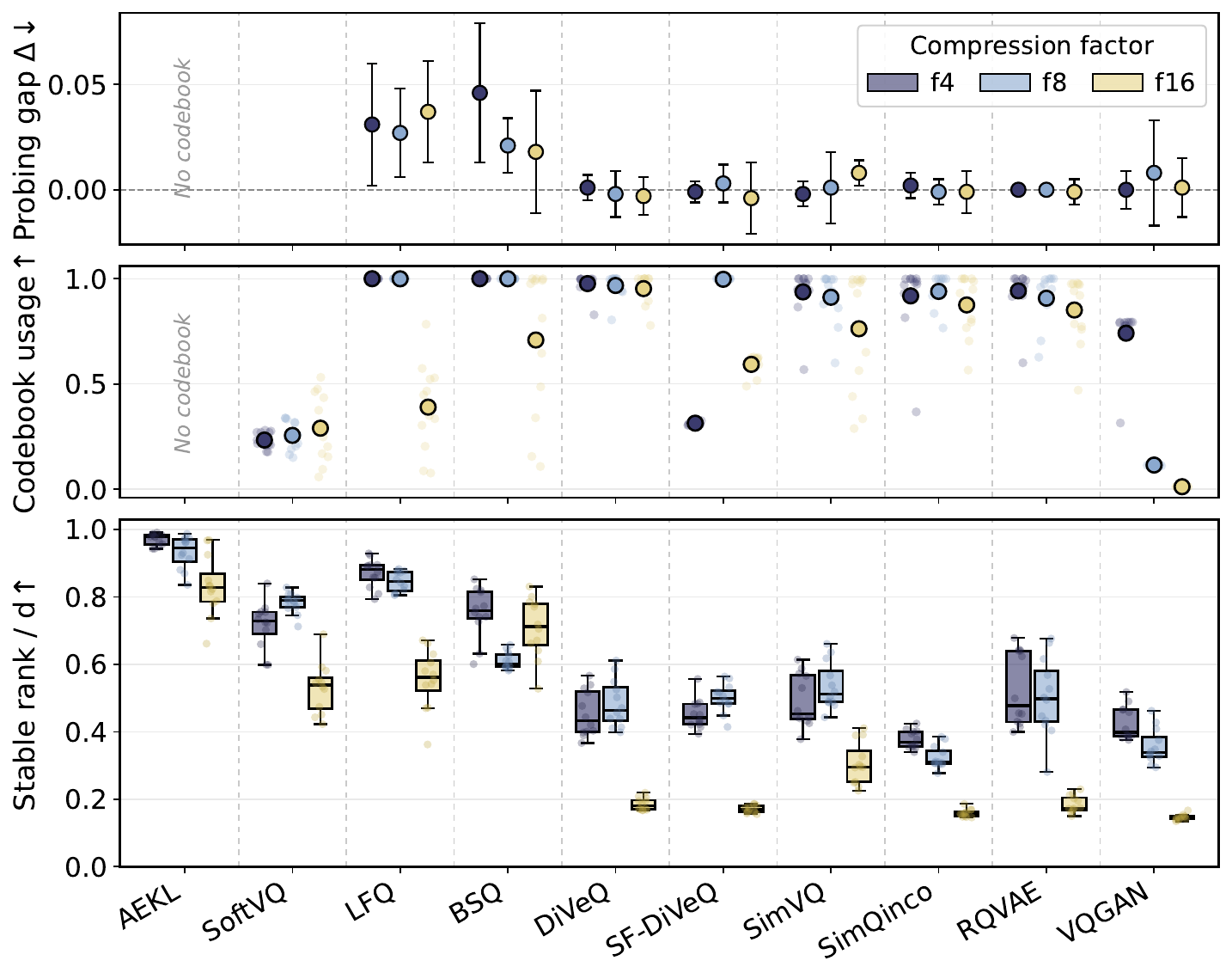}
    \caption{\textbf{\textit{Codebook utilization and latent geometry are decoupled.}} Three tokenizer properties grouped by family and compression. Modern schemes reach near-full codebook utilization \emph{(mid)} yet collapse their latents onto a thin, low-rank subspace \emph{(bottom)}, showing that a fully used codebook does not imply an expressive latent space and that codebook usage alone is a misleading indicator of tokenizer quality. Discretization largely preserves downstream-useful information \emph{(top)}.}
    \label{fig:codebook_geometry}
\end{figure}

\begin{figure}[t]
    \centering
    \includegraphics[width=\linewidth]{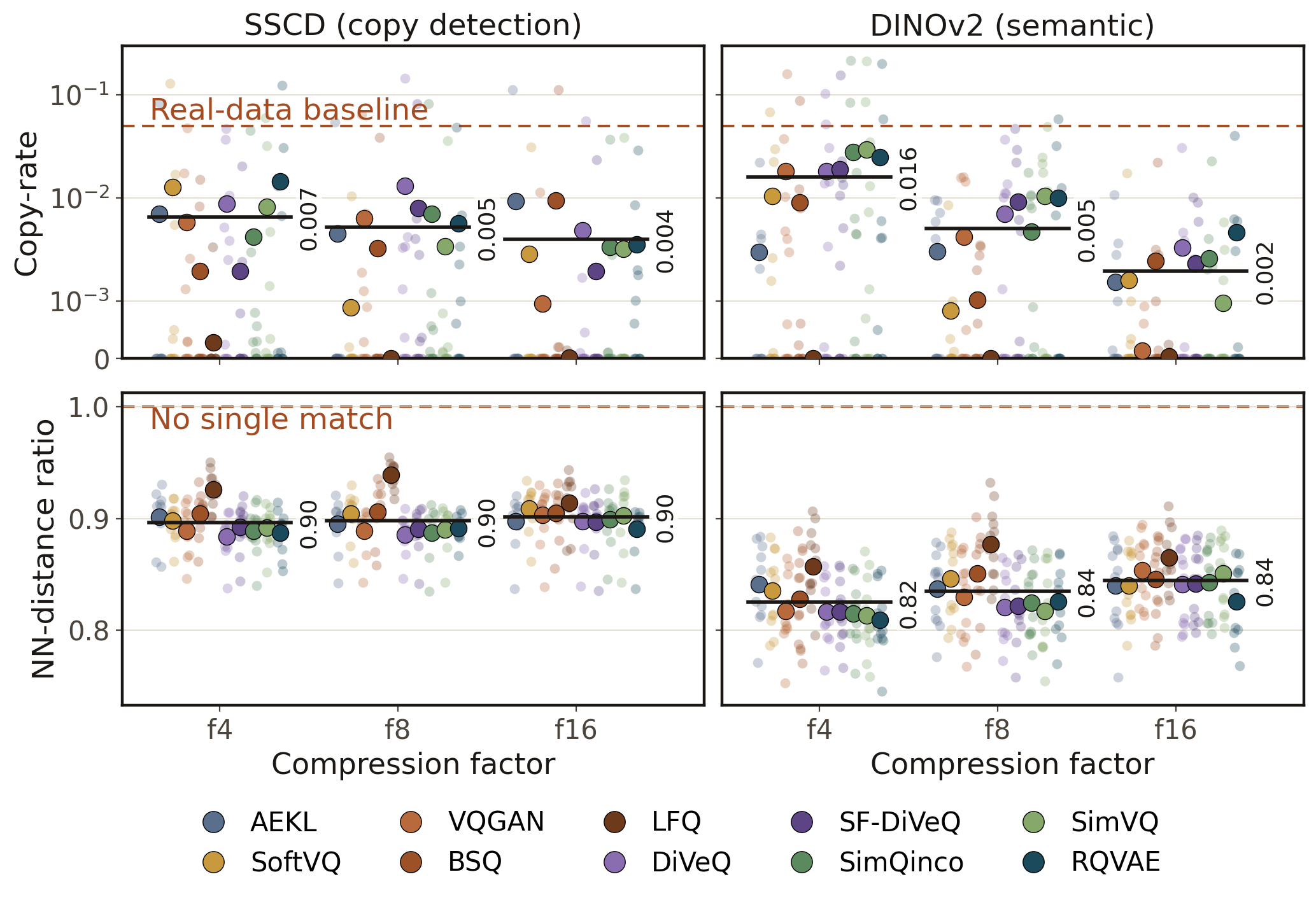}
    \caption{\textbf{Memorization is mild and weakens under stronger compression.}
    \emph{Columns:} SSCD (copy-detection) and DINOv2 (semantic) extractors.
    \emph{Top:} copy-rate vs.\ compression factor f, with one point per (family, dataset) and black bars denoting per-f means. Rates remain far below the real-data baseline (dashed) and decrease as the bottleneck tightens. The NN-distance ratio stays well under the "no single match" line.}
    \label{fig:memorization}
\end{figure}

\subsection{Memorization}
\label{sec:exp-memorization}
To test whether a tokenizer's bottleneck leaks training data, we adopt a nearest-neighbor replication test in the spirit of Somepalli et al.~\cite{somepalli2023diffusion}. For each (tokenizer, dataset) configuration we embed every generated image and, as a calibrated reference, every held-out test image, and retrieve their nearest neighbors in the full training set. Embeddings come from two complementary encoders: SSCD~\cite{sscd}, purpose-built for copy detection, and DINOv2~\cite{dinov2}, which captures higher-level semantic similarity; a configuration is flagged only when both agree.
We report \emph{copy-rate}, the fraction of generated images whose top-1 training similarity exceeds the $95$th percentile of the test-to-train similarity distribution, a per-dataset threshold that makes the real held-out data its own baseline ($\approx0.05$ by construction), so that values above baseline indicate copying beyond what genuinely novel images exhibit. As a complementary view we also report the distance ratio between each generated image's nearest and its next-$k$ training neighbors ($k{=}50$), where values near one indicate no single memorized match. We cap each dataset at $10{,}000$ generated images and compare against the complete training split.

\subsection{Semantics and Downstream Evaluation}
\label{sec:exp-downstream}
To test if quantization preserves the semantically useful information needed for downstream tasks, we evaluate each frozen tokenizer through linear probing on the datasets' labeled classes. For every model we extract two representation variants, the continuous pre-quantization latent $z_e$ and its quantized counterpart $z_q$, both flattened across the spatial dimensions into a single feature vector and standardized to zero mean and unit variance. On top of each variant, we fit a single linear classifier for 50 epochs with the tokenizer held frozen, select the best checkpoint on the validation split, and report test performance under each dataset's task-appropriate primary metric (top-1 accuracy for multi-class tasks, macro-AUROC for multi-label tasks, and quadratic-weighted kappa for the ordinal grading task). The per-dataset gap $\Delta = \mathrm{M}(z_e) - \mathrm{M}(z_q)$ in the primary metric $\mathrm{M}$ quantifies the change in class-discriminative structure across the discretization step itself, isolating the effect of quantization from the encoder's representation capacity.


\section{Results and Discussion}
\label{sec:results}



\noindent\textbf{\textit{Reconstruction and generation performance in medical imaging strongly correlate.}}
\label{sec:recon-gen}
Across all configurations, reconstruction metrics are strongly predictive of generation quality: rFID and gFID correlate at $r{=}0.82$ with $95\%$ bootstrap CI $[0.77, 0.86]$ ($B{=}10{,}000$ resamples over $n{=}360$ paired observations; Spearman $\rho{=}0.81\,[0.77, 0.85]$) (\cref{fig:correlations}), and PSNR/LPIPS show the same trend (per-method values in \cref{tab:recongen}). Moreover, \cref{sec:supp-correlation} disentangles the correlation by dataset and by compression factor, showing it is not an artifact of pooling since the same relationship holds across each stratum. This contrasts with \emph{reported} natural-image results, where reconstruction is a poor
proxy for generation~\cite{recon-vs-gen,xu2026makingrfid,rfid-gfid}. We
conjecture this reflects the distinct statistics of medical data, stronger anatomical priors, lower intra-class variance, and limited texture diversity, which yield latent spaces whose reconstructability and generative modelability move together; whether representation-aligned tokenizers would follow a different regime remains open. This is practically valuable since judging a tokenizer by generation requires training a generator per configuration while reconstruction is free from the autoencoder, the correlation lets practitioners select tokenizers and search architectures using rFID alone, reserving generator training for a shortlist. It further makes reconstruction a legitimate optimization target, since reducing reconstruction error is expected to improve generation.

\noindent\textbf{\textit{Decoupled codebook usage and latent geometry.}}
\label{sec:geometry}
Nearly every modern scheme cures the canonical VQ collapse: as $f$ grows from $4$ to $16$, VQGAN usage falls from $79\%$ to $1.2\%$, yet reparameterized (SimVQ, SimQINCo), residual (RQVAE), and lookup-free (LFQ, BSQ) methods all hold $\approx100\%$ utilization (\cref{fig:codebook_geometry}; only SF-DiVeQ and SoftVQ collapse partially). But high usage does not imply an efficient latent space: explicit-codebook methods, \emph{even at full utilization}, compress their latents onto a thin, anisotropic subspace (stable rank $0.15$--$0.54$ of $d$), whereas lookup-free ($0.55$--$0.87$) and continuous ($0.83$--$0.97$) methods stay near-isotropic. This could be driven by the general quantization scheme, not the data or the gradient estimator: nearest-neighbor lookup lets the encoder collapse onto a low-rank manifold the codebook can cover, and it is no straight-through artifact, since DiVeQ propagates true gradients through the quantizer yet collapses just like the rest. The immediate consequence is definitional: since the encoder packs information into a fraction of the available dimensions, the nominal latent dimensionality overstates how much of the space each tokenizer actually uses. Beyond this, the collapse appears benign; stable rank shows no clear effect on either reconstruction or generation, so we treat it as an intrinsic geometric property of these quantizers and an axis independent of downstream quality rather than a driver of it.

\noindent\textbf{\textit{Codebook collapse and misalignment.}}
\label{sec:misalignment}
A joint t-SNE of the pre-quant latents $\mathcal{P}_z$ and the selected codes $\mathcal{C}_z$ (\cref{fig:misalignment}) serves as a more diagnostic read-out of tokenizer quality than codebook usage alone, capturing both how much of the codebook a model uses and how well it places that codebook onto the data manifold. It reveals that usage and geometric alignment with the data are independent axes, and that each family trades them off in a distinct way. Vanilla VQGAN displays the canonical codebook collapse pattern: only a few live codewords concentrate tightly on the densest region of $\mathcal{P}_z$ and reconstruct what they cover at low distortion, but most codewords go unused, leaving most of the latent cloud unreached. Lookup-free schemes (LFQ, BSQ) sit at the opposite extreme: their implicit codebooks spread broadly across the embedding space, yielding nearly full usage but settling far from the data manifold, which translates to high quantization error despite the healthy utilization numbers. Codebook-reparameterization methods fall in between: SimVQ exhibits a usage shift in which some datasets are covered densely while others are left sparse, whereas SimQINCo blankets all twelve datasets uniformly and at lower distortion than SimVQ. The differentiable schemes achieve both wide coverage and low distortion, with DiVeQ attaining the lowest $\mathcal{D}_{\text{per-bit}}$ among the single-codebook methods; SF-DiVeQ, despite interpolating along codeword segments, does not improve on it here. Residual quantization (RQVAE) attains the lowest distortion overall, though at a higher bit rate (\cref{tab:bitbudget}), so its low $\mathcal{D}_{\text{per-bit}}$ partly reflects a higher operating rate.

\noindent\textbf{\textit{Memorization is mild and decreases under stronger compression.}}
\label{sec:memorization}
A tokenizer that reproduces its training data undermines the privacy argument for synthetic imaging, so we ask whether the tokenizer leaks data. \cref{fig:memorization} shows it largely does not: across all $720$ configurations mean copy-rate stays well below the real-data baseline, and the nearest-neighbor distance ratio remains near one, meaning generated images sit
in generic high-density regions rather than locking onto any single example. Outright replication is rare, so on aggregate the bottleneck withholds identity as required. This is not an underfitting artifact as copy-rate is essentially uncorrelated with generation quality and the few elevated copy-rates come from the best generators (\cref{fig:mem-vs-gfid}).

The exceptions concentrate on Path and OCT, for partly different reasons (\cref{tab:dataset-variance}). Path is among the least diverse datasets by every measure (low effective rank, low pairwise distance, high nearest-neighbor similarity), so its near-duplicate scans leave a generator little room for novelty. OCT only looks diverse in raw pixels, where speckle inflates variance; its B-scans share a stereotyped layered structure and stay semantically repetitive. Thus, memorization is rather driven by data than by tokenizer design. 

The only design choice that matters is compression: memorization falls steadily as the bottleneck tightens, because a coarser latent simply cannot carry the detail that makes a sample traceable. This is the same trade-off we see on our other axes: the compression that protects privacy also discards diagnostic and representational signal.

\noindent\textbf{\textit{Discretization largely preserves downstream-useful information.}}
\label{sec:downstream}
Finally, we probe how much semantically useful information survives
quantization by training linear classifiers on the frozen latents
(\cref{tab:probe-gap}). Discretization is largely information-preserving: for
explicit-codebook (VQGAN, SimVQ, SimQINCo), residual (RQVAE), and
differentiable (DiVeQ, SF-DiVeQ) schemes the probe gap
$\Delta=\mathrm{M}(z_e)-\mathrm{M}(z_q)$ is statistically indistinguishable
from zero, i.e.\ the discrete latent $z_q$ matches the continuous bottleneck
$z_e$ to within noise. The exception is the lookup-free family (LFQ, BSQ),
whose per-dimension binarization incurs a consistent gap
($\Delta\approx 0.02$--$0.05$) across all three compression factors.
Strikingly, this is \emph{independent} of latent rank: lookup-free latents are
the highest-rank of all families (\cref{sec:geometry}), yet lose the most
downstream-useful structure---so the erosion stems from the hard scalar
quantization grid, not from the low-rank geometry of explicit-codebook methods.
Downstream utility is thus a largely \emph{separate} axis from reconstruction
and generation, one on which most quantizers are safe and only lookup-free
binarization pays a price, a consideration absent from natural-image tokenizer
design.

\noindent\textbf{\textit{Summary: Practitioner's Guide in medical imaging.}}
Under our evaluated setting we observe:
\label{sec:pract-guide}
\begin{itemize}
\item \textbf{Reconstruction still has headroom.} Reconstruction has not yet
saturated and, since it tends to predict generation ($r{=}0.82$), it appears to
be a reasonable lever to keep improving, with limited privacy risk given that
memorization remains mild in our experiments.
\item \textbf{Discrete tokenizers are competitive substrates for continuous-latent diffusion.} In our setup, discrete
tokenizers tend to match or outperform the continuous baseline on reconstruction
and generation while preserving downstream utility comparably.
\item \textbf{Residual quantization is a strong default.} For a favorable
compute-to-performance trade-off, residual quantization (RQVAE) is a reasonable
first choice, performing well across compression factors with little tuning.
\end{itemize}

\section{Conclusion}
\label{sec:conclusion}
Within our benchmark, medical image tokenization departs from natural-image expectations. The field's inherited assumptions break: reconstruction predicts generation, a healthy codebook hides a collapsed latent, and
discretization is largely free for both privacy and downstream use. Within this setting, there is no single winner, but a clear practitioner's guide emerges, and reconstruction remains a lever worth pushing.

\noindent\textbf{Limitations and Future Work.}
Our study covers only 2D medical data. Higher-resolution and volumetric data, where compression and memory constraints differ, remain unexplored. We assess each tokenizer through a continuous-latent diffusion model rather than autoregressive modeling of the discrete codes, with a single fixed DiT-B/2 generator, so generation quality reflects latent fidelity but not code-sequence learnability or generator capacity. Downstream utility is gauged via linear probing, a deliberately simple proxy for richer clinical tasks. Our natural-versus-medical comparison also relies on reported natural-image numbers rather than a matched control, which we leave to future work alongside representation-aligned tokenizers. Looking forward, the decoupling of codebook usage from latent geometry suggests designing quantizers that target the latent-rank axis directly rather than chasing utilization. Finally, our work establishes \emph{how} medical tokenization departs from the natural-image regime but not yet \emph{why}; pinning down the cause, be it data scarcity, low inter-sample variability, limited texture diversity, or strong anatomical priors, is the key open question.



{
    \small
    \bibliographystyle{ieeenat_fullname_seq}
    \bibliography{main}
}

\setcounter{dbltopnumber}{2}
\setcounter{totalnumber}{4}
\renewcommand{\dbltopfraction}{0.95}
\renewcommand{\dblfloatpagefraction}{0.55}
\renewcommand{\topfraction}{0.95}
\renewcommand{\floatpagefraction}{0.85}
\renewcommand{\textfraction}{0.05}
\setlength{\floatsep}{14pt plus 2pt minus 2pt}
\setlength{\dblfloatsep}{14pt plus 2pt minus 2pt}
\setlength{\textfloatsep}{16pt plus 2pt minus 4pt}

\appendix

\twocolumn[{%
  \centering
  \vspace*{0.5em}
  {\Large\bfseries What Makes a Good Medical Image Tokenizer? \\Rethinking Reconstruction and Generation in Medical Image Tokenization\par}
  \vspace{0.6em}
  {\large\bfseries Supplementary Material\par}
  \vspace{1.4em}
}]

\renewcommand{\thesection}{\Alph{section}}
\renewcommand{\thesubsection}{\Alph{section}.\arabic{subsection}}
\renewcommand{\thetable}{S\arabic{table}}
\renewcommand{\thefigure}{S\arabic{figure}}

\section{Correlation}
\label{sec:supp-correlation}
\begin{table}[h]
  
  \centering
  \caption{Per-dataset correlation of \textbf{PSNR} with DiT-B/2 generation-FID, with $95\%$ bootstrap CIs ($B{=}10{,}000$). $n{=}30$ \emph{(tokenizer, compression)} configs per dataset; pooled $n{=}360$. PSNR correlates negatively (higher PSNR $\rightarrow$ lower/better gFID).}
  \label{supp:tab-corr-psnr}
  \setlength{\tabcolsep}{4pt}
  \adjustbox{max width=0.48\textwidth,max totalheight=0.43\textheight,center}{%
  \begin{tabular}{lcc}
  \toprule
  Dataset & Pearson $r$ \,[95\% CI] & Spearman $\rho$ \,[95\% CI] \\
  \midrule
  Path   & $-0.63$ \,$[-0.78, -0.37]$ & $-0.67$ \,$[-0.85, -0.38]$ \\
  Chest  & $-0.73$ \,$[-0.89, -0.65]$ & $-0.89$ \,$[-0.95, -0.73]$ \\
  Derma  & $-0.76$ \,$[-0.88, -0.59]$ & $-0.74$ \,$[-0.86, -0.50]$ \\
  OCT    & $-0.76$ \,$[-0.92, -0.59]$ & $-0.82$ \,$[-0.94, -0.57]$ \\
  Pneu   & $-0.51$ \,$[-0.70, -0.18]$ & $-0.43$ \,$[-0.69, -0.07]$ \\
  Reti   & $-0.72$ \,$[-0.84, -0.54]$ & $-0.77$ \,$[-0.87, -0.57]$ \\
  Breast & $-0.83$ \,$[-0.91, -0.71]$ & $-0.89$ \,$[-0.95, -0.74]$ \\
  Blood  & $-0.57$ \,$[-0.75, -0.34]$ & $-0.57$ \,$[-0.75, -0.26]$ \\
  Tiss   & $-0.66$ \,$[-0.80, -0.47]$ & $-0.63$ \,$[-0.78, -0.36]$ \\
  OrgA   & $-0.82$ \,$[-0.91, -0.72]$ & $-0.88$ \,$[-0.94, -0.70]$ \\
  OrgC   & $-0.78$ \,$[-0.87, -0.67]$ & $-0.84$ \,$[-0.91, -0.69]$ \\
  OrgS   & $-0.76$ \,$[-0.85, -0.66]$ & $-0.80$ \,$[-0.88, -0.64]$ \\
  \midrule
  \textbf{Pooled} & $\mathbf{-0.67}$ \,$[-0.71, -0.64]$ & $\mathbf{-0.74}$ \,$[-0.79, -0.69]$ \\
  \bottomrule
  \end{tabular}}
\end{table}

\begin{table}[h]
  
  \centering
  \caption{Per-dataset correlation of \textbf{rFID} with DiT-B/2 generation-FID, with $95\%$ bootstrap CIs ($B{=}10{,}000$). $n{=}30$ \emph{(tokenizer, compression)} configs per dataset; pooled $n{=}360$. rFID correlates positively with gFID across every dataset.}
  \label{supp:tab-corr-rfid}
  \setlength{\tabcolsep}{4pt}
  \adjustbox{max width=0.48\textwidth,max totalheight=0.43\textheight,center}{%
  \begin{tabular}{lcc}
  \toprule
  Dataset & Pearson $r$ \,[95\% CI] & Spearman $\rho$ \,[95\% CI] \\
  \midrule
  Path   & $+0.61$ \,$[+0.17, +0.84]$ & $+0.54$ \,$[+0.17, +0.78]$ \\
  Chest  & $+0.79$ \,$[+0.73, +0.97]$ & $+0.87$ \,$[+0.69, +0.95]$ \\
  Derma  & $+0.75$ \,$[+0.60, +0.90]$ & $+0.76$ \,$[+0.55, +0.87]$ \\
  OCT    & $+0.80$ \,$[+0.67, +0.98]$ & $+0.93$ \,$[+0.79, +0.98]$ \\
  Pneu   & $+0.48$ \,$[+0.07, +0.76]$ & $+0.38$ \,$[+0.06, +0.63]$ \\
  Reti   & $+0.82$ \,$[+0.59, +0.93]$ & $+0.77$ \,$[+0.55, +0.88]$ \\
  Breast & $+0.92$ \,$[+0.84, +0.97]$ & $+0.92$ \,$[+0.82, +0.96]$ \\
  Blood  & $+0.42$ \,$[+0.11, +0.80]$ & $+0.42$ \,$[+0.10, +0.65]$ \\
  Tiss   & $+0.71$ \,$[+0.24, +0.95]$ & $+0.65$ \,$[+0.35, +0.83]$ \\
  OrgA   & $+0.77$ \,$[+0.62, +0.92]$ & $+0.87$ \,$[+0.69, +0.95]$ \\
  OrgC   & $+0.72$ \,$[+0.53, +0.88]$ & $+0.79$ \,$[+0.62, +0.89]$ \\
  OrgS   & $+0.69$ \,$[+0.49, +0.87]$ & $+0.74$ \,$[+0.54, +0.85]$ \\
  \midrule
  \textbf{Pooled} & $\mathbf{+0.82}$ \,$[+0.77, +0.86]$ & $\mathbf{+0.81}$ \,$[+0.77, +0.85]$ \\
  \bottomrule
  \end{tabular}}
\end{table}

\begin{table}[t]
\centering
\caption{Per-compression correlation of \textbf{PSNR} with DiT-B/2
generation-FID, with 95\% bootstrap CIs ($B{=}10{,}000$). PSNR correlates negatively (higher PSNR $\rightarrow$ lower/better
gFID) at every compression.}
\label{tab:corr-psnr-comp}
\resizebox{\columnwidth}{!}{%
\begin{tabular}{lcc}
\toprule
Compression & Pearson $r$ [95\% CI] & Spearman $\rho$ [95\% CI] \\
\midrule
$f{=}4$  & $-0.68$ $[-0.75, -0.61]$ & $-0.73$ $[-0.80, -0.64]$ \\
$f{=}8$  & $-0.72$ $[-0.78, -0.66]$ & $-0.79$ $[-0.85, -0.70]$ \\
$f{=}16$ & $-0.61$ $[-0.69, -0.52]$ & $-0.62$ $[-0.73, -0.48]$ \\
\midrule
\textbf{Pooled} & $\mathbf{-0.67}$ $[-0.71, -0.64]$ & $\mathbf{-0.74}$ $[-0.79, -0.69]$ \\
\bottomrule
\end{tabular}}
\end{table}

\begin{table}[t]
\centering
\caption{Per-compression correlation of \textbf{rFID} with DiT-B/2
generation-FID, with 95\% bootstrap CIs ($B{=}10{,}000$). rFID correlates positively with gFID at every compression, and the
association strengthens as the bottleneck tightens.}
\label{tab:corr-rfid-comp}
\resizebox{\columnwidth}{!}{%
\begin{tabular}{lcc}
\toprule
Compression & Pearson $r$ [95\% CI] & Spearman $\rho$ [95\% CI] \\
\midrule
$f{=}4$  & $+0.71$ $[+0.63, +0.78]$ & $+0.76$ $[+0.68, +0.82]$ \\
$f{=}8$  & $+0.86$ $[+0.81, +0.91]$ & $+0.88$ $[+0.82, +0.91]$ \\
$f{=}16$ & $+0.90$ $[+0.84, +0.94]$ & $+0.85$ $[+0.76, +0.91]$ \\
\midrule
\textbf{Pooled} & $\mathbf{+0.82}$ $[+0.77, +0.86]$ & $\mathbf{+0.81}$ $[+0.77, +0.85]$ \\
\bottomrule
\end{tabular}}
\end{table}

\begin{figure}[t]
  \centering
  \includegraphics[width=\linewidth]{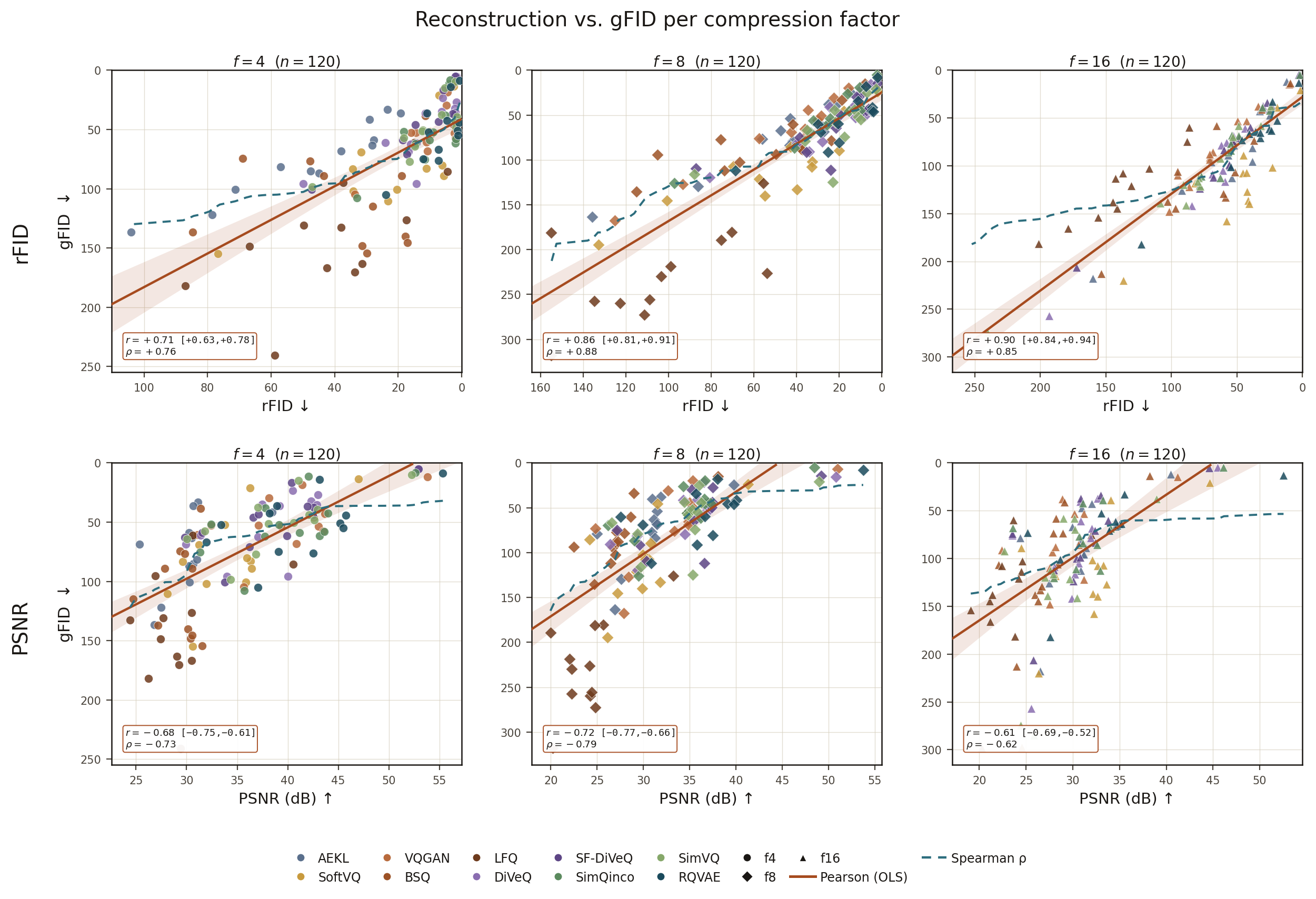}
  \caption{\textbf{The reconstruction--generation coupling strengthens with
  compression.} Reconstruction metric vs.\ DiT-B/2 gFID per compression factor
  ($n{=}120$ each). \emph{Top:} rFID; \emph{bottom:} PSNR. Pearson OLS ($95\%$ band)
  and Spearman fits, with $r$ $[95\%$ CI$]$ and $\rho$ inset. The coupling tightens
  as $f{=}4\!\rightarrow\!8\!\rightarrow\!16$.}
  \label{fig:corr-percomp}
\end{figure}

\begin{figure*}[t]
  \centering
  \includegraphics[width=\textwidth]{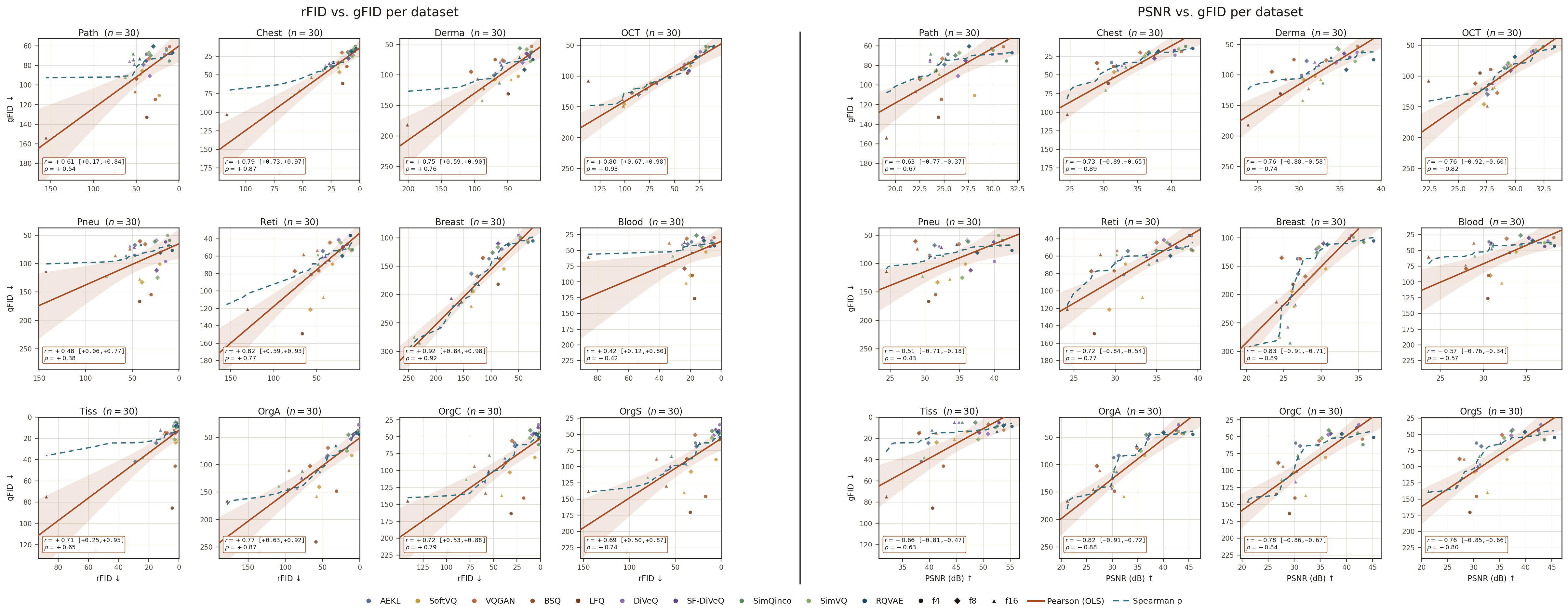}
  \caption{\textbf{Reconstruction predicts generation \emph{within} every
  dataset.} Per-dataset scatter of a reconstruction metric against DiT-B/2 gFID,
  one panel per MedMNIST+ dataset ($n{=}30$ (tokenizer, compression) configs each),
  colored by tokenizer family and marked by compression factor. \emph{Left block:}
  rFID; \emph{right block:} PSNR. Each panel overlays a Pearson OLS fit (with a
  $95\%$ band) and a Spearman monotone fit; the inset box reports Pearson $r$
  $[95\%$ CI$]$ and Spearman $\rho$. rFID correlates positively and PSNR negatively
  with gFID in all twelve datasets, confirming that the pooled trend is not an
  artifact of aggregation.}
  \label{fig:corr-perds}
\end{figure*}

\section{Bit-Compression}
\label{sec:bit-comp}
While we aim to keep the bit budget comparable across quantizers, some variation is inevitable, since each family encodes its latents differently: single-codebook methods spend one index per token, RQVAE spends several across its residual stages, and lookup-free schemes spend one bit per latent dimension. \Cref{tab:bitbudget} lists the resulting per-token budget $B$ for each family, which we use to normalize distortion into the distortion-per-bit $\mathcal{D}_{\text{per-bit}}=\mathcal{D}/B$ of \cref{sec:exp-misalignment}.

\begin{table}[h]
  \centering
  \caption{Per-token bit budget $B=\sum_{\ell=1}^{L}\log_2 K_\ell$ used to
  normalize distortion, $\mathcal{D}_{\text{per-bit}}=\mathcal{D}/B$
  (\cref{sec:exp-misalignment}). Single-codebook families spend one index
  ($B=\log_2 K$, $L{=}1$); RQVAE uses $L{=}4$ residual stages of size $K$;
  lookup-free families spend one bit per latent dimension ($B=d$). Codebooks are
  $K{=}8192$ at $f{=}4$ and $K{=}16384$ at $f{=}8,16$. AEKL is continuous (no
  discrete budget); SoftVQ's budget is nominal as its assignment is soft.}
  \label{tab:bitbudget}
  \setlength{\tabcolsep}{5pt}
  \begin{tabular}{lccc}
  \toprule
  Family ($B$) & $f{=}4$ & $f{=}8$ & $f{=}16$ \\
  \midrule
  VQGAN ($\log_2 K$)           & 13  & 14  & 14  \\
  SimVQ ($\log_2 K$)           & 13  & 14  & 14  \\
  SimQINCo ($\log_2 K$)     & 13  & 14  & 14  \\
  DiVeQ ($\log_2 K$)           & 13  & 14  & 14  \\
  SF-DiVeQ ($\log_2 K$)        & 13  & 14  & 14  \\
  RQVAE ($4\log_2 K$)          & 52  & 56  & 56  \\
  LFQ ($d$)                    & 10  & 14  & 18  \\
  BSQ ($d$)                    & 10  & 14  & 18  \\
  \bottomrule
  \end{tabular}
\end{table}

\section{Quantitative Reconstruction}
\label{sec:supp-recon}
We report pixel and perceptual fidelity per dataset---PSNR (\cref{tab:psnr}),
SSIM (\cref{tab:ssim}), LPIPS (\cref{tab:lpips}), and pixel-MSE
(\cref{tab:mse})---together with reconstruction-FID (\cref{tab:rfid}). RQVAE is
the strongest reconstructor on nearly every dataset and compression factor,
whereas the lookup-free families (LFQ, BSQ) trail throughout and degrade sharply
at $f{=}16$.

\begin{table*}[h]
  \centering
  \caption{Per-dataset PSNR ($\uparrow$). Cells: $\text{mean}_{\pm\sigma}$ over test images.}
  \label{tab:psnr}
  \setlength{\tabcolsep}{3pt}
  \adjustbox{max width=\textwidth,max totalheight=0.43\textheight,center}{%
  \begin{tabular}{lcccccccccccc|c}
  \toprule
  Method & Path & Chest & Derma & OCT & Pneu & Reti & Breast & Blood & Tiss & OrgA & OrgC & OrgS & mean \\
  \midrule
  \multicolumn{14}{l}{\textit{$f = 4$}} \\
  AEKL & 25.4$_{\pm 3.1}$ & 31.1$_{\pm 1.3}$ & 30.3$_{\pm 2.0}$ & 27.5$_{\pm 1.6}$ & 30.7$_{\pm 1.5}$ & 31.0$_{\pm 1.6}$ & 26.8$_{\pm 1.1}$ & 30.6$_{\pm 1.9}$ & 38.5$_{\pm 2.0}$ & 30.3$_{\pm 2.6}$ & 30.3$_{\pm 3.2}$ & 30.5$_{\pm 3.2}$ & 30.2 \\
  SoftVQ & 28.1$_{\pm 3.1}$ & 36.2$_{\pm 1.6}$ & 32.0$_{\pm 2.2}$ & 29.6$_{\pm 1.8}$ & 35.8$_{\pm 1.8}$ & 31.2$_{\pm 2.5}$ & 30.6$_{\pm 1.3}$ & 33.8$_{\pm 1.2}$ & 47.0$_{\pm 2.2}$ & 36.3$_{\pm 3.9}$ & 36.0$_{\pm 4.3}$ & 36.4$_{\pm 4.2}$ & 34.4 \\
  VQGAN & 31.1$_{\pm 3.4}$ & 41.4$_{\pm 1.5}$ & 37.1$_{\pm 2.1}$ & 32.4$_{\pm 2.0}$ & 40.8$_{\pm 1.5}$ & \textbf{39.4$_{\pm 1.4}$} & 35.7$_{\pm 1.3}$ & 38.1$_{\pm 1.8}$ & 53.8$_{\pm 0.8}$ & 43.6$_{\pm 4.4}$ & 43.0$_{\pm 4.7}$ & 43.5$_{\pm 4.7}$ & 40.0 \\
  SimVQ & 30.0$_{\pm 3.5}$ & 41.1$_{\pm 1.6}$ & 36.8$_{\pm 2.2}$ & 31.8$_{\pm 2.0}$ & 40.6$_{\pm 1.8}$ & 38.1$_{\pm 1.5}$ & 34.3$_{\pm 1.4}$ & 37.0$_{\pm 1.8}$ & 52.2$_{\pm 1.6}$ & 42.9$_{\pm 4.8}$ & 42.1$_{\pm 5.1}$ & 42.6$_{\pm 5.1}$ & 39.1 \\
  SimQINCo & 31.4$_{\pm 3.8}$ & 42.1$_{\pm 1.4}$ & 37.7$_{\pm 1.9}$ & 32.4$_{\pm 2.1}$ & 41.1$_{\pm 1.5}$ & 39.2$_{\pm 1.3}$ & 35.7$_{\pm 1.4}$ & 37.8$_{\pm 1.6}$ & 52.6$_{\pm 1.6}$ & 44.0$_{\pm 4.2}$ & 43.1$_{\pm 4.5}$ & 43.6$_{\pm 4.5}$ & 40.1 \\
  RQVAE & \textbf{31.9$_{\pm 3.5}$} & \textbf{43.1$_{\pm 1.6}$} & \textbf{39.1$_{\pm 2.2}$} & \textbf{33.4$_{\pm 2.1}$} & \textbf{42.5$_{\pm 1.6}$} & 39.0$_{\pm 1.4}$ & \textbf{37.0$_{\pm 1.4}$} & \textbf{38.2$_{\pm 1.4}$} & \textbf{55.3$_{\pm 0.9}$} & \textbf{45.8$_{\pm 4.2}$} & \textbf{45.2$_{\pm 4.8}$} & \textbf{45.5$_{\pm 4.7}$} & 41.3 \\
  DiVeQ & 29.9$_{\pm 3.9}$ & 40.5$_{\pm 1.6}$ & 37.1$_{\pm 2.2}$ & 31.4$_{\pm 2.0}$ & 40.0$_{\pm 1.9}$ & 39.2$_{\pm 1.3}$ & 34.0$_{\pm 1.3}$ & 37.4$_{\pm 1.9}$ & 52.9$_{\pm 1.6}$ & 43.0$_{\pm 5.3}$ & 42.3$_{\pm 5.6}$ & 42.9$_{\pm 5.5}$ & 39.2 \\
  SF-DiVeQ & 29.8$_{\pm 3.5}$ & 40.4$_{\pm 1.6}$ & 36.2$_{\pm 2.1}$ & 31.6$_{\pm 2.0}$ & 39.9$_{\pm 1.8}$ & 36.6$_{\pm 1.9}$ & 33.8$_{\pm 1.4}$ & 37.1$_{\pm 1.8}$ & 52.9$_{\pm 1.9}$ & 42.6$_{\pm 5.0}$ & 41.9$_{\pm 5.4}$ & 42.5$_{\pm 5.3}$ & 38.8 \\
  BSQ & 24.7$_{\pm 3.0}$ & 31.4$_{\pm 1.5}$ & 29.4$_{\pm 2.2}$ & 27.8$_{\pm 1.5}$ & 31.5$_{\pm 1.7}$ & 29.8$_{\pm 2.1}$ & 27.2$_{\pm 1.3}$ & 30.6$_{\pm 1.8}$ & 42.6$_{\pm 2.6}$ & 30.4$_{\pm 3.2}$ & 30.1$_{\pm 3.5}$ & 30.5$_{\pm 3.5}$ & 30.5 \\
  LFQ & 24.4$_{\pm 3.1}$ & 30.6$_{\pm 1.6}$ & 27.7$_{\pm 2.7}$ & 26.9$_{\pm 1.6}$ & 30.5$_{\pm 1.8}$ & 27.4$_{\pm 2.4}$ & 26.2$_{\pm 1.2}$ & 30.5$_{\pm 2.1}$ & 40.5$_{\pm 3.9}$ & 29.4$_{\pm 2.5}$ & 29.0$_{\pm 2.9}$ & 29.3$_{\pm 2.8}$ & 29.4 \\
  \midrule
  \multicolumn{14}{l}{\textit{$f = 8$}} \\
  AEKL & 25.0$_{\pm 3.1}$ & 31.8$_{\pm 1.3}$ & 30.9$_{\pm 1.9}$ & 27.6$_{\pm 1.6}$ & 31.4$_{\pm 1.5}$ & 31.5$_{\pm 1.4}$ & 26.9$_{\pm 1.1}$ & 31.0$_{\pm 1.7}$ & 39.8$_{\pm 1.7}$ & 31.2$_{\pm 3.3}$ & 31.1$_{\pm 3.7}$ & 31.4$_{\pm 3.6}$ & 30.8 \\
  SoftVQ & 24.2$_{\pm 3.5}$ & 31.5$_{\pm 1.6}$ & 30.6$_{\pm 2.1}$ & 27.2$_{\pm 1.5}$ & 31.8$_{\pm 1.7}$ & 29.2$_{\pm 1.8}$ & 26.1$_{\pm 1.4}$ & 30.8$_{\pm 2.1}$ & 41.3$_{\pm 2.5}$ & 29.9$_{\pm 3.1}$ & 29.9$_{\pm 3.7}$ & 30.2$_{\pm 3.5}$ & 30.2 \\
  VQGAN & 24.8$_{\pm 3.9}$ & 35.4$_{\pm 1.9}$ & 33.3$_{\pm 2.2}$ & 28.4$_{\pm 1.8}$ & 34.9$_{\pm 2.1}$ & 35.0$_{\pm 1.5}$ & 27.9$_{\pm 1.6}$ & 32.7$_{\pm 2.6}$ & 51.0$_{\pm 1.4}$ & 34.9$_{\pm 6.4}$ & 34.8$_{\pm 6.9}$ & 35.5$_{\pm 6.9}$ & 34.0 \\
  SimVQ & 26.6$_{\pm 3.9}$ & 36.0$_{\pm 1.5}$ & 34.8$_{\pm 2.3}$ & 29.4$_{\pm 1.9}$ & 35.4$_{\pm 1.8}$ & 36.2$_{\pm 1.6}$ & 29.7$_{\pm 1.4}$ & 34.4$_{\pm 2.3}$ & 49.0$_{\pm 2.4}$ & 35.6$_{\pm 4.6}$ & 35.3$_{\pm 5.0}$ & 35.8$_{\pm 5.0}$ & 34.8 \\
  SimQINCo & 26.2$_{\pm 3.9}$ & 36.6$_{\pm 1.7}$ & 34.7$_{\pm 2.4}$ & 29.0$_{\pm 1.8}$ & 36.2$_{\pm 2.1}$ & 35.0$_{\pm 1.4}$ & 29.5$_{\pm 1.5}$ & 34.3$_{\pm 2.4}$ & 48.5$_{\pm 1.2}$ & 36.6$_{\pm 5.6}$ & 36.6$_{\pm 6.4}$ & 37.2$_{\pm 6.3}$ & 35.0 \\
  RQVAE & \textbf{27.6$_{\pm 3.9}$} & \textbf{38.0$_{\pm 1.7}$} & \textbf{35.8$_{\pm 2.3}$} & \textbf{29.9$_{\pm 1.9}$} & \textbf{37.5$_{\pm 2.0}$} & 36.6$_{\pm 1.4}$ & \textbf{30.8$_{\pm 1.5}$} & \textbf{35.7$_{\pm 2.1}$} & \textbf{53.8$_{\pm 0.8}$} & \textbf{40.1$_{\pm 6.5}$} & \textbf{39.1$_{\pm 6.6}$} & \textbf{39.8$_{\pm 6.6}$} & 37.1 \\
  DiVeQ & 26.4$_{\pm 4.0}$ & 36.6$_{\pm 1.7}$ & 34.5$_{\pm 2.2}$ & 29.3$_{\pm 1.9}$ & 36.1$_{\pm 2.0}$ & 35.9$_{\pm 1.6}$ & 29.4$_{\pm 1.5}$ & 34.3$_{\pm 2.4}$ & 50.8$_{\pm 1.9}$ & 37.0$_{\pm 5.8}$ & 36.5$_{\pm 6.2}$ & 37.2$_{\pm 6.2}$ & 35.3 \\
  SF-DiVeQ & 27.2$_{\pm 4.2}$ & 37.5$_{\pm 1.8}$ & 35.5$_{\pm 2.4}$ & 29.6$_{\pm 1.9}$ & 36.6$_{\pm 2.0}$ & \textbf{37.4$_{\pm 1.3}$} & 30.3$_{\pm 1.4}$ & 35.2$_{\pm 2.3}$ & 49.2$_{\pm 2.1}$ & 37.2$_{\pm 5.1}$ & 37.0$_{\pm 5.7}$ & 37.5$_{\pm 5.6}$ & 35.8 \\
  BSQ & 22.5$_{\pm 3.0}$ & 29.0$_{\pm 1.4}$ & 26.7$_{\pm 2.2}$ & 26.5$_{\pm 1.4}$ & 28.6$_{\pm 1.6}$ & 27.1$_{\pm 2.4}$ & 24.7$_{\pm 1.4}$ & 28.0$_{\pm 1.9}$ & 38.0$_{\pm 3.0}$ & 27.0$_{\pm 2.6}$ & 27.0$_{\pm 3.1}$ & 27.3$_{\pm 3.0}$ & 27.7 \\
  LFQ & 20.0$_{\pm 2.9}$ & 25.7$_{\pm 2.0}$ & 24.2$_{\pm 3.0}$ & 24.4$_{\pm 1.6}$ & 24.8$_{\pm 1.9}$ & 24.8$_{\pm 2.2}$ & 20.2$_{\pm 2.1}$ & 24.2$_{\pm 2.3}$ & 33.2$_{\pm 3.9}$ & 22.2$_{\pm 2.5}$ & 22.0$_{\pm 2.9}$ & 22.2$_{\pm 3.0}$ & 24.0 \\
  \midrule
  \multicolumn{14}{l}{\textit{$f = 16$}} \\
  AEKL & 24.3$_{\pm 3.6}$ & 32.0$_{\pm 1.4}$ & 31.2$_{\pm 1.9}$ & 27.4$_{\pm 1.6}$ & 31.4$_{\pm 1.6}$ & 30.0$_{\pm 1.5}$ & 26.5$_{\pm 1.3}$ & 30.9$_{\pm 2.3}$ & 40.5$_{\pm 1.6}$ & 30.8$_{\pm 3.5}$ & 30.8$_{\pm 4.1}$ & 31.2$_{\pm 4.0}$ & 30.6 \\
  SoftVQ & 24.5$_{\pm 4.0}$ & 34.1$_{\pm 1.6}$ & 32.6$_{\pm 2.4}$ & 28.2$_{\pm 1.7}$ & 33.6$_{\pm 2.0}$ & 33.3$_{\pm 1.6}$ & 26.3$_{\pm 1.5}$ & 32.1$_{\pm 2.6}$ & 44.6$_{\pm 2.3}$ & 32.3$_{\pm 4.7}$ & 32.1$_{\pm 5.3}$ & 32.7$_{\pm 5.3}$ & 32.2 \\
  VQGAN & 22.4$_{\pm 4.3}$ & 31.2$_{\pm 1.8}$ & 31.2$_{\pm 2.5}$ & 27.5$_{\pm 1.6}$ & 30.6$_{\pm 2.0}$ & 30.2$_{\pm 1.6}$ & 24.5$_{\pm 1.7}$ & 28.9$_{\pm 2.7}$ & 41.2$_{\pm 2.6}$ & 27.5$_{\pm 3.5}$ & 27.8$_{\pm 4.5}$ & 28.1$_{\pm 4.5}$ & 29.3 \\
  SimVQ & 22.7$_{\pm 4.0}$ & 30.2$_{\pm 1.6}$ & 30.5$_{\pm 2.3}$ & 27.0$_{\pm 1.5}$ & 29.6$_{\pm 1.7}$ & 30.1$_{\pm 1.3}$ & 24.4$_{\pm 1.5}$ & 29.0$_{\pm 2.6}$ & 39.0$_{\pm 2.6}$ & 27.3$_{\pm 3.2}$ & 27.6$_{\pm 4.1}$ & 27.9$_{\pm 4.0}$ & 28.8 \\
  SimQINCo & 23.6$_{\pm 4.5}$ & 33.2$_{\pm 1.8}$ & 32.9$_{\pm 2.6}$ & 28.0$_{\pm 1.7}$ & 32.6$_{\pm 2.1}$ & 34.1$_{\pm 1.3}$ & 25.8$_{\pm 1.6}$ & 31.0$_{\pm 3.0}$ & 46.1$_{\pm 2.4}$ & 30.3$_{\pm 4.4}$ & 30.7$_{\pm 5.5}$ & 31.1$_{\pm 5.4}$ & 31.6 \\
  RQVAE & \textbf{25.2$_{\pm 4.3}$} & \textbf{35.5$_{\pm 1.9}$} & \textbf{33.9$_{\pm 2.5}$} & \textbf{28.6$_{\pm 1.9}$} & \textbf{34.6$_{\pm 2.1}$} & \textbf{35.0$_{\pm 1.4}$} & \textbf{27.6$_{\pm 1.6}$} & \textbf{33.0$_{\pm 2.7}$} & \textbf{52.6$_{\pm 1.5}$} & \textbf{34.7$_{\pm 6.3}$} & \textbf{34.5$_{\pm 6.9}$} & \textbf{35.2$_{\pm 6.8}$} & 34.2 \\
  DiVeQ & 23.5$_{\pm 4.4}$ & 32.7$_{\pm 1.7}$ & 31.9$_{\pm 2.3}$ & 28.0$_{\pm 1.7}$ & 32.0$_{\pm 2.1}$ & 30.8$_{\pm 1.0}$ & 25.5$_{\pm 1.6}$ & 30.8$_{\pm 2.9}$ & 45.5$_{\pm 2.6}$ & 29.9$_{\pm 4.2}$ & 30.2$_{\pm 5.3}$ & 30.6$_{\pm 5.2}$ & 31.0 \\
  SF-DiVeQ & 23.5$_{\pm 4.3}$ & 33.0$_{\pm 1.8}$ & 32.4$_{\pm 2.4}$ & 28.0$_{\pm 1.7}$ & 32.4$_{\pm 2.0}$ & 33.8$_{\pm 1.4}$ & 25.7$_{\pm 1.6}$ & 30.8$_{\pm 3.0}$ & 44.6$_{\pm 2.5}$ & 30.1$_{\pm 4.2}$ & 30.4$_{\pm 5.4}$ & 30.8$_{\pm 5.2}$ & 31.3 \\
  BSQ & 22.1$_{\pm 3.4}$ & 29.1$_{\pm 1.7}$ & 28.4$_{\pm 2.4}$ & 26.0$_{\pm 1.5}$ & 28.9$_{\pm 1.9}$ & 28.2$_{\pm 2.3}$ & 24.0$_{\pm 1.3}$ & 27.9$_{\pm 2.2}$ & 38.2$_{\pm 3.0}$ & 26.3$_{\pm 2.9}$ & 26.5$_{\pm 3.7}$ & 26.7$_{\pm 3.5}$ & 27.7 \\
  LFQ & 19.1$_{\pm 3.6}$ & 24.6$_{\pm 1.9}$ & 23.8$_{\pm 2.6}$ & 22.4$_{\pm 1.6}$ & 24.5$_{\pm 1.9}$ & 24.2$_{\pm 2.1}$ & 20.2$_{\pm 1.5}$ & 23.6$_{\pm 2.0}$ & 32.0$_{\pm 3.2}$ & 21.2$_{\pm 2.4}$ & 21.1$_{\pm 2.9}$ & 21.4$_{\pm 2.8}$ & 23.2 \\
  \bottomrule
  \end{tabular}}
\end{table*}

\begin{table*}[h]
  \centering
  \caption{Per-dataset SSIM ($\uparrow$). Cells: $\text{mean}_{\pm\sigma}$ over test images.}
  \label{tab:ssim}
  \setlength{\tabcolsep}{3pt}
  \adjustbox{max width=\textwidth,max totalheight=0.43\textheight,center}{%
  \begin{tabular}{lcccccccccccc|c}
  \toprule
  Method & Path & Chest & Derma & OCT & Pneu & Reti & Breast & Blood & Tiss & OrgA & OrgC & OrgS & mean \\
  \midrule
  \multicolumn{14}{l}{\textit{$f = 4$}} \\
  AEKL & .714$_{\pm .119}$ & .883$_{\pm .027}$ & .838$_{\pm .057}$ & .578$_{\pm .088}$ & .854$_{\pm .031}$ & .887$_{\pm .024}$ & .707$_{\pm .027}$ & .891$_{\pm .035}$ & .965$_{\pm .009}$ & .856$_{\pm .091}$ & .848$_{\pm .102}$ & .857$_{\pm .100}$ & .823 \\
  SoftVQ & .846$_{\pm .068}$ & .953$_{\pm .011}$ & .907$_{\pm .031}$ & .700$_{\pm .066}$ & .939$_{\pm .012}$ & .938$_{\pm .027}$ & .852$_{\pm .021}$ & .935$_{\pm .017}$ & .993$_{\pm .003}$ & .961$_{\pm .039}$ & .954$_{\pm .047}$ & .957$_{\pm .044}$ & .911 \\
  VQGAN & .906$_{\pm .049}$ & .981$_{\pm .004}$ & .943$_{\pm .019}$ & .832$_{\pm .038}$ & .973$_{\pm .006}$ & .960$_{\pm .021}$ & .941$_{\pm .008}$ & .954$_{\pm .015}$ & .997$_{\pm .000}$ & .986$_{\pm .016}$ & .983$_{\pm .024}$ & .985$_{\pm .021}$ & .953 \\
  SimVQ & .884$_{\pm .059}$ & .977$_{\pm .006}$ & .937$_{\pm .022}$ & .808$_{\pm .045}$ & .968$_{\pm .007}$ & .961$_{\pm .013}$ & .926$_{\pm .010}$ & .950$_{\pm .016}$ & .997$_{\pm .001}$ & .984$_{\pm .023}$ & .979$_{\pm .031}$ & .980$_{\pm .028}$ & .946 \\
  SimQINCo & .905$_{\pm .055}$ & .983$_{\pm .004}$ & .946$_{\pm .019}$ & .828$_{\pm .040}$ & .975$_{\pm .006}$ & .949$_{\pm .019}$ & .941$_{\pm .008}$ & .955$_{\pm .015}$ & .997$_{\pm .001}$ & .987$_{\pm .016}$ & .984$_{\pm .025}$ & .985$_{\pm .022}$ & .953 \\
  RQVAE & \textbf{.921$_{\pm .044}$} & \textbf{.986$_{\pm .003}$} & \textbf{.951$_{\pm .017}$} & \textbf{.865$_{\pm .033}$} & \textbf{.980$_{\pm .004}$} & .962$_{\pm .015}$ & \textbf{.958$_{\pm .007}$} & \textbf{.959$_{\pm .014}$} & \textbf{.998$_{\pm .000}$} & \textbf{.992$_{\pm .010}$} & \textbf{.989$_{\pm .017}$} & \textbf{.991$_{\pm .015}$} & .963 \\
  DiVeQ & .869$_{\pm .068}$ & .975$_{\pm .007}$ & .932$_{\pm .024}$ & .785$_{\pm .049}$ & .964$_{\pm .008}$ & \textbf{.966$_{\pm .011}$} & .916$_{\pm .011}$ & .949$_{\pm .017}$ & .997$_{\pm .000}$ & .980$_{\pm .029}$ & .976$_{\pm .037}$ & .978$_{\pm .034}$ & .940 \\
  SF-DiVeQ & .879$_{\pm .058}$ & .977$_{\pm .006}$ & .935$_{\pm .022}$ & .798$_{\pm .046}$ & .966$_{\pm .007}$ & .965$_{\pm .019}$ & .918$_{\pm .011}$ & .949$_{\pm .016}$ & .998$_{\pm .000}$ & .980$_{\pm .025}$ & .976$_{\pm .032}$ & .978$_{\pm .030}$ & .943 \\
  BSQ & .743$_{\pm .100}$ & .912$_{\pm .021}$ & .873$_{\pm .043}$ & .670$_{\pm .067}$ & .893$_{\pm .025}$ & .882$_{\pm .034}$ & .771$_{\pm .021}$ & .901$_{\pm .032}$ & .985$_{\pm .009}$ & .903$_{\pm .069}$ & .888$_{\pm .079}$ & .895$_{\pm .077}$ & .860 \\
  LFQ & .755$_{\pm .096}$ & .903$_{\pm .024}$ & .866$_{\pm .043}$ & .666$_{\pm .063}$ & .887$_{\pm .026}$ & .900$_{\pm .033}$ & .784$_{\pm .023}$ & .903$_{\pm .030}$ & .982$_{\pm .008}$ & .895$_{\pm .061}$ & .887$_{\pm .074}$ & .893$_{\pm .071}$ & .860 \\
  \midrule
  \multicolumn{14}{l}{\textit{$f = 8$}} \\
  AEKL & .679$_{\pm .140}$ & .892$_{\pm .026}$ & .837$_{\pm .057}$ & .576$_{\pm .089}$ & .861$_{\pm .030}$ & .913$_{\pm .023}$ & .701$_{\pm .029}$ & .892$_{\pm .036}$ & .972$_{\pm .008}$ & .867$_{\pm .100}$ & .856$_{\pm .110}$ & .867$_{\pm .107}$ & .826 \\
  SoftVQ & .636$_{\pm .159}$ & .894$_{\pm .027}$ & .854$_{\pm .055}$ & .582$_{\pm .086}$ & .875$_{\pm .031}$ & .902$_{\pm .021}$ & .667$_{\pm .039}$ & .887$_{\pm .040}$ & .982$_{\pm .007}$ & .850$_{\pm .106}$ & .836$_{\pm .119}$ & .842$_{\pm .119}$ & .817 \\
  VQGAN & .644$_{\pm .163}$ & .929$_{\pm .020}$ & .866$_{\pm .053}$ & .601$_{\pm .087}$ & .898$_{\pm .026}$ & .942$_{\pm .017}$ & .727$_{\pm .037}$ & .897$_{\pm .041}$ & .996$_{\pm .001}$ & .895$_{\pm .103}$ & .879$_{\pm .117}$ & .889$_{\pm .115}$ & .847 \\
  SimVQ & .741$_{\pm .122}$ & .945$_{\pm .014}$ & .890$_{\pm .042}$ & .672$_{\pm .073}$ & .922$_{\pm .020}$ & .947$_{\pm .019}$ & .811$_{\pm .025}$ & .922$_{\pm .027}$ & .996$_{\pm .001}$ & .930$_{\pm .078}$ & .920$_{\pm .085}$ & .926$_{\pm .081}$ & .885 \\
  SimQINCo & .713$_{\pm .137}$ & .946$_{\pm .015}$ & .889$_{\pm .043}$ & .644$_{\pm .079}$ & .923$_{\pm .020}$ & \textbf{.951$_{\pm .013}$} & .794$_{\pm .026}$ & .919$_{\pm .030}$ & .994$_{\pm .003}$ & .920$_{\pm .085}$ & .912$_{\pm .094}$ & .919$_{\pm .091}$ & .877 \\
  RQVAE & \textbf{.791$_{\pm .099}$} & \textbf{.959$_{\pm .011}$} & \textbf{.907$_{\pm .034}$} & \textbf{.709$_{\pm .066}$} & \textbf{.941$_{\pm .014}$} & .939$_{\pm .016}$ & \textbf{.849$_{\pm .020}$} & \textbf{.934$_{\pm .022}$} & \textbf{.997$_{\pm .000}$} & \textbf{.958$_{\pm .057}$} & \textbf{.948$_{\pm .063}$} & \textbf{.953$_{\pm .059}$} & .907 \\
  DiVeQ & .730$_{\pm .127}$ & .946$_{\pm .015}$ & .890$_{\pm .042}$ & .663$_{\pm .077}$ & .924$_{\pm .019}$ & .947$_{\pm .018}$ & .801$_{\pm .027}$ & .921$_{\pm .028}$ & .997$_{\pm .001}$ & .934$_{\pm .077}$ & .923$_{\pm .085}$ & .930$_{\pm .081}$ & .884 \\
  SF-DiVeQ & .759$_{\pm .115}$ & .953$_{\pm .013}$ & .897$_{\pm .040}$ & .684$_{\pm .070}$ & .932$_{\pm .018}$ & .921$_{\pm .021}$ & .825$_{\pm .022}$ & .929$_{\pm .026}$ & .997$_{\pm .000}$ & .942$_{\pm .073}$ & .932$_{\pm .079}$ & .938$_{\pm .076}$ & .892 \\
  BSQ & .629$_{\pm .147}$ & .877$_{\pm .030}$ & .838$_{\pm .058}$ & .594$_{\pm .081}$ & .851$_{\pm .035}$ & .871$_{\pm .038}$ & .685$_{\pm .032}$ & .866$_{\pm .046}$ & .965$_{\pm .015}$ & .844$_{\pm .094}$ & .833$_{\pm .109}$ & .841$_{\pm .106}$ & .808 \\
  LFQ & .502$_{\pm .190}$ & .806$_{\pm .047}$ & .795$_{\pm .082}$ & .540$_{\pm .086}$ & .745$_{\pm .061}$ & .798$_{\pm .054}$ & .499$_{\pm .061}$ & .802$_{\pm .071}$ & .935$_{\pm .028}$ & .699$_{\pm .125}$ & .685$_{\pm .151}$ & .694$_{\pm .148}$ & .708 \\
  \midrule
  \multicolumn{14}{l}{\textit{$f = 16$}} \\
  AEKL & .615$_{\pm .167}$ & .889$_{\pm .027}$ & .826$_{\pm .061}$ & .571$_{\pm .092}$ & .851$_{\pm .034}$ & .901$_{\pm .027}$ & .666$_{\pm .038}$ & .878$_{\pm .046}$ & .974$_{\pm .006}$ & .850$_{\pm .113}$ & .837$_{\pm .128}$ & .846$_{\pm .125}$ & .809 \\
  SoftVQ & .612$_{\pm .176}$ & .912$_{\pm .024}$ & .849$_{\pm .059}$ & .587$_{\pm .090}$ & .878$_{\pm .032}$ & .910$_{\pm .029}$ & .650$_{\pm .048}$ & .889$_{\pm .046}$ & .989$_{\pm .009}$ & .864$_{\pm .115}$ & .842$_{\pm .132}$ & .854$_{\pm .128}$ & .820 \\
  VQGAN & .490$_{\pm .214}$ & .873$_{\pm .035}$ & .817$_{\pm .075}$ & .572$_{\pm .092}$ & .828$_{\pm .045}$ & .853$_{\pm .037}$ & .566$_{\pm .068}$ & .840$_{\pm .068}$ & .977$_{\pm .006}$ & .754$_{\pm .133}$ & .742$_{\pm .162}$ & .755$_{\pm .159}$ & .756 \\
  SimVQ & .514$_{\pm .207}$ & .864$_{\pm .035}$ & .816$_{\pm .074}$ & .557$_{\pm .091}$ & .819$_{\pm .043}$ & .890$_{\pm .025}$ & .573$_{\pm .063}$ & .845$_{\pm .065}$ & .970$_{\pm .008}$ & .769$_{\pm .131}$ & .758$_{\pm .155}$ & .769$_{\pm .151}$ & .762 \\
  SimQINCo & .553$_{\pm .196}$ & .901$_{\pm .029}$ & .838$_{\pm .068}$ & .581$_{\pm .091}$ & .861$_{\pm .037}$ & .915$_{\pm .025}$ & .621$_{\pm .058}$ & .871$_{\pm .056}$ & .990$_{\pm .002}$ & .816$_{\pm .125}$ & .802$_{\pm .149}$ & .812$_{\pm .146}$ & .797 \\
  RQVAE & \textbf{.647$_{\pm .160}$} & \textbf{.926$_{\pm .022}$} & \textbf{.862$_{\pm .055}$} & \textbf{.603$_{\pm .089}$} & \textbf{.892$_{\pm .030}$} & \textbf{.933$_{\pm .017}$} & \textbf{.709$_{\pm .041}$} & \textbf{.901$_{\pm .042}$} & \textbf{.997$_{\pm .000}$} & \textbf{.885$_{\pm .109}$} & \textbf{.868$_{\pm .122}$} & \textbf{.878$_{\pm .119}$} & .842 \\
  DiVeQ & .551$_{\pm .197}$ & .899$_{\pm .029}$ & .835$_{\pm .068}$ & .583$_{\pm .092}$ & .856$_{\pm .039}$ & .907$_{\pm .028}$ & .612$_{\pm .058}$ & .868$_{\pm .057}$ & .990$_{\pm .002}$ & .817$_{\pm .126}$ & .802$_{\pm .148}$ & .812$_{\pm .145}$ & .794 \\
  SF-DiVeQ & .549$_{\pm .198}$ & .898$_{\pm .029}$ & .834$_{\pm .068}$ & .581$_{\pm .092}$ & .859$_{\pm .038}$ & .900$_{\pm .022}$ & .621$_{\pm .058}$ & .868$_{\pm .058}$ & .990$_{\pm .002}$ & .813$_{\pm .125}$ & .800$_{\pm .150}$ & .810$_{\pm .146}$ & .793 \\
  BSQ & .543$_{\pm .189}$ & .864$_{\pm .034}$ & .827$_{\pm .067}$ & .566$_{\pm .087}$ & .830$_{\pm .044}$ & .881$_{\pm .026}$ & .608$_{\pm .050}$ & .849$_{\pm .057}$ & .972$_{\pm .009}$ & .784$_{\pm .125}$ & .771$_{\pm .147}$ & .784$_{\pm .143}$ & .773 \\
  LFQ & .402$_{\pm .228}$ & .766$_{\pm .059}$ & .777$_{\pm .089}$ & .499$_{\pm .087}$ & .706$_{\pm .074}$ & .775$_{\pm .045}$ & .454$_{\pm .068}$ & .766$_{\pm .082}$ & .910$_{\pm .033}$ & .600$_{\pm .155}$ & .599$_{\pm .183}$ & .608$_{\pm .181}$ & .655 \\
  \bottomrule
  \end{tabular}}
\end{table*}

\begin{table*}[h]
  \centering
  \caption{Per-dataset LPIPS ($\downarrow$). Cells: $\text{mean}_{\pm\sigma}$ over test images.}
  \label{tab:lpips}
  \setlength{\tabcolsep}{3pt}
  \adjustbox{max width=\textwidth,max totalheight=0.43\textheight,center}{%
  \begin{tabular}{lcccccccccccc|c}
  \toprule
  Method & Path & Chest & Derma & OCT & Pneu & Reti & Breast & Blood & Tiss & OrgA & OrgC & OrgS & mean \\
  \midrule
  \multicolumn{14}{l}{\textit{$f = 4$}} \\
  AEKL & .205$_{\pm .071}$ & .113$_{\pm .022}$ & .207$_{\pm .056}$ & .240$_{\pm .044}$ & .141$_{\pm .029}$ & .105$_{\pm .025}$ & .241$_{\pm .031}$ & .068$_{\pm .018}$ & .040$_{\pm .010}$ & .110$_{\pm .058}$ & .106$_{\pm .061}$ & .102$_{\pm .057}$ & .140 \\
  SoftVQ & .147$_{\pm .065}$ & .051$_{\pm .011}$ & .123$_{\pm .040}$ & .222$_{\pm .055}$ & .073$_{\pm .023}$ & .068$_{\pm .029}$ & .184$_{\pm .032}$ & .033$_{\pm .007}$ & .005$_{\pm .001}$ & .046$_{\pm .049}$ & .048$_{\pm .054}$ & .043$_{\pm .049}$ & .087 \\
  VQGAN & .076$_{\pm .043}$ & .017$_{\pm .004}$ & .051$_{\pm .018}$ & .096$_{\pm .027}$ & .029$_{\pm .009}$ & .014$_{\pm .006}$ & .082$_{\pm .015}$ & .013$_{\pm .003}$ & .001$_{\pm .000}$ & .013$_{\pm .022}$ & .013$_{\pm .024}$ & .011$_{\pm .018}$ & .035 \\
  SimVQ & .097$_{\pm .052}$ & .019$_{\pm .005}$ & .064$_{\pm .024}$ & .104$_{\pm .027}$ & .032$_{\pm .008}$ & .022$_{\pm .008}$ & .100$_{\pm .017}$ & .017$_{\pm .004}$ & .001$_{\pm .000}$ & .017$_{\pm .029}$ & .018$_{\pm .031}$ & .015$_{\pm .025}$ & .042 \\
  SimQINCo & .072$_{\pm .041}$ & .015$_{\pm .004}$ & .050$_{\pm .018}$ & .094$_{\pm .026}$ & .027$_{\pm .008}$ & .015$_{\pm .007}$ & .068$_{\pm .013}$ & .012$_{\pm .003}$ & .002$_{\pm .000}$ & .011$_{\pm .020}$ & .012$_{\pm .022}$ & .010$_{\pm .017}$ & .032 \\
  RQVAE & \textbf{.059$_{\pm .032}$} & \textbf{.011$_{\pm .003}$} & \textbf{.040$_{\pm .015}$} & \textbf{.068$_{\pm .019}$} & \textbf{.019$_{\pm .005}$} & \textbf{.013$_{\pm .004}$} & \textbf{.049$_{\pm .010}$} & \textbf{.010$_{\pm .002}$} & \textbf{.001$_{\pm .000}$} & \textbf{.008$_{\pm .013}$} & \textbf{.009$_{\pm .016}$} & \textbf{.007$_{\pm .012}$} & .024 \\
  DiVeQ & .100$_{\pm .055}$ & .023$_{\pm .006}$ & .066$_{\pm .023}$ & .125$_{\pm .030}$ & .039$_{\pm .012}$ & .020$_{\pm .006}$ & .096$_{\pm .018}$ & .017$_{\pm .004}$ & .002$_{\pm .001}$ & .017$_{\pm .032}$ & .018$_{\pm .032}$ & .015$_{\pm .027}$ & .045 \\
  SF-DiVeQ & .098$_{\pm .052}$ & .020$_{\pm .005}$ & .078$_{\pm .025}$ & .116$_{\pm .029}$ & .032$_{\pm .009}$ & .028$_{\pm .012}$ & .091$_{\pm .017}$ & .018$_{\pm .004}$ & .002$_{\pm .000}$ & .016$_{\pm .027}$ & .018$_{\pm .030}$ & .016$_{\pm .025}$ & .044 \\
  BSQ & .195$_{\pm .070}$ & .083$_{\pm .016}$ & .166$_{\pm .049}$ & .203$_{\pm .043}$ & .104$_{\pm .019}$ & .095$_{\pm .027}$ & .206$_{\pm .034}$ & .060$_{\pm .014}$ & .014$_{\pm .004}$ & .083$_{\pm .050}$ & .085$_{\pm .053}$ & .080$_{\pm .049}$ & .114 \\
  LFQ & .211$_{\pm .081}$ & .088$_{\pm .016}$ & .179$_{\pm .058}$ & .157$_{\pm .035}$ & .110$_{\pm .019}$ & .118$_{\pm .028}$ & .210$_{\pm .035}$ & .061$_{\pm .022}$ & .019$_{\pm .006}$ & .092$_{\pm .049}$ & .093$_{\pm .061}$ & .090$_{\pm .057}$ & .119 \\
  \midrule
  \multicolumn{14}{l}{\textit{$f = 8$}} \\
  AEKL & .219$_{\pm .083}$ & .094$_{\pm .020}$ & .163$_{\pm .051}$ & .232$_{\pm .050}$ & .113$_{\pm .020}$ & .082$_{\pm .025}$ & .219$_{\pm .028}$ & .065$_{\pm .017}$ & .027$_{\pm .007}$ & .104$_{\pm .064}$ & .100$_{\pm .066}$ & .095$_{\pm .064}$ & .126 \\
  SoftVQ & .239$_{\pm .082}$ & .100$_{\pm .023}$ & .175$_{\pm .054}$ & .195$_{\pm .038}$ & .116$_{\pm .033}$ & .114$_{\pm .027}$ & .257$_{\pm .044}$ & .070$_{\pm .019}$ & .013$_{\pm .003}$ & .111$_{\pm .068}$ & .119$_{\pm .083}$ & .112$_{\pm .081}$ & .135 \\
  VQGAN & .230$_{\pm .085}$ & .077$_{\pm .018}$ & .158$_{\pm .046}$ & .185$_{\pm .038}$ & .097$_{\pm .022}$ & .069$_{\pm .018}$ & .227$_{\pm .035}$ & .059$_{\pm .019}$ & .002$_{\pm .000}$ & .088$_{\pm .073}$ & .091$_{\pm .080}$ & .083$_{\pm .077}$ & .114 \\
  SimVQ & .174$_{\pm .083}$ & .049$_{\pm .010}$ & .100$_{\pm .032}$ & .145$_{\pm .029}$ & .073$_{\pm .017}$ & .038$_{\pm .015}$ & \textbf{.170$_{\pm .028}$} & .035$_{\pm .010}$ & .003$_{\pm .001}$ & .055$_{\pm .056}$ & .054$_{\pm .056}$ & .049$_{\pm .052}$ & .079 \\
  SimQINCo & .199$_{\pm .091}$ & .056$_{\pm .012}$ & .115$_{\pm .035}$ & .184$_{\pm .041}$ & .078$_{\pm .019}$ & .048$_{\pm .014}$ & .204$_{\pm .031}$ & .041$_{\pm .012}$ & .003$_{\pm .000}$ & .061$_{\pm .065}$ & .062$_{\pm .065}$ & .057$_{\pm .062}$ & .092 \\
  RQVAE & \textbf{.153$_{\pm .076}$} & \textbf{.038$_{\pm .009}$} & \textbf{.088$_{\pm .029}$} & .151$_{\pm .035}$ & \textbf{.053$_{\pm .013}$} & .032$_{\pm .009}$ & .173$_{\pm .030}$ & \textbf{.026$_{\pm .006}$} & \textbf{.001$_{\pm .000}$} & \textbf{.035$_{\pm .048}$} & \textbf{.035$_{\pm .048}$} & \textbf{.031$_{\pm .043}$} & .068 \\
  DiVeQ & .189$_{\pm .088}$ & .055$_{\pm .012}$ & .105$_{\pm .034}$ & .164$_{\pm .030}$ & .077$_{\pm .018}$ & .041$_{\pm .013}$ & .208$_{\pm .033}$ & .038$_{\pm .010}$ & .003$_{\pm .001}$ & .055$_{\pm .061}$ & .056$_{\pm .061}$ & .051$_{\pm .057}$ & .087 \\
  SF-DiVeQ & .172$_{\pm .086}$ & .043$_{\pm .010}$ & .091$_{\pm .029}$ & \textbf{.125$_{\pm .026}$} & .064$_{\pm .016}$ & \textbf{.032$_{\pm .011}$} & .184$_{\pm .030}$ & .031$_{\pm .008}$ & .003$_{\pm .001}$ & .046$_{\pm .058}$ & .046$_{\pm .056}$ & .041$_{\pm .052}$ & .073 \\
  BSQ & .269$_{\pm .086}$ & .118$_{\pm .024}$ & .210$_{\pm .057}$ & .179$_{\pm .031}$ & .139$_{\pm .026}$ & .137$_{\pm .033}$ & .255$_{\pm .037}$ & .087$_{\pm .023}$ & .031$_{\pm .008}$ & .136$_{\pm .063}$ & .140$_{\pm .076}$ & .135$_{\pm .074}$ & .153 \\
  LFQ & .344$_{\pm .086}$ & .204$_{\pm .040}$ & .320$_{\pm .100}$ & .279$_{\pm .046}$ & .237$_{\pm .039}$ & .210$_{\pm .042}$ & .435$_{\pm .056}$ & .158$_{\pm .043}$ & .089$_{\pm .034}$ & .257$_{\pm .083}$ & .275$_{\pm .100}$ & .270$_{\pm .093}$ & .257 \\
  \midrule
  \multicolumn{14}{l}{\textit{$f = 16$}} \\
  AEKL & .241$_{\pm .090}$ & .099$_{\pm .021}$ & .160$_{\pm .048}$ & .240$_{\pm .046}$ & .122$_{\pm .022}$ & .082$_{\pm .021}$ & .244$_{\pm .033}$ & .074$_{\pm .023}$ & .028$_{\pm .008}$ & .122$_{\pm .078}$ & .121$_{\pm .083}$ & .114$_{\pm .080}$ & .137 \\
  SoftVQ & .227$_{\pm .092}$ & .073$_{\pm .017}$ & .132$_{\pm .047}$ & .214$_{\pm .047}$ & .089$_{\pm .020}$ & .068$_{\pm .021}$ & .243$_{\pm .038}$ & .060$_{\pm .024}$ & .007$_{\pm .002}$ & .110$_{\pm .078}$ & .109$_{\pm .083}$ & .103$_{\pm .080}$ & .120 \\
  VQGAN & .317$_{\pm .106}$ & .118$_{\pm .028}$ & .193$_{\pm .062}$ & .297$_{\pm .063}$ & .145$_{\pm .029}$ & .097$_{\pm .025}$ & .330$_{\pm .044}$ & .117$_{\pm .045}$ & .025$_{\pm .010}$ & .199$_{\pm .093}$ & .198$_{\pm .108}$ & .190$_{\pm .103}$ & .186 \\
  SimVQ & .298$_{\pm .099}$ & .131$_{\pm .031}$ & .200$_{\pm .067}$ & .247$_{\pm .050}$ & .156$_{\pm .027}$ & .106$_{\pm .029}$ & .320$_{\pm .043}$ & .104$_{\pm .040}$ & .041$_{\pm .015}$ & .194$_{\pm .088}$ & .188$_{\pm .099}$ & .181$_{\pm .095}$ & .180 \\
  SimQINCo & .262$_{\pm .099}$ & .083$_{\pm .020}$ & .155$_{\pm .051}$ & .227$_{\pm .049}$ & .111$_{\pm .025}$ & .068$_{\pm .019}$ & .271$_{\pm .039}$ & .075$_{\pm .031}$ & .008$_{\pm .003}$ & .144$_{\pm .084}$ & .138$_{\pm .093}$ & .131$_{\pm .090}$ & .139 \\
  RQVAE & \textbf{.223$_{\pm .096}$} & \textbf{.060$_{\pm .014}$} & \textbf{.124$_{\pm .045}$} & \textbf{.179$_{\pm .036}$} & \textbf{.083$_{\pm .018}$} & \textbf{.050$_{\pm .014}$} & \textbf{.206$_{\pm .034}$} & \textbf{.055$_{\pm .022}$} & \textbf{.002$_{\pm .000}$} & \textbf{.091$_{\pm .076}$} & \textbf{.091$_{\pm .078}$} & \textbf{.084$_{\pm .076}$} & .104 \\
  DiVeQ & .266$_{\pm .101}$ & .090$_{\pm .022}$ & .154$_{\pm .052}$ & .241$_{\pm .048}$ & .116$_{\pm .024}$ & .082$_{\pm .020}$ & .274$_{\pm .036}$ & .083$_{\pm .032}$ & .008$_{\pm .004}$ & .146$_{\pm .082}$ & .142$_{\pm .092}$ & .135$_{\pm .089}$ & .145 \\
  SF-DiVeQ & .263$_{\pm .096}$ & .086$_{\pm .021}$ & .147$_{\pm .050}$ & .217$_{\pm .045}$ & .110$_{\pm .023}$ & .066$_{\pm .019}$ & .270$_{\pm .038}$ & .079$_{\pm .031}$ & .009$_{\pm .004}$ & .150$_{\pm .085}$ & .144$_{\pm .095}$ & .137$_{\pm .092}$ & .140 \\
  BSQ & .318$_{\pm .099}$ & .128$_{\pm .028}$ & .231$_{\pm .068}$ & .234$_{\pm .043}$ & .154$_{\pm .029}$ & .132$_{\pm .031}$ & .317$_{\pm .051}$ & .112$_{\pm .035}$ & .030$_{\pm .010}$ & .173$_{\pm .083}$ & .185$_{\pm .111}$ & .179$_{\pm .111}$ & .183 \\
  LFQ & .459$_{\pm .101}$ & .241$_{\pm .043}$ & .365$_{\pm .106}$ & .306$_{\pm .054}$ & .281$_{\pm .042}$ & .227$_{\pm .049}$ & .489$_{\pm .066}$ & .212$_{\pm .064}$ & .118$_{\pm .034}$ & .356$_{\pm .113}$ & .384$_{\pm .141}$ & .379$_{\pm .138}$ & .318 \\
  \bottomrule
  \end{tabular}}
\end{table*}

\begin{table*}[h]
  \centering
  \caption{Per-dataset pixel-MSE ($\downarrow$). Cells: $\text{mean}_{\pm\sigma}$ over test images.}
  \label{tab:mse}
  \setlength{\tabcolsep}{3pt}
  \adjustbox{max width=\textwidth,max totalheight=0.43\textheight,center}{%
  \begin{tabular}{lcccccccccccc|c}
  \toprule
  Method & Path & Chest & Derma & OCT & Pneu & Reti & Breast & Blood & Tiss & OrgA & OrgC & OrgS & mean \\
  \midrule
  \multicolumn{14}{l}{\textit{$f = 4$}} \\
  AEKL & .0037$_{\pm .0025}$ & 8.1e-4$_{\pm 2.5e-4}$ & .0011$_{\pm 6.4e-4}$ & .0019$_{\pm 7.8e-4}$ & 9.1e-4$_{\pm 3.1e-4}$ & 8.5e-4$_{\pm 3.7e-4}$ & .0021$_{\pm 5.1e-4}$ & 9.4e-4$_{\pm 4.1e-4}$ & 1.6e-4$_{\pm 8.1e-5}$ & .0011$_{\pm 7.1e-4}$ & .0012$_{\pm 8.5e-4}$ & .0011$_{\pm 8.2e-4}$ & .0013 \\
  SoftVQ & .0019$_{\pm .0013}$ & 2.5e-4$_{\pm 1.1e-4}$ & 7.4e-4$_{\pm 5.3e-4}$ & .0012$_{\pm 5.8e-4}$ & 2.8e-4$_{\pm 1.2e-4}$ & 9.0e-4$_{\pm 6.5e-4}$ & 9.1e-4$_{\pm 2.8e-4}$ & 4.4e-4$_{\pm 1.4e-4}$ & 2.3e-5$_{\pm 1.4e-5}$ & 3.5e-4$_{\pm 3.7e-4}$ & 4.1e-4$_{\pm 4.5e-4}$ & 3.7e-4$_{\pm 4.2e-4}$ & 6.5e-4 \\
  VQGAN & .0010$_{\pm 7.7e-4}$ & 7.6e-5$_{\pm 2.9e-5}$ & 2.2e-4$_{\pm 1.5e-4}$ & 6.4e-4$_{\pm 3.4e-4}$ & 8.8e-5$_{\pm 3.1e-5}$ & \textbf{1.2e-4$_{\pm 5.0e-5}$} & 2.8e-4$_{\pm 8.5e-5}$ & 1.7e-4$_{\pm 6.7e-5}$ & 4.2e-6$_{\pm 9.6e-7}$ & 8.0e-5$_{\pm 1.2e-4}$ & 1.0e-4$_{\pm 2.0e-4}$ & 9.3e-5$_{\pm 1.7e-4}$ & 2.4e-4 \\
  SimVQ & .0013$_{\pm .0011}$ & 8.4e-5$_{\pm 3.6e-5}$ & 2.4e-4$_{\pm 2.1e-4}$ & 7.4e-4$_{\pm 3.9e-4}$ & 9.5e-5$_{\pm 4.0e-5}$ & 1.7e-4$_{\pm 6.5e-5}$ & 3.9e-4$_{\pm 1.3e-4}$ & 2.1e-4$_{\pm 8.3e-5}$ & 6.5e-6$_{\pm 2.9e-6}$ & 1.1e-4$_{\pm 1.7e-4}$ & 1.4e-4$_{\pm 2.5e-4}$ & 1.3e-4$_{\pm 2.2e-4}$ & 3.0e-4 \\
  SimQINCo & .0010$_{\pm 9.3e-4}$ & 6.6e-5$_{\pm 2.4e-5}$ & 1.9e-4$_{\pm 1.4e-4}$ & 6.4e-4$_{\pm 3.5e-4}$ & 8.2e-5$_{\pm 2.9e-5}$ & 1.3e-4$_{\pm 5.1e-5}$ & 2.8e-4$_{\pm 9.0e-5}$ & 1.8e-4$_{\pm 6.6e-5}$ & 6.0e-6$_{\pm 3.5e-6}$ & 7.3e-5$_{\pm 1.1e-4}$ & 9.8e-5$_{\pm 1.9e-4}$ & 8.8e-5$_{\pm 1.6e-4}$ & 2.4e-4 \\
  RQVAE & \textbf{8.7e-4$_{\pm 7.5e-4}$} & \textbf{5.2e-5$_{\pm 2.1e-5}$} & \textbf{1.4e-4$_{\pm 1.1e-4}$} & \textbf{5.1e-4$_{\pm 2.8e-4}$} & \textbf{6.0e-5$_{\pm 2.2e-5}$} & 1.3e-4$_{\pm 4.5e-5}$ & \textbf{2.1e-4$_{\pm 6.9e-5}$} & \textbf{1.6e-4$_{\pm 5.2e-5}$} & \textbf{3.0e-6$_{\pm 8.1e-7}$} & \textbf{4.8e-5$_{\pm 7.8e-5}$} & \textbf{6.7e-5$_{\pm 1.5e-4}$} & \textbf{6.0e-5$_{\pm 1.2e-4}$} & 1.9e-4 \\
  DiVeQ & .0014$_{\pm .0012}$ & 9.6e-5$_{\pm 4.1e-5}$ & 2.3e-4$_{\pm 1.9e-4}$ & 8.1e-4$_{\pm 4.2e-4}$ & 1.1e-4$_{\pm 4.6e-5}$ & 1.3e-4$_{\pm 4.6e-5}$ & 4.2e-4$_{\pm 1.3e-4}$ & 2.0e-4$_{\pm 8.3e-5}$ & 5.6e-6$_{\pm 3.0e-6}$ & 1.2e-4$_{\pm 2.0e-4}$ & 1.5e-4$_{\pm 2.8e-4}$ & 1.3e-4$_{\pm 2.5e-4}$ & 3.2e-4 \\
  SF-DiVeQ & .0014$_{\pm .0010}$ & 9.8e-5$_{\pm 4.0e-5}$ & 2.8e-4$_{\pm 2.0e-4}$ & 7.8e-4$_{\pm 4.1e-4}$ & 1.1e-4$_{\pm 4.9e-5}$ & 2.5e-4$_{\pm 1.5e-4}$ & 4.4e-4$_{\pm 1.4e-4}$ & 2.1e-4$_{\pm 8.3e-5}$ & 5.7e-6$_{\pm 2.9e-6}$ & 1.2e-4$_{\pm 1.9e-4}$ & 1.5e-4$_{\pm 2.7e-4}$ & 1.4e-4$_{\pm 2.4e-4}$ & 3.3e-4 \\
  BSQ & .0042$_{\pm .0027}$ & 7.8e-4$_{\pm 4.4e-4}$ & .0013$_{\pm .0010}$ & .0018$_{\pm 6.8e-4}$ & 7.6e-4$_{\pm 3.5e-4}$ & .0012$_{\pm 6.8e-4}$ & .0020$_{\pm 5.8e-4}$ & 9.6e-4$_{\pm 4.0e-4}$ & 6.7e-5$_{\pm 5.3e-5}$ & .0012$_{\pm 1.0e-3}$ & .0013$_{\pm .0013}$ & .0012$_{\pm .0023}$ & .0014 \\
  LFQ & .0049$_{\pm .0059}$ & 9.2e-4$_{\pm 3.6e-4}$ & .0021$_{\pm .0018}$ & .0022$_{\pm 8.7e-4}$ & 9.6e-4$_{\pm 3.9e-4}$ & .0021$_{\pm .0013}$ & .0025$_{\pm 7.0e-4}$ & .0010$_{\pm 9.5e-4}$ & 1.4e-4$_{\pm 1.6e-4}$ & .0014$_{\pm 7.8e-4}$ & .0015$_{\pm 9.8e-4}$ & .0014$_{\pm 9.3e-4}$ & .0018 \\
  \midrule
  \multicolumn{14}{l}{\textit{$f = 8$}} \\
  AEKL & .0040$_{\pm .0025}$ & 6.8e-4$_{\pm 2.0e-4}$ & 9.0e-4$_{\pm 5.7e-4}$ & .0019$_{\pm 7.7e-4}$ & 7.7e-4$_{\pm 2.6e-4}$ & 7.5e-4$_{\pm 2.5e-4}$ & .0021$_{\pm 5.1e-4}$ & 8.6e-4$_{\pm 3.3e-4}$ & 1.1e-4$_{\pm 1.6e-4}$ & 9.9e-4$_{\pm 7.7e-4}$ & .0011$_{\pm 9.0e-4}$ & .0010$_{\pm 8.6e-4}$ & .0013 \\
  SoftVQ & .0050$_{\pm .0032}$ & 7.5e-4$_{\pm 3.1e-4}$ & .0010$_{\pm 7.2e-4}$ & .0020$_{\pm 7.4e-4}$ & 7.2e-4$_{\pm 2.9e-4}$ & .0013$_{\pm 5.7e-4}$ & .0026$_{\pm 7.6e-4}$ & 9.4e-4$_{\pm 4.5e-4}$ & 8.8e-5$_{\pm 5.9e-5}$ & .0013$_{\pm 9.9e-4}$ & .0014$_{\pm .0012}$ & .0013$_{\pm .0011}$ & .0015 \\
  VQGAN & .0046$_{\pm .0033}$ & 3.2e-4$_{\pm 1.5e-4}$ & 5.4e-4$_{\pm 4.5e-4}$ & .0016$_{\pm 7.1e-4}$ & 3.6e-4$_{\pm 1.7e-4}$ & 3.4e-4$_{\pm 1.3e-4}$ & .0017$_{\pm 5.9e-4}$ & 6.4e-4$_{\pm 3.8e-4}$ & 8.4e-6$_{\pm 3.0e-6}$ & 7.3e-4$_{\pm 8.5e-4}$ & 8.5e-4$_{\pm .0010}$ & 7.6e-4$_{\pm 9.9e-4}$ & .0010 \\
  SimVQ & .0031$_{\pm .0023}$ & 2.6e-4$_{\pm 9.6e-5}$ & 3.9e-4$_{\pm 3.2e-4}$ & .0013$_{\pm 6.2e-4}$ & 3.2e-4$_{\pm 1.3e-4}$ & 2.6e-4$_{\pm 1.3e-4}$ & .0011$_{\pm 3.4e-4}$ & 4.1e-4$_{\pm 2.0e-4}$ & 1.5e-5$_{\pm 9.7e-6}$ & 4.8e-4$_{\pm 5.7e-4}$ & 5.6e-4$_{\pm 6.8e-4}$ & 5.0e-4$_{\pm 6.4e-4}$ & 7.2e-4 \\
  SimQINCo & .0033$_{\pm .0024}$ & 2.3e-4$_{\pm 9.9e-5}$ & 4.0e-4$_{\pm 3.3e-4}$ & .0014$_{\pm 6.5e-4}$ & 2.7e-4$_{\pm 1.2e-4}$ & 3.4e-4$_{\pm 1.1e-4}$ & .0012$_{\pm 3.8e-4}$ & 4.3e-4$_{\pm 2.4e-4}$ & 1.5e-5$_{\pm 4.3e-6}$ & 4.7e-4$_{\pm 6.2e-4}$ & 5.5e-4$_{\pm 7.3e-4}$ & 4.9e-4$_{\pm 7.0e-4}$ & 7.6e-4 \\
  RQVAE & \textbf{.0024$_{\pm .0018}$} & \textbf{1.7e-4$_{\pm 7.4e-5}$} & \textbf{3.1e-4$_{\pm 2.7e-4}$} & \textbf{.0011$_{\pm 5.6e-4}$} & \textbf{2.0e-4$_{\pm 8.8e-5}$} & 2.3e-4$_{\pm 6.7e-5}$ & \textbf{8.8e-4$_{\pm 2.9e-4}$} & \textbf{3.0e-4$_{\pm 1.4e-4}$} & \textbf{4.3e-6$_{\pm 1.1e-6}$} & \textbf{2.8e-4$_{\pm 4.2e-4}$} & \textbf{3.4e-4$_{\pm 5.2e-4}$} & \textbf{3.1e-4$_{\pm 4.8e-4}$} & 5.5e-4 \\
  DiVeQ & .0032$_{\pm .0023}$ & 2.4e-4$_{\pm 1.0e-4}$ & 4.2e-4$_{\pm 3.4e-4}$ & .0013$_{\pm 6.3e-4}$ & 2.7e-4$_{\pm 1.2e-4}$ & 2.8e-4$_{\pm 1.1e-4}$ & .0012$_{\pm 4.1e-4}$ & 4.3e-4$_{\pm 2.3e-4}$ & 9.2e-6$_{\pm 4.9e-6}$ & 4.5e-4$_{\pm 6.0e-4}$ & 5.3e-4$_{\pm 7.0e-4}$ & 4.8e-4$_{\pm 6.7e-4}$ & 7.3e-4 \\
  SF-DiVeQ & .0027$_{\pm .0020}$ & 1.9e-4$_{\pm 8.2e-5}$ & 3.4e-4$_{\pm 2.9e-4}$ & .0012$_{\pm 6.0e-4}$ & 2.4e-4$_{\pm 1.1e-4}$ & \textbf{1.9e-4$_{\pm 6.2e-5}$} & 9.7e-4$_{\pm 3.0e-4}$ & 3.4e-4$_{\pm 1.8e-4}$ & 1.3e-5$_{\pm 6.5e-6}$ & 3.8e-4$_{\pm 5.2e-4}$ & 4.5e-4$_{\pm 6.2e-4}$ & 4.1e-4$_{\pm 5.8e-4}$ & 6.2e-4 \\
  BSQ & .0070$_{\pm .0054}$ & .0013$_{\pm 5.1e-4}$ & .0024$_{\pm .0014}$ & .0024$_{\pm 8.8e-4}$ & .0015$_{\pm 5.0e-4}$ & .0023$_{\pm .0013}$ & .0036$_{\pm .0012}$ & .0018$_{\pm 7.7e-4}$ & 2.0e-4$_{\pm 1.6e-4}$ & .0024$_{\pm .0014}$ & .0025$_{\pm .0017}$ & .0024$_{\pm .0016}$ & .0025 \\
  LFQ & .0123$_{\pm .0082}$ & .0030$_{\pm .0016}$ & .0050$_{\pm .0067}$ & .0039$_{\pm .0019}$ & .0036$_{\pm .0016}$ & .0038$_{\pm .0021}$ & .0108$_{\pm .0062}$ & .0043$_{\pm .0020}$ & 7.4e-4$_{\pm .0015}$ & .0070$_{\pm .0047}$ & .0077$_{\pm .0053}$ & .0075$_{\pm .0055}$ & .0058 \\
  \midrule
  \multicolumn{14}{l}{\textit{$f = 16$}} \\
  AEKL & .0049$_{\pm .0032}$ & 6.6e-4$_{\pm 2.2e-4}$ & 8.6e-4$_{\pm 5.4e-4}$ & .0019$_{\pm 8.1e-4}$ & 7.7e-4$_{\pm 2.8e-4}$ & .0011$_{\pm 3.9e-4}$ & .0023$_{\pm 6.7e-4}$ & 9.2e-4$_{\pm 4.6e-4}$ & 9.6e-5$_{\pm 4.0e-5}$ & .0011$_{\pm 9.3e-4}$ & .0012$_{\pm .0011}$ & .0011$_{\pm .0011}$ & .0014 \\
  SoftVQ & .0050$_{\pm .0035}$ & 4.2e-4$_{\pm 1.7e-4}$ & 6.5e-4$_{\pm 5.6e-4}$ & .0017$_{\pm 7.5e-4}$ & 4.8e-4$_{\pm 2.2e-4}$ & 5.1e-4$_{\pm 2.1e-4}$ & .0025$_{\pm 8.0e-4}$ & 7.3e-4$_{\pm 4.4e-4}$ & 4.0e-5$_{\pm 2.3e-5}$ & 9.7e-4$_{\pm 9.5e-4}$ & .0011$_{\pm .0012}$ & .0010$_{\pm .0011}$ & .0013 \\
  VQGAN & .0082$_{\pm .0061}$ & 8.2e-4$_{\pm 3.4e-4}$ & 9.3e-4$_{\pm 8.3e-4}$ & .0019$_{\pm 7.9e-4}$ & 9.7e-4$_{\pm 4.4e-4}$ & .0010$_{\pm 3.6e-4}$ & .0038$_{\pm .0013}$ & .0015$_{\pm 8.7e-4}$ & 9.2e-5$_{\pm 6.7e-5}$ & .0023$_{\pm .0017}$ & .0026$_{\pm .0022}$ & .0024$_{\pm .0022}$ & .0022 \\
  SimVQ & .0074$_{\pm .0050}$ & .0010$_{\pm 3.7e-4}$ & .0011$_{\pm 8.3e-4}$ & .0021$_{\pm 8.3e-4}$ & .0012$_{\pm 4.6e-4}$ & .0010$_{\pm 3.0e-4}$ & .0038$_{\pm .0012}$ & .0015$_{\pm 8.1e-4}$ & 1.5e-4$_{\pm 9.7e-5}$ & .0024$_{\pm .0016}$ & .0025$_{\pm .0019}$ & .0023$_{\pm .0019}$ & .0022 \\
  SimQINCo & .0064$_{\pm .0047}$ & 5.2e-4$_{\pm 2.3e-4}$ & 6.3e-4$_{\pm 6.2e-4}$ & .0017$_{\pm 7.7e-4}$ & 6.2e-4$_{\pm 2.9e-4}$ & 4.1e-4$_{\pm 1.3e-4}$ & .0028$_{\pm 9.6e-4}$ & 9.7e-4$_{\pm 6.2e-4}$ & 2.9e-5$_{\pm 1.8e-5}$ & .0014$_{\pm .0012}$ & .0016$_{\pm .0015}$ & .0014$_{\pm .0015}$ & .0015 \\
  RQVAE & \textbf{.0044$_{\pm .0031}$} & \textbf{3.1e-4$_{\pm 1.5e-4}$} & \textbf{5.0e-4$_{\pm 4.8e-4}$} & \textbf{.0015$_{\pm 7.3e-4}$} & \textbf{3.9e-4$_{\pm 1.8e-4}$} & \textbf{3.3e-4$_{\pm 1.1e-4}$} & \textbf{.0018$_{\pm 6.2e-4}$} & \textbf{6.0e-4$_{\pm 3.8e-4}$} & \textbf{5.9e-6$_{\pm 2.4e-6}$} & \textbf{7.3e-4$_{\pm 8.5e-4}$} & \textbf{8.7e-4$_{\pm .0010}$} & \textbf{7.8e-4$_{\pm 1.0e-3}$} & .0010 \\
  DiVeQ & .0065$_{\pm .0047}$ & 5.9e-4$_{\pm 2.4e-4}$ & 7.7e-4$_{\pm 7.2e-4}$ & .0017$_{\pm 7.7e-4}$ & 7.0e-4$_{\pm 3.2e-4}$ & 8.5e-4$_{\pm 1.9e-4}$ & .0030$_{\pm .0010}$ & .0010$_{\pm 6.4e-4}$ & 3.4e-5$_{\pm 2.5e-5}$ & .0015$_{\pm .0013}$ & .0017$_{\pm .0016}$ & .0015$_{\pm .0016}$ & .0017 \\
  SF-DiVeQ & .0064$_{\pm .0047}$ & 5.4e-4$_{\pm 2.3e-4}$ & 6.9e-4$_{\pm 6.4e-4}$ & .0017$_{\pm 7.7e-4}$ & 6.5e-4$_{\pm 3.0e-4}$ & 4.4e-4$_{\pm 1.4e-4}$ & .0028$_{\pm 9.7e-4}$ & .0010$_{\pm 6.6e-4}$ & 4.1e-5$_{\pm 2.9e-5}$ & .0015$_{\pm .0012}$ & .0016$_{\pm .0016}$ & .0015$_{\pm .0015}$ & .0016 \\
  BSQ & .0080$_{\pm .0052}$ & .0013$_{\pm 5.2e-4}$ & .0017$_{\pm .0012}$ & .0027$_{\pm 9.5e-4}$ & .0014$_{\pm 6.7e-4}$ & .0017$_{\pm 9.5e-4}$ & .0042$_{\pm .0012}$ & .0018$_{\pm .0010}$ & 1.9e-4$_{\pm 1.4e-4}$ & .0029$_{\pm .0018}$ & .0030$_{\pm .0022}$ & .0029$_{\pm .0023}$ & .0027 \\
  LFQ & .0162$_{\pm .0112}$ & .0038$_{\pm .0017}$ & .0050$_{\pm .0035}$ & .0062$_{\pm .0027}$ & .0039$_{\pm .0017}$ & .0043$_{\pm .0022}$ & .0101$_{\pm .0036}$ & .0048$_{\pm .0021}$ & 8.2e-4$_{\pm 6.6e-4}$ & .0088$_{\pm .0045}$ & .0094$_{\pm .0054}$ & .0088$_{\pm .0053}$ & .0068 \\
  \bottomrule
  \end{tabular}}
\end{table*}

\begin{table}[h]
  \centering
  \caption{Per-dataset reconstruction-FID ($\downarrow$). Dataset-level metric (no per-image variance).}
  \label{tab:rfid}
  \setlength{\tabcolsep}{3pt}
  \adjustbox{max width=0.48\textwidth,max totalheight=0.43\textheight,center}{%
  \begin{tabular}{lcccccccccccc|c}
  \toprule
  Method & Path & Chest & Derma & OCT & Pneu & Reti & Breast & Blood & Tiss & OrgA & OrgC & OrgS & mean \\
  \midrule
  \multicolumn{14}{l}{\textit{$f = 4$}} \\
  AEKL & 37.9 & 23.3 & 71.3 & 78.6 & 47.5 & 56.9 & 104 & 19.2 & 29.0 & 45.0 & 27.6 & 28.1 & 47.4 \\
  SoftVQ & 23.2 & 7.06 & 34.2 & 34.3 & 20.4 & 31.7 & 76.8 & 10.1 & 2.52 & 11.0 & 6.14 & 5.65 & 21.9 \\
  VQGAN & 11.1 & 4.53 & 14.4 & 15.9 & 10.8 & 9.44 & 33.6 & 4.79 & 2.77 & 1.68 & 2.24 & 1.96 & 9.44 \\
  SimVQ & 14.7 & 5.32 & 16.5 & 18.3 & 12.2 & 12.5 & 47.1 & 6.35 & 1.96 & 1.51 & 2.04 & 1.85 & 11.7 \\
  SimQINCo & 11.6 & 4.03 & 14.5 & 18.3 & 10.1 & \textbf{8.88} & 33.0 & 5.17 & 3.69 & 1.47 & 2.02 & 1.78 & 9.54 \\
  RQVAE & \textbf{7.29} & \textbf{3.47} & \textbf{12.2} & \textbf{10.5} & \textbf{7.34} & 10.9 & \textbf{23.8} & \textbf{4.56} & \textbf{0.58} & \textbf{0.85} & \textbf{1.32} & \textbf{1.18} & 6.99 \\
  DiVeQ & 16.8 & 5.93 & 16.0 & 24.0 & 14.2 & 11.5 & 49.8 & 6.35 & 1.72 & 1.65 & 2.16 & 2.01 & 12.7 \\
  SF-DiVeQ & 14.8 & 5.77 & 17.3 & 18.5 & 14.2 & 14.6 & 47.2 & 7.27 & 1.90 & 2.16 & 2.78 & 2.52 & 12.4 \\
  BSQ & 28.0 & 11.4 & 69.0 & 43.4 & 29.9 & 47.7 & 84.7 & 18.8 & 2.51 & 31.4 & 17.7 & 17.1 & 33.5 \\
  LFQ & 37.9 & 15.0 & 49.6 & 37.2 & 42.4 & 66.8 & 87.1 & 17.4 & 4.48 & 58.9 & 31.4 & 33.7 & 40.1 \\
  \midrule
  \multicolumn{14}{l}{\textit{$f = 8$}} \\
  AEKL & 41.8 & 25.1 & 56.1 & 86.2 & 47.5 & 43.1 & 136 & 21.8 & 15.0 & 43.3 & 27.7 & 27.2 & 47.5 \\
  SoftVQ & 42.9 & 17.9 & 69.7 & 101 & 39.8 & 57.6 & 133 & 20.0 & 2.12 & 54.8 & 32.8 & 33.0 & 50.4 \\
  VQGAN & 35.6 & 15.6 & 57.7 & 93.3 & 36.1 & 34.7 & 125 & 22.5 & \textbf{1.95} & 42.6 & 30.2 & 28.5 & 43.7 \\
  SimVQ & 35.2 & \textbf{6.90} & 21.9 & 36.1 & 23.0 & 22.3 & 88.0 & 13.0 & 2.75 & 17.6 & 11.3 & 9.98 & 24.0 \\
  SimQINCo & 33.3 & 10.5 & 32.1 & 41.1 & 25.7 & 24.4 & 97.2 & 16.2 & 2.23 & 15.4 & 10.8 & 10.2 & 26.6 \\
  RQVAE & \textbf{29.9} & 7.64 & 25.2 & \textbf{28.7} & \textbf{20.1} & 20.5 & \textbf{68.7} & \textbf{11.3} & 2.11 & \textbf{4.67} & \textbf{4.69} & \textbf{3.98} & 19.0 \\
  DiVeQ & 34.3 & 9.71 & 28.0 & 39.9 & 24.3 & 24.6 & 80.7 & 13.8 & 5.39 & 12.3 & 8.93 & 8.20 & 24.2 \\
  SF-DiVeQ & 38.6 & 10.0 & \textbf{21.7} & 35.4 & 24.0 & \textbf{19.6} & 87.2 & 11.9 & 2.10 & 8.72 & 7.15 & 6.24 & 22.7 \\
  BSQ & 49.8 & 18.9 & 105 & 73.9 & 41.9 & 75.7 & 115 & 23.8 & 7.98 & 66.8 & 37.4 & 37.9 & 54.5 \\
  LFQ & 75.3 & 70.4 & 123 & 109 & 111 & 155 & 155 & 53.8 & 55.5 & 135 & 99.0 & 103 & 104 \\
  \midrule
  \multicolumn{14}{l}{\textit{$f = 16$}} \\
  AEKL & 52.3 & 32.6 & 69.3 & 92.0 & 56.5 & 40.6 & 160 & 24.9 & 12.2 & 53.6 & 38.0 & 38.4 & 55.9 \\
  SoftVQ & \textbf{44.8} & \textbf{19.1} & 45.4 & 81.8 & 42.3 & 42.4 & 136 & 22.8 & \textbf{1.74} & 57.9 & 41.4 & 40.5 & 48.1 \\
  VQGAN & 70.6 & 42.5 & 86.4 & 102 & 68.4 & 49.5 & 251 & 33.9 & 9.86 & 95.4 & 70.6 & 70.6 & 79.2 \\
  SimVQ & 62.9 & 52.9 & 88.7 & 90.9 & 78.0 & 48.8 & 240 & 31.6 & 24.2 & 109 & 79.1 & 80.1 & 82.2 \\
  SimQINCo & 53.8 & 30.4 & 62.7 & 78.2 & 56.8 & 48.5 & 231 & 24.8 & 1.94 & 77.3 & 54.8 & 54.5 & 64.5 \\
  RQVAE & 46.1 & 22.9 & \textbf{32.3} & \textbf{54.8} & \textbf{40.7} & \textbf{35.7} & \textbf{123} & \textbf{19.2} & 2.83 & \textbf{32.1} & \textbf{25.4} & \textbf{23.6} & 38.2 \\
  DiVeQ & 57.8 & 30.1 & 62.0 & 68.6 & 52.8 & 43.6 & 193 & 25.4 & 1.83 & 84.5 & 59.0 & 60.5 & 61.6 \\
  SF-DiVeQ & 53.2 & 26.8 & 60.0 & 67.8 & 48.2 & 40.0 & 172 & 23.8 & 2.70 & 78.4 & 55.3 & 55.7 & 57.0 \\
  BSQ & 51.5 & 30.8 & 95.4 & 103 & 52.6 & 65.6 & 154 & 37.0 & 9.27 & 96.9 & 58.7 & 60.2 & 67.9 \\
  LFQ & 156 & 117 & 201 & 137 & 143 & 131 & 252 & 86.3 & 88.1 & 179 & 141 & 145 & 148 \\
  \bottomrule
  \end{tabular}}
\end{table}

\begin{table}[h]
  \centering
  \caption{Per-dataset reconstruction-FMD (rFMD, $\downarrow$). Dataset-level metric using each dataset's own MedMNIST feature extractor; the aggregate is the \emph{median} over datasets (scales differ per extractor).}
  \label{tab:rfmd}
  \setlength{\tabcolsep}{3pt}
  \adjustbox{max width=0.48\textwidth,max totalheight=0.43\textheight,center}{%
  \begin{tabular}{lcccccccccccc|c}
  \toprule
  Method & Path & Chest & Derma & OCT & Pneu & Reti & Breast & Blood & Tiss & OrgA & OrgC & OrgS & median \\
  \midrule
  \multicolumn{14}{l}{\textit{$f = 4$}} \\
  AEKL & 17.3 & 0.829 & 4.22 & 2.30 & 0.770 & 7.09 & 59.9 & 16.8 & 1.04 & 0.611 & 0.581 & 0.573 & 1.67 \\
  SoftVQ & 9.01 & 0.053 & 4.26 & 0.295 & 0.104 & 24.8 & 20.6 & 6.69 & 0.165 & 0.143 & 0.150 & 0.147 & 0.230 \\
  VQGAN & \textbf{2.01} & 0.005 & 1.50 & 0.069 & 0.021 & 1.32 & 1.29 & \textbf{0.606} & \textbf{0.023} & 0.013 & 0.010 & \textbf{0.010} & 0.046 \\
  SimVQ & 5.62 & 0.007 & 1.19 & 0.116 & 0.007 & \textbf{1.24} & 1.19 & 2.61 & 0.030 & 0.058 & 0.042 & 0.046 & 0.087 \\
  SimpleQINCo & 6.16 & 0.004 & \textbf{0.528} & 0.128 & 0.016 & 11.8 & 5.39 & 2.70 & 0.024 & 0.070 & 0.062 & 0.068 & 0.099 \\
  RQVAE & 2.41 & 0.004 & 0.781 & 0.043 & 0.012 & 1.64 & \textbf{0.851} & 3.22 & 0.029 & 0.021 & 0.018 & 0.015 & \textbf{0.036} \\
  DiVeQ & 4.83 & 0.009 & 2.14 & 0.105 & 0.057 & 2.07 & 6.51 & 2.55 & 0.036 & 0.091 & 0.064 & 0.062 & 0.098 \\
  SF-DiVeQ & 5.45 & \textbf{0.003} & 1.08 & \textbf{0.036} & \textbf{0.007} & 5.44 & 1.96 & 3.36 & 0.040 & \textbf{0.008} & \textbf{0.007} & 0.010 & 0.038 \\
  BSQ & 26.2 & 0.168 & 8.99 & 1.05 & 0.189 & 22.1 & 66.2 & 21.8 & 0.585 & 0.470 & 0.569 & 0.557 & 0.819 \\
  LFQ & 25.9 & 0.186 & 11.3 & 1.29 & 0.180 & 107 & 37.6 & 21.1 & 0.449 & 0.318 & 0.424 & 0.593 & 0.939 \\
  \midrule
  \multicolumn{14}{l}{\textit{$f = 8$}} \\
  AEKL & 18.3 & 0.447 & 3.49 & 2.29 & 0.230 & 32.5 & 56.0 & 12.4 & 0.545 & 0.561 & 0.295 & 0.469 & 1.43 \\
  SoftVQ & 23.5 & 0.137 & 3.85 & 1.76 & 0.268 & 32.9 & 117 & 7.27 & 0.281 & 0.780 & 0.428 & 0.647 & 1.27 \\
  VQGAN & 15.7 & 0.039 & 7.45 & 0.547 & 0.032 & 4.90 & 32.8 & 2.58 & 0.105 & 0.177 & 0.155 & 0.325 & 0.436 \\
  SimVQ & 12.5 & 0.025 & \textbf{0.538} & 0.279 & 0.039 & 4.30 & 8.63 & 2.87 & \textbf{0.029} & 0.110 & 0.058 & 0.103 & 0.194 \\
  SimpleQINCo & 14.2 & 0.025 & 1.99 & 0.298 & 0.074 & 40.5 & 23.8 & 8.78 & 0.522 & 0.237 & 0.155 & 0.186 & 0.410 \\
  RQVAE & 7.45 & \textbf{0.009} & 0.945 & \textbf{0.121} & \textbf{0.014} & 4.43 & \textbf{3.30} & 3.43 & 0.039 & 0.041 & \textbf{0.016} & \textbf{0.031} & \textbf{0.081} \\
  DiVeQ & \textbf{7.44} & 0.015 & 1.79 & 0.308 & 0.016 & 5.77 & 8.03 & \textbf{2.54} & 0.086 & \textbf{0.025} & 0.030 & 0.047 & 0.197 \\
  SF-DiVeQ & 8.58 & 0.024 & 0.842 & 0.283 & 0.037 & \textbf{1.76} & 9.74 & 2.97 & 0.064 & 0.074 & 0.063 & 0.099 & 0.191 \\
  BSQ & 53.6 & 1.01 & 14.7 & 2.32 & 0.363 & 167 & 111 & 64.4 & 2.65 & 0.685 & 0.744 & 1.15 & 2.48 \\
  LFQ & 61.5 & 2.42 & 17.8 & 5.43 & 8.99 & 152 & 540 & 165 & 7.48 & 9.44 & 13.2 & 13.5 & 13.4 \\
  \midrule
  \multicolumn{14}{l}{\textit{$f = 16$}} \\
  AEKL & 20.9 & 0.971 & 7.05 & 3.05 & 0.421 & 7.20 & 69.0 & 15.4 & 0.611 & 0.567 & 0.519 & 0.833 & 2.01 \\
  SoftVQ & 22.2 & 0.108 & 4.16 & 1.37 & 0.160 & 29.8 & 83.4 & 11.2 & 0.334 & 0.493 & 0.290 & 0.473 & 0.933 \\
  VQGAN & 52.9 & 0.700 & 4.04 & 2.43 & 0.420 & 62.4 & 249 & 64.1 & 1.17 & 1.72 & 1.59 & 2.01 & 2.22 \\
  SimVQ & 47.3 & 2.12 & 4.71 & 3.61 & 1.00 & 12.8 & 234 & 39.0 & 1.13 & 2.04 & 1.79 & 2.09 & 2.86 \\
  SimpleQINCo & 31.1 & 0.192 & \textbf{1.10} & 0.982 & 0.180 & 6.02 & 103 & 11.5 & 0.155 & 0.747 & 0.626 & 0.993 & 0.988 \\
  RQVAE & \textbf{17.9} & \textbf{0.025} & 3.17 & \textbf{0.587} & \textbf{0.061} & 8.37 & \textbf{41.5} & \textbf{9.22} & \textbf{0.028} & \textbf{0.203} & \textbf{0.175} & \textbf{0.291} & \textbf{0.439} \\
  DiVeQ & 27.1 & 0.204 & 5.97 & 1.49 & 0.189 & 52.5 & 121 & 11.6 & 0.060 & 1.06 & 0.693 & 1.18 & 1.33 \\
  SF-DiVeQ & 26.1 & 0.174 & 3.36 & 1.53 & 0.115 & \textbf{1.14} & 147 & 13.9 & 0.177 & 1.23 & 0.851 & 1.13 & 1.18 \\
  BSQ & 50.3 & 0.867 & 16.0 & 3.03 & 0.449 & 32.9 & 215 & 41.0 & 0.623 & 1.74 & 1.70 & 1.86 & 2.45 \\
  LFQ & 62.6 & 6.16 & 18.4 & 9.56 & 7.89 & 111 & 655 & 115 & 8.44 & 15.5 & 23.5 & 20.1 & 19.2 \\
  \bottomrule
  \end{tabular}}
\end{table}

\section{Qualitative Reconstructions}
\label{sec:supp-qual}

\begin{figure*}[h]
  \centering
  \begin{minipage}{\textwidth}
    \centering
    \includegraphics[width=\textwidth,height=0.27\textheight,keepaspectratio]{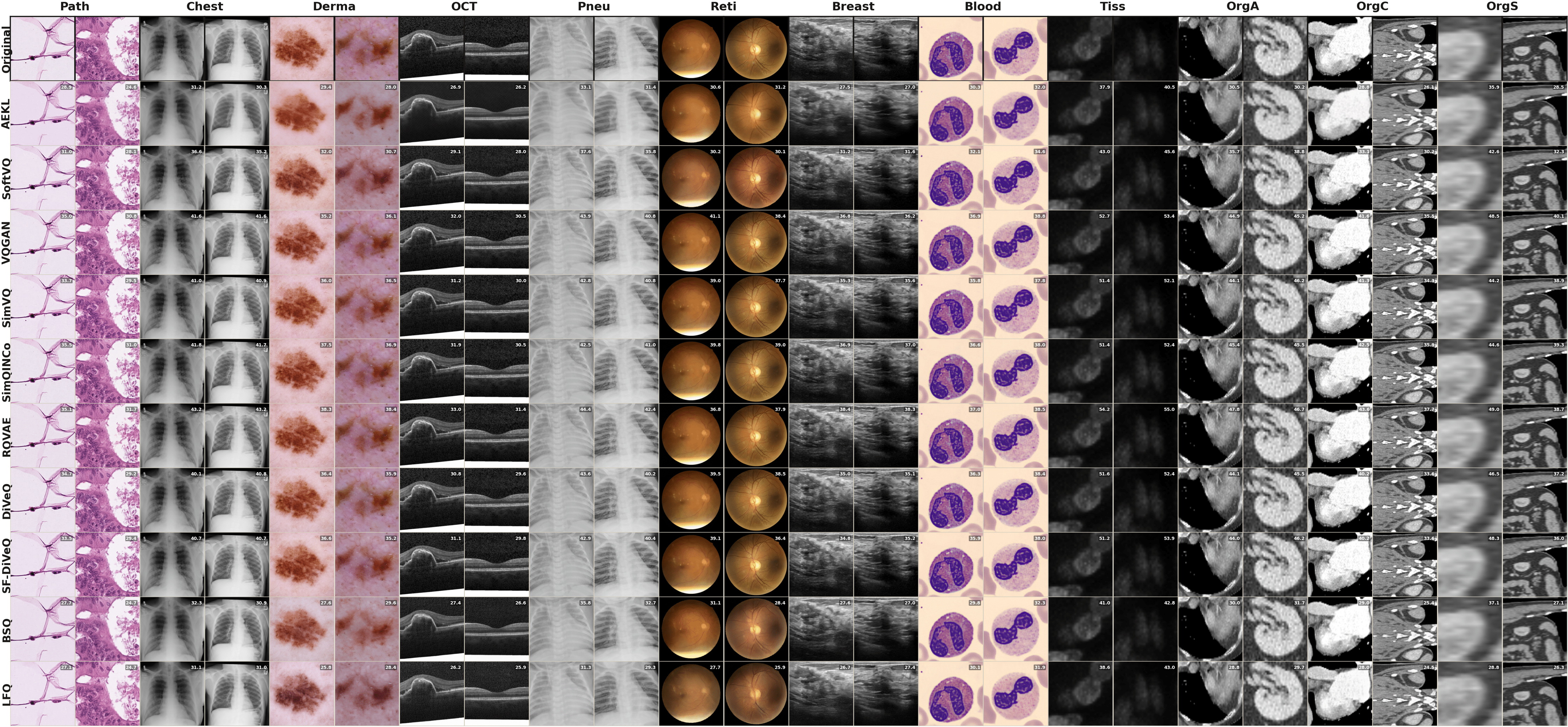}
    \subcaption{$f{=}4$}
    \label{fig:recon-f4}
  \end{minipage}

  \vspace{4pt}
  \begin{minipage}{\textwidth}
    \centering
    \includegraphics[width=\textwidth,height=0.27\textheight,keepaspectratio]{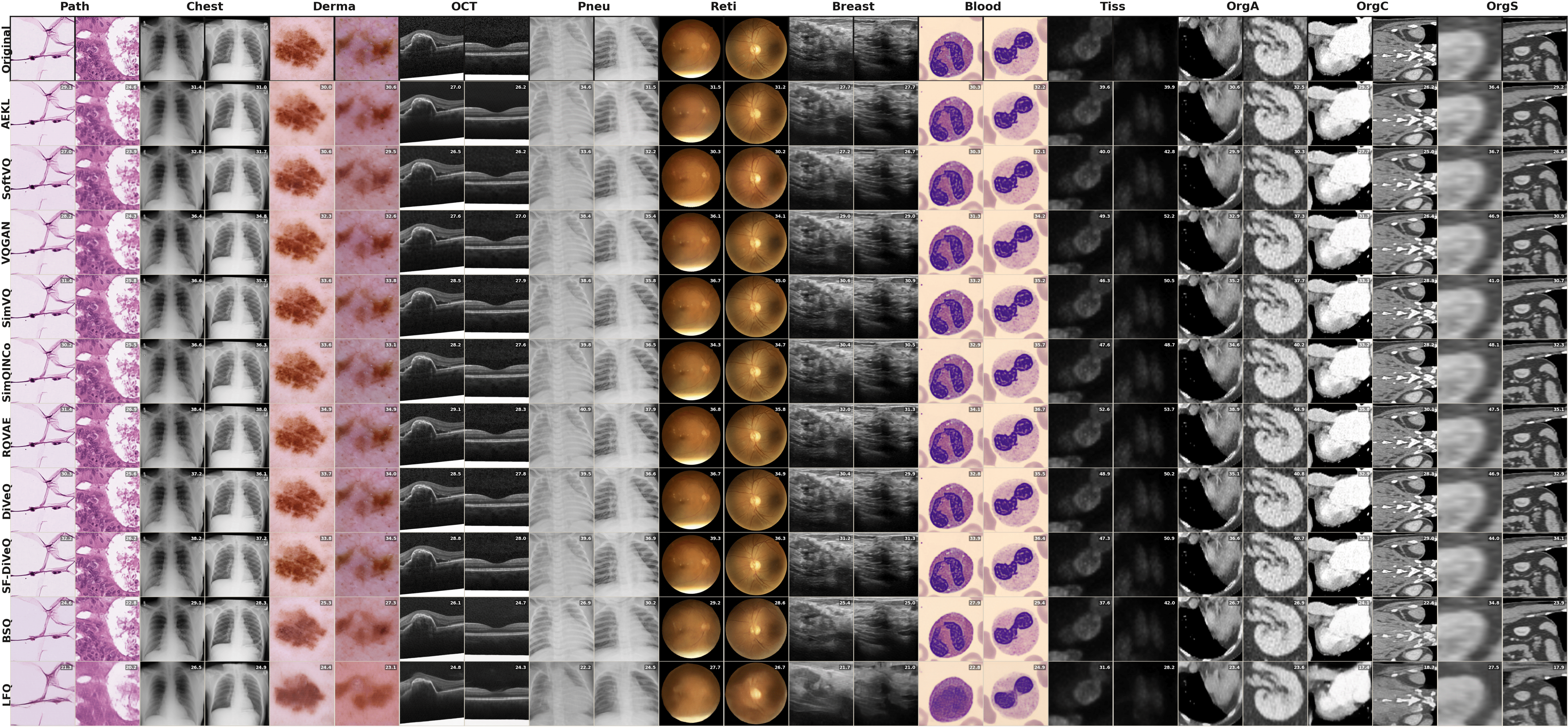}
    \subcaption{$f{=}8$}
    \label{fig:recon-f8}
  \end{minipage}

  \vspace{4pt}
  \begin{minipage}{\textwidth}
    \centering
    \includegraphics[width=\textwidth,height=0.27\textheight,keepaspectratio]{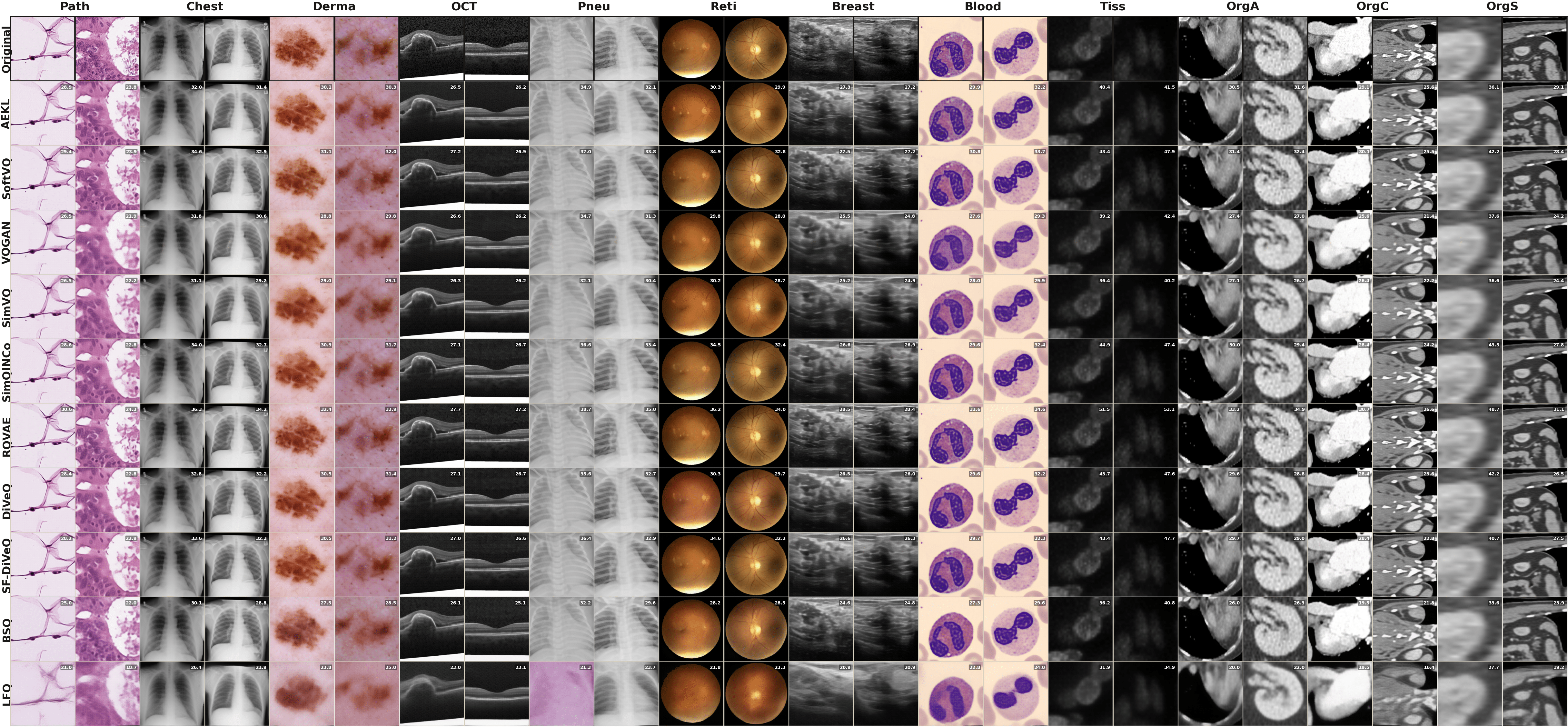}
    \subcaption{$f{=}16$}
    \label{fig:recon-f16}
  \end{minipage}

  \caption{Qualitative reconstructions of two fixed test subjects per dataset
  (columns) by every tokenizer family (rows), at compression factors
  \subref{fig:recon-f4}~$f{=}4$, \subref{fig:recon-f8}~$f{=}8$, and
  \subref{fig:recon-f16}~$f{=}16$. Top row of each panel is the original;
  top-right of each cell is per-image PSNR (dB). Reconstruction differences
  across families grow with compression.}
  \label{fig:recon-qual}
\end{figure*}

Figure~\ref{fig:recon-qual} shows two fixed test subjects (indices 7 and 12)
from every MedMNIST dataset, reconstructed by each tokenizer family at the
three compression factors. The top row of each panel is the original; the
number in the top-right of each cell is that image's PSNR (dB) against its
reference.

\section{Quantitative Generation}
\label{sec:supp-gen}
\Cref{tab:gfid} gives per-dataset DiT-B/2 generation-FID. The ranking tracks
reconstruction closely---consistent with the $r{=}0.82$ correlation in the main
paper---with the soft and space-filling schemes (DiVeQ, SF-DiVeQ) leading at
low-to-mid compression and RQVAE retaking the lead at $f{=}16$.

\begin{table}[h]
  \centering
  \caption{Per-dataset DiT-B/2 generation-FID ($\downarrow$). Dataset-level metric.}
  \label{tab:gfid}
  \setlength{\tabcolsep}{3pt}
  \adjustbox{max width=0.48\textwidth,max totalheight=0.43\textheight,center}{%
  \begin{tabular}{lcccccccccccc|c}
  \toprule
  Method & Path & Chest & Derma & OCT & Pneu & Reti & Breast & Blood & Tiss & OrgA & OrgC & OrgS & mean \\
  \midrule
  \multicolumn{14}{l}{\textit{$f = 4$}} \\
  AEKL & 68.4 & 33.2 & 100 & 122 & 85.1 & 81.3 & 137 & 36.4 & 41.7 & 86.8 & 58.9 & 63.3 & 76.2 \\
  SoftVQ & 111 & 21.1 & 102 & 83.3 & 101 & 69.0 & 155 & 52.0 & 13.5 & 82.9 & 80.3 & 88.9 & 79.9 \\
  VQGAN & \textbf{60.7} & 18.4 & \textbf{52.8} & 52.7 & 68.1 & 53.7 & 105 & \textbf{29.4} & 11.9 & 42.7 & 53.3 & 57.7 & 50.5 \\
  SimVQ & 64.3 & 14.9 & 77.3 & 57.3 & \textbf{50.4} & 51.2 & 98.3 & 37.7 & 9.78 & 39.2 & 43.8 & 48.4 & 49.4 \\
  SimQINCo & 75.3 & \textbf{11.2} & 62.0 & \textbf{51.6} & 58.8 & 52.4 & 108 & 38.1 & 8.29 & 42.6 & 61.5 & 58.2 & 52.3 \\
  RQVAE & 66.8 & 14.1 & 74.9 & 52.1 & 76.3 & \textbf{36.2} & 105 & 42.4 & 8.64 & 44.0 & 50.9 & 54.8 & 52.2 \\
  DiVeQ & 68.7 & 23.2 & 62.3 & 60.9 & 95.9 & 36.3 & \textbf{95.7} & 34.8 & \textbf{5.07} & \textbf{27.0} & \textbf{32.0} & \textbf{35.2} & 48.1 \\
  SF-DiVeQ & 63.0 & 16.6 & 70.7 & 59.0 & 61.4 & 46.0 & 101 & 43.5 & 5.12 & 37.5 & 36.5 & 44.2 & 48.7 \\
  BSQ & 115 & 38.6 & 74.6 & 89.1 & 154 & 76.7 & 137 & 89.3 & 46.1 & 148 & 140 & 146 & 104 \\
  LFQ & 133 & 61.2 & 131 & 95.1 & 167 & 149 & 182 & 126 & 85.4 & 241 & 163 & 170 & 142 \\
  \midrule
  \multicolumn{14}{l}{\textit{$f = 8$}} \\
  AEKL & 78.9 & 38.0 & 76.9 & 130 & 67.0 & 53.9 & 163 & 39.9 & 24.4 & 83.5 & 63.1 & 68.5 & 73.9 \\
  SoftVQ & 85.4 & 45.7 & 107 & 145 & 133 & 121 & 194 & 89.3 & 23.8 & 140 & 102 & 108 & 108 \\
  VQGAN & 73.1 & 19.7 & 76.1 & 127 & 65.5 & 44.1 & 168 & 31.0 & 6.80 & 69.2 & 55.6 & 50.6 & 65.5 \\
  SimVQ & 66.9 & 24.7 & 57.7 & 78.7 & 125 & 44.2 & 116 & 43.6 & 20.9 & 74.5 & 51.9 & 55.1 & 63.2 \\
  SimQINCo & 69.7 & 19.4 & \textbf{56.2} & 86.5 & 61.1 & 53.7 & 126 & \textbf{26.1} & \textbf{5.01} & 45.4 & \textbf{40.3} & 44.7 & 52.8 \\
  RQVAE & {\textbf{60.2}} & {\textbf{17.9}} & {91.2} & {\textbf{68.8}} & {80.8} & {59.7} & {112} & {34.3} & {7.98} & {\textbf{41.3}} & {45.6} & {46.1} & {55.5} \\
  DiVeQ & 90.5 & 21.4 & 79.6 & 81.5 & \textbf{59.0} & \textbf{41.1} & 120 & 41.6 & 15.8 & 52.5 & 46.3 & 51.6 & 58.4 \\
  SF-DiVeQ & 74.8 & 27.6 & 64.4 & 91.1 & 112 & 49.4 & \textbf{110} & 29.9 & 14.5 & 44.8 & 45.1 & \textbf{43.4} & 58.8 \\
  BSQ & 93.6 & 33.5 & 94.3 & 112 & 60.2 & 76.9 & 135 & 78.6 & 15.1 & 102 & 88.1 & 88.0 & 81.5 \\
  LFQ & 189 & 181 & 259 & 256 & 272 & 181 & 318 & 226 & 126 & 257 & 218 & 230 & 226 \\
  \midrule
  \multicolumn{14}{l}{\textit{$f = 16$}} \\
  AEKL & 78.7 & 42.9 & 95.9 & 126 & 89.0 & 66.6 & 218 & 47.2 & 12.1 & 113 & 81.2 & 95.5 & 88.8 \\
  SoftVQ & {88.7} & {39.0} & {108} & {118} & {127} & {107} & {220} & {102} & {20.5} & {158} & {137} & {140} & {114} \\
  VQGAN & 91.0 & 53.0 & 122 & 148 & 85.7 & \textbf{53.4} & 298 & 37.9 & 14.7 & 110 & 93.3 & 87.8 & 99.5 \\
  SimVQ & 92.1 & 69.9 & 141 & 120 & 121 & 58.2 & 274 & 58.2 & 37.7 & 139 & 113 & 116 & 112 \\
  SimQINCo & \textbf{68.1} & 38.7 & 113 & 120 & 85.5 & 68.8 & 285 & 42.4 & \textbf{4.56} & 111 & 76.8 & 84.1 & 91.5 \\
  RQVAE & 73.0 & \textbf{32.9} & \textbf{71.0} & \textbf{101} & \textbf{66.4} & 63.9 & \textbf{182} & 52.6 & 13.1 & \textbf{65.0} & \textbf{61.5} & \textbf{63.3} & 70.5 \\
  DiVeQ & 75.6 & 37.1 & 78.5 & 109 & 68.6 & 60.7 & 257 & 40.3 & 4.79 & 142 & 116 & 105 & 91.2 \\
  SF-DiVeQ & 74.1 & 34.1 & 83.5 & 112 & 70.7 & 60.0 & 206 & \textbf{36.9} & 4.99 & 124 & 100 & 99.0 & 83.7 \\
  BSQ & 107 & 41.0 & 106 & 138 & 73.7 & 58.1 & 213 & 73.6 & 13.5 & 145 & 133 & 129 & 102 \\
  LFQ & 154 & 103 & 181 & 108 & 113 & 121 & 298 & 59.9 & 74.9 & 166 & 144 & 138 & 138 \\
  \bottomrule
  \end{tabular}}
\end{table}

\begin{table}[h]
  \centering
  \caption{Per-dataset DiT-B/2 generation-FMD (gFMD, $\downarrow$). Dataset-level metric using each dataset's own MedMNIST feature extractor; the aggregate is the \emph{median} over datasets.}
  \label{tab:gfmd}
  \setlength{\tabcolsep}{3pt}
  \adjustbox{max width=0.48\textwidth,max totalheight=0.43\textheight,center}{%
  \begin{tabular}{lcccccccccccc|c}
  \toprule
  Method & Path & Chest & Derma & OCT & Pneu & Reti & Breast & Blood & Tiss & OrgA & OrgC & OrgS & median \\
  \midrule
  \multicolumn{14}{l}{\textit{$f = 4$}} \\
  AEKL & 98.3 & 1.63 & 18.0 & 4.70 & 28.2 & 49.8 & 450 & 62.7 & 8.18 & 20.6 & 21.7 & 23.0 & 22.4 \\
  SoftVQ & 43.7 & 2.07 & 63.8 & 3.07 & 11.9 & 35.5 & 37.9 & 87.1 & 17.1 & 12.2 & 18.9 & 21.5 & 20.2 \\
  VQGAN & 38.2 & 4.13 & 26.2 & 3.15 & 41.2 & 210 & 234 & 61.5 & 10.4 & 25.9 & 26.5 & 20.0 & 26.4 \\
  SimVQ & 41.5 & 3.35 & 15.4 & 4.18 & 7.59 & 83.6 & 45.1 & 51.5 & 8.51 & 9.24 & 14.5 & 24.0 & 14.9 \\
  SimpleQINCo & 55.1 & \textbf{0.799} & 21.4 & 5.62 & \textbf{6.24} & 80.8 & 322 & \textbf{34.6} & 10.2 & 29.7 & 26.3 & 38.1 & 28.0 \\
  RQVAE & 32.0 & 1.45 & 33.9 & \textbf{2.89} & 16.8 & 67.7 & \textbf{6.56} & 82.1 & 9.85 & 23.0 & 28.8 & 52.7 & 25.9 \\
  DiVeQ & 35.9 & 1.91 & \textbf{7.44} & 3.28 & 55.8 & \textbf{29.8} & 79.1 & 85.6 & \textbf{4.53} & \textbf{5.55} & \textbf{10.9} & \textbf{12.1} & \textbf{11.5} \\
  SF-DiVeQ & \textbf{28.5} & 4.33 & 67.8 & 5.14 & 45.4 & 39.4 & 46.3 & 73.4 & 4.61 & 16.4 & 14.2 & 23.7 & 26.1 \\
  BSQ & 37.1 & 13.6 & 20.7 & 13.7 & 111 & 78.3 & 15.6 & 90.3 & 27.8 & 34.2 & 62.3 & 87.3 & 35.6 \\
  LFQ & 69.1 & 5.66 & 64.6 & 3.32 & 47.0 & 126 & 217 & 221 & 32.1 & 51.2 & 50.2 & 60.8 & 56.0 \\
  \midrule
  \multicolumn{14}{l}{\textit{$f = 8$}} \\
  AEKL & 56.1 & 2.10 & 11.6 & 8.98 & 24.8 & \textbf{18.9} & 159 & 34.6 & 5.93 & 29.7 & 12.0 & 18.1 & 18.5 \\
  SoftVQ & 54.4 & 3.15 & 60.4 & 6.07 & 50.2 & 586 & 522 & 72.1 & 16.0 & 20.5 & 27.6 & 25.5 & 38.9 \\
  VQGAN & 104 & \textbf{0.529} & 18.3 & 6.43 & 20.0 & 54.1 & 208 & \textbf{17.7} & 9.27 & 7.92 & 12.6 & 21.9 & 18.0 \\
  SimVQ & 46.1 & 7.44 & 34.5 & 4.09 & 70.7 & 35.4 & 185 & 36.9 & 14.9 & 18.9 & 14.8 & 28.5 & 31.5 \\
  SimpleQINCo & 79.5 & 0.564 & 25.9 & 5.54 & 15.1 & 50.6 & \textbf{54.9} & 39.4 & 6.25 & 17.2 & 13.5 & 21.2 & 19.2 \\
  RQVAE & {\textbf{31.6}} & {5.54} & {40.9} & {\textbf{2.77}} & {6.45} & {58.7} & {76.7} & {26.6} & {\textbf{4.79}} & {\textbf{7.33}} & {\textbf{9.80}} & {\textbf{10.3}} & {10.0} \\
  DiVeQ & 56.5 & 1.40 & \textbf{7.60} & 3.11 & \textbf{2.73} & 221 & 144 & 21.6 & 7.56 & 13.5 & 13.0 & 27.3 & 13.2 \\
  SF-DiVeQ & 64.5 & 4.11 & 85.3 & 3.91 & 49.8 & 22.1 & 89.2 & 43.2 & 8.96 & 10.4 & 12.0 & 11.4 & 17.1 \\
  BSQ & 81.5 & 1.74 & 15.6 & 9.15 & 12.2 & 335 & 199 & 66.4 & 20.2 & 19.9 & 34.0 & 32.5 & 26.4 \\
  LFQ & 54.9 & 4.24 & 24.3 & 4.46 & 19.4 & 453 & 555 & 165 & 27.2 & 32.9 & 44.4 & 68.1 & 38.7 \\
  \midrule
  \multicolumn{14}{l}{\textit{$f = 16$}} \\
  AEKL & 232 & 1.56 & 52.6 & 12.7 & 69.2 & \textbf{19.6} & 226 & 56.8 & 8.09 & 13.5 & 14.3 & 22.4 & 21.0 \\
  SoftVQ & {\textbf{53.6}} & {3.51} & {27.9} & {5.83} & {\textbf{2.81}} & {45.7} & {\textbf{49.0}} & {\textbf{24.6}} & {11.9} & {\textbf{6.48}} & {8.08} & {\textbf{8.39}} & {10.1} \\
  VQGAN & 73.9 & 2.78 & \textbf{6.38} & 16.2 & 25.4 & 77.8 & 405 & 109 & 9.82 & 13.8 & 9.84 & 11.8 & 15.0 \\
  SimVQ & 86.7 & 1.63 & 11.9 & 13.0 & 75.9 & 163 & 467 & 127 & 10.7 & 38.0 & 30.1 & 18.5 & 34.0 \\
  SimpleQINCo & 168 & 0.875 & 78.4 & \textbf{3.95} & 76.1 & 43.2 & 385 & 45.4 & 5.06 & 23.2 & 12.8 & 20.6 & 33.2 \\
  RQVAE & 130 & 1.00 & 17.5 & 6.00 & 44.7 & 133 & 276 & 58.4 & 5.96 & 13.8 & 8.76 & 11.3 & 15.7 \\
  DiVeQ & 206 & 1.02 & 13.5 & 5.86 & 23.7 & 104 & 428 & 25.7 & \textbf{2.90} & 25.4 & 16.7 & 18.6 & 21.2 \\
  SF-DiVeQ & 74.4 & \textbf{0.602} & 10.8 & 8.98 & 3.67 & 38.9 & 323 & 43.5 & 3.13 & 11.7 & \textbf{7.73} & 13.9 & 11.2 \\
  BSQ & 98.2 & 4.61 & 58.8 & 7.34 & 5.07 & 126 & 195 & 62.7 & 10.2 & 11.2 & 20.8 & 21.6 & 21.2 \\
  LFQ & 72.6 & 6.88 & 36.5 & 10.1 & 34.1 & 66.2 & 670 & 142 & 11.1 & 19.1 & 29.7 & 24.7 & 31.9 \\
  \bottomrule
  \end{tabular}}
\end{table}

\section{Qualitative Generations}
\label{sec:supp-qual-gen}
\begin{figure*}[h]
  \centering
  \begin{minipage}{\textwidth}
    \centering
    \includegraphics[width=\textwidth,height=0.27\textheight,keepaspectratio]{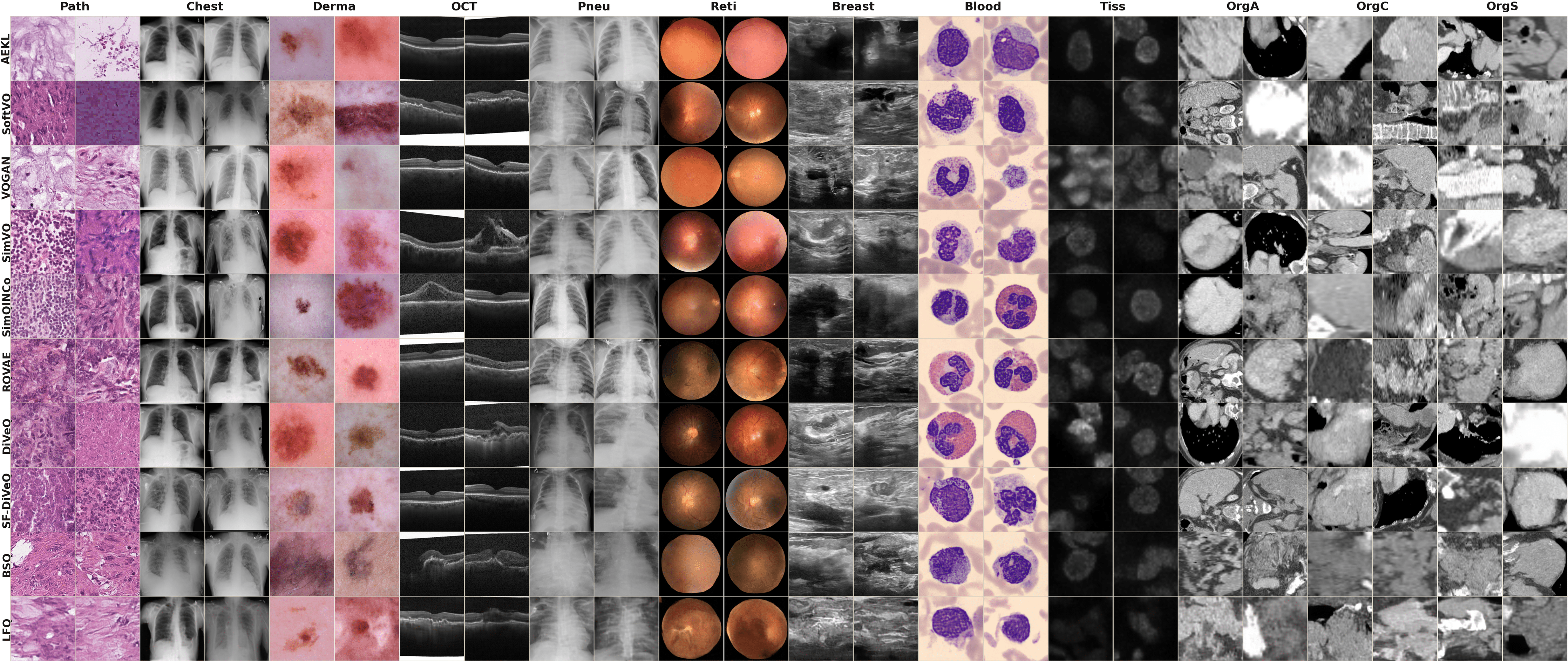}
    \subcaption{$f{=}4$}
    \label{fig:gen-f4}
  \end{minipage}

  \vspace{4pt}
  \begin{minipage}{\textwidth}
    \centering
    \includegraphics[width=\textwidth,height=0.27\textheight,keepaspectratio]{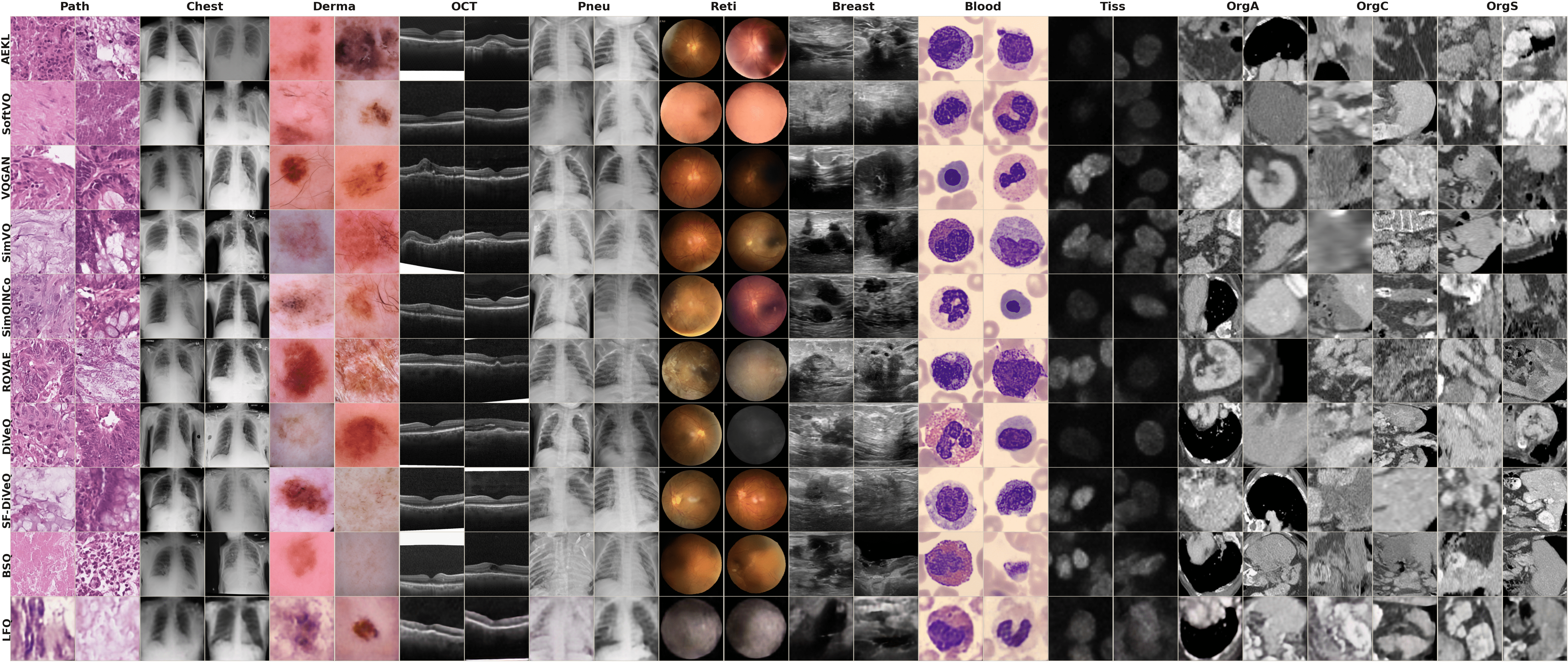}
    \subcaption{$f{=}8$}
    \label{fig:gen-f8}
  \end{minipage}

  \vspace{4pt}
  \begin{minipage}{\textwidth}
    \centering
    \includegraphics[width=\textwidth,height=0.27\textheight,keepaspectratio]{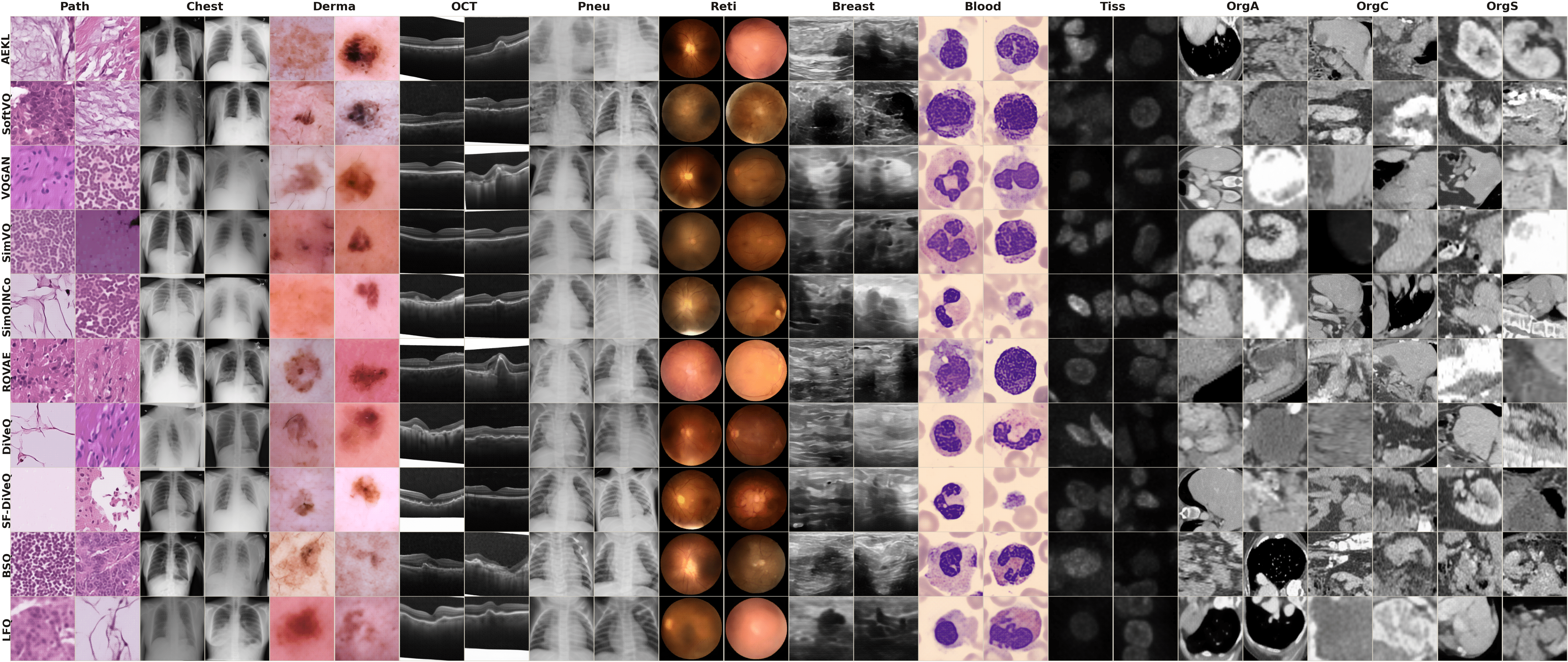}
    \subcaption{$f{=}16$}
    \label{fig:gen-f16}
  \end{minipage}

  \caption{{Qualitative generations of two unconditional DiT samples per dataset (columns) by every tokenizer family (rows), at compression factors \subref{fig:gen-f4}~$f{=}4$, \subref{fig:gen-f8}~$f{=}8$, and \subref{fig:gen-f16}~$f{=}16$. Sample quality and diversity degrade with compression.}}
  \label{fig:gen-qual}
\end{figure*}

\Cref{fig:gen-qual} shows two conditional DiT samples per dataset from every tokenizer family at each compression factor, complementing the quantitative gFID results with a visual sense of how sample quality and diversity degrade as the bottleneck tightens.

\section{Codebook Statistics}
\label{sec:supp-codebook}
We characterise the discrete bottleneck by \emph{Usage} (\cref{tab:usage}), the
fraction of codes ever selected; normalised \emph{Entropy} (\cref{tab:entropy}),
$H(\mathcal{C})/\log_2 K$ of the empirical code distribution; and \emph{Utility}
(\cref{tab:utility}), $1/(1+\mathrm{CV})$ with $\mathrm{CV}=\sigma(p)/\mu(p)$.
Continuous baselines have no codebook and are omitted. Usage saturates near
$100\%$ for most modern schemes yet collapses for VQGAN as $f$ grows---motivating
the latent-geometry analysis, since high usage does not imply an efficiently used
latent.

\begin{table}[h]
  \centering
  \caption{Per-dataset codebook \emph{Usage} ($\uparrow$).}
  \label{tab:usage}
  \setlength{\tabcolsep}{3pt}
  \adjustbox{max width=0.48\textwidth,max totalheight=0.43\textheight,center}{%
  \begin{tabular}{lcccccccccccc|c}
  \toprule
  Method & Path & Chest & Derma & OCT & Pneu & Reti & Breast & Blood & Tiss & OrgA & OrgC & OrgS & mean \\
  \midrule
  \multicolumn{14}{l}{\textit{$f = 4$}} \\
  VQGAN & 0.79 & 0.79 & 0.79 & 0.79 & 0.78 & 0.73 & 0.78 & 0.74 & 0.31 & 0.79 & 0.79 & 0.79 & 0.74 \\
  SimVQ & 1.00 & 1.00 & 0.95 & 0.98 & 0.98 & 0.87 & 0.96 & 0.95 & 0.57 & \textbf{1.00} & 1.00 & 1.00 & 0.94 \\
  SimQINCo & \textbf{1.00} & 1.00 & 0.99 & 0.98 & 0.97 & 0.82 & 0.97 & 0.93 & 0.37 & 1.00 & 1.00 & 1.00 & 0.92 \\
  RQVAE & 0.99 & \textbf{1.00} & 0.96 & 0.99 & 0.97 & 0.92 & 0.97 & 0.92 & 0.60 & 1.00 & \textbf{1.00} & \textbf{1.00} & 0.94 \\
  DiVeQ & 1.00 & 1.00 & 0.99 & 1.00 & 0.99 & 0.96 & 0.99 & 0.96 & 0.83 & 1.00 & 1.00 & 1.00 & 0.98 \\
  SF-DiVeQ & 0.32 & 0.33 & 0.31 & 0.31 & 0.31 & 0.31 & 0.30 & 0.30 & 0.31 & 0.32 & 0.32 & 0.32 & 0.31 \\
  BSQ & \textbf{1.00} & \textbf{1.00} & \textbf{1.00} & \textbf{1.00} & \textbf{1.00} & \textbf{1.00} & \textbf{1.00} & \textbf{1.00} & \textbf{1.00} & \textbf{1.00} & \textbf{1.00} & \textbf{1.00} & 1.00 \\
  LFQ & \textbf{1.00} & \textbf{1.00} & \textbf{1.00} & \textbf{1.00} & \textbf{1.00} & \textbf{1.00} & \textbf{1.00} & \textbf{1.00} & \textbf{1.00} & \textbf{1.00} & \textbf{1.00} & \textbf{1.00} & 1.00 \\
  \midrule
  \multicolumn{14}{l}{\textit{$f = 8$}} \\
  VQGAN & 0.12 & 0.12 & 0.11 & 0.12 & 0.12 & 0.11 & 0.12 & 0.12 & 0.11 & 0.12 & 0.12 & 0.12 & 0.11 \\
  SimVQ & 1.00 & 1.00 & 0.86 & 0.94 & 0.88 & 0.60 & 0.77 & 0.93 & 0.97 & 1.00 & 1.00 & 1.00 & 0.91 \\
  SimQINCo & \textbf{1.00} & 1.00 & 0.92 & 0.93 & 0.93 & 0.77 & 0.92 & 0.98 & 0.84 & 1.00 & 1.00 & 1.00 & 0.94 \\
  RQVAE & 0.97 & 1.00 & 0.88 & 0.96 & 0.91 & 0.63 & 0.89 & 0.95 & 0.70 & \textbf{1.00} & \textbf{1.00} & \textbf{1.00} & 0.91 \\
  DiVeQ & 1.00 & \textbf{1.00} & 0.94 & 0.98 & 0.96 & 0.80 & 0.94 & 0.99 & 1.00 & \textbf{1.00} & 1.00 & 1.00 & 0.97 \\
  SF-DiVeQ & 1.00 & 1.00 & 1.00 & 1.00 & 1.00 & 0.98 & 0.99 & 1.00 & 1.00 & 1.00 & 1.00 & 1.00 & 1.00 \\
  BSQ & \textbf{1.00} & \textbf{1.00} & \textbf{1.00} & \textbf{1.00} & \textbf{1.00} & 1.00 & \textbf{1.00} & \textbf{1.00} & \textbf{1.00} & \textbf{1.00} & \textbf{1.00} & \textbf{1.00} & 1.00 \\
  LFQ & \textbf{1.00} & \textbf{1.00} & \textbf{1.00} & \textbf{1.00} & \textbf{1.00} & \textbf{1.00} & 1.00 & \textbf{1.00} & \textbf{1.00} & \textbf{1.00} & \textbf{1.00} & \textbf{1.00} & 1.00 \\
  \midrule
  \multicolumn{14}{l}{\textit{$f = 16$}} \\
  VQGAN & 0.01 & 0.01 & 0.01 & 0.01 & 0.01 & 0.01 & 0.01 & 0.01 & 0.01 & 0.01 & 0.01 & 0.01 & 0.01 \\
  SimVQ & 0.98 & 0.99 & 0.65 & 0.44 & 0.56 & 0.33 & 0.29 & 0.93 & 0.98 & 1.00 & 1.00 & 1.00 & 0.76 \\
  SimQINCo & \textbf{1.00} & 0.99 & 0.79 & 0.77 & 0.81 & 0.57 & 0.70 & 0.95 & 0.93 & \textbf{1.00} & 1.00 & 1.00 & 0.88 \\
  RQVAE & 0.97 & 0.98 & 0.77 & 0.76 & 0.78 & 0.47 & 0.69 & 0.94 & 0.92 & 0.98 & 0.98 & 0.97 & 0.85 \\
  DiVeQ & 1.00 & \textbf{1.00} & \textbf{0.95} & \textbf{0.95} & \textbf{0.89} & \textbf{0.87} & \textbf{0.78} & \textbf{1.00} & \textbf{1.00} & \textbf{1.00} & \textbf{1.00} & \textbf{1.00} & 0.95 \\
  SF-DiVeQ & 0.62 & 0.62 & 0.59 & 0.59 & 0.58 & 0.49 & 0.52 & 0.61 & 0.63 & 0.63 & 0.62 & 0.62 & 0.59 \\
  BSQ & 0.98 & 1.00 & 0.65 & 0.49 & 0.34 & 0.16 & 0.11 & 0.81 & \textbf{1.00} & 1.00 & 0.99 & 0.99 & 0.71 \\
  LFQ & 0.47 & 0.45 & 0.30 & 0.35 & 0.20 & 0.08 & 0.09 & 0.34 & 0.78 & 0.57 & 0.52 & 0.53 & 0.39 \\
  \bottomrule
  \end{tabular}}
\end{table}

\begin{table}[h]
  \centering
  \caption{Per-dataset codebook \emph{Entropy} ($\uparrow$).}
  \label{tab:entropy}
  \setlength{\tabcolsep}{3pt}
  \adjustbox{max width=0.48\textwidth,max totalheight=0.43\textheight,center}{%
  \begin{tabular}{lcccccccccccc|c}
  \toprule
  Method & Path & Chest & Derma & OCT & Pneu & Reti & Breast & Blood & Tiss & OrgA & OrgC & OrgS & mean \\
  \midrule
  \multicolumn{14}{l}{\textit{$f = 4$}} \\
  VQGAN & 0.95 & 0.90 & 0.88 & 0.92 & 0.90 & 0.85 & 0.93 & 0.83 & 0.71 & 0.88 & 0.88 & 0.88 & 0.88 \\
  SimVQ & 0.95 & 0.93 & 0.89 & 0.92 & 0.92 & 0.86 & 0.94 & 0.87 & 0.75 & 0.93 & 0.93 & 0.93 & 0.90 \\
  SimQINCo & 0.97 & 0.93 & 0.90 & 0.94 & 0.94 & 0.85 & 0.96 & 0.86 & 0.69 & 0.91 & 0.91 & 0.91 & 0.90 \\
  RQVAE & 0.71 & 0.69 & 0.69 & 0.71 & 0.70 & 0.69 & 0.71 & 0.68 & 0.63 & 0.69 & 0.69 & 0.69 & 0.69 \\
  DiVeQ & 0.95 & 0.93 & 0.91 & 0.96 & 0.93 & 0.86 & 0.97 & 0.89 & 0.86 & 0.93 & 0.93 & 0.93 & 0.92 \\
  SF-DiVeQ & 0.83 & 0.83 & 0.77 & 0.83 & 0.82 & 0.80 & 0.85 & 0.77 & 0.81 & 0.82 & 0.83 & 0.83 & 0.82 \\
  BSQ & \textbf{1.00} & \textbf{1.00} & 0.99 & \textbf{0.99} & \textbf{1.00} & 0.99 & 1.00 & \textbf{0.99} & 0.99 & 0.99 & 0.99 & 0.99 & 0.99 \\
  LFQ & 1.00 & 0.99 & \textbf{0.99} & 0.99 & 0.99 & \textbf{0.99} & \textbf{1.00} & 0.99 & \textbf{0.99} & \textbf{1.00} & \textbf{1.00} & \textbf{1.00} & 0.99 \\
  \midrule
  \multicolumn{14}{l}{\textit{$f = 8$}} \\
  VQGAN & 0.74 & 0.74 & 0.69 & 0.71 & 0.74 & 0.67 & 0.75 & 0.70 & 0.66 & 0.76 & 0.76 & 0.76 & 0.72 \\
  SimVQ & 0.86 & 0.78 & 0.75 & 0.86 & 0.77 & 0.78 & 0.87 & 0.78 & 0.69 & 0.82 & 0.81 & 0.81 & 0.80 \\
  SimQINCo & 0.97 & 0.96 & 0.91 & 0.93 & 0.95 & 0.88 & 0.96 & 0.92 & 0.85 & 0.97 & 0.97 & 0.97 & 0.94 \\
  RQVAE & 0.77 & 0.75 & 0.74 & 0.76 & 0.75 & 0.73 & 0.76 & 0.74 & 0.68 & 0.75 & 0.76 & 0.75 & 0.74 \\
  DiVeQ & 0.97 & 0.96 & 0.92 & 0.95 & 0.95 & 0.90 & 0.97 & 0.93 & 0.94 & 0.97 & 0.97 & 0.97 & 0.95 \\
  SF-DiVeQ & 0.99 & 0.98 & 0.97 & 0.99 & 0.98 & 0.95 & 0.98 & 0.96 & 0.93 & 0.98 & 0.98 & 0.98 & 0.97 \\
  BSQ & \textbf{1.00} & \textbf{1.00} & \textbf{0.99} & \textbf{0.99} & \textbf{0.99} & 0.97 & \textbf{0.99} & \textbf{0.99} & \textbf{1.00} & \textbf{1.00} & \textbf{1.00} & \textbf{1.00} & 0.99 \\
  LFQ & 1.00 & 0.99 & 0.99 & 0.99 & 0.99 & \textbf{0.97} & 0.99 & 0.99 & 1.00 & 1.00 & 1.00 & 1.00 & 0.99 \\
  \midrule
  \multicolumn{14}{l}{\textit{$f = 16$}} \\
  VQGAN & 0.53 & 0.51 & 0.48 & 0.48 & 0.51 & 0.48 & 0.53 & 0.51 & 0.48 & 0.54 & 0.54 & 0.54 & 0.51 \\
  SimVQ & 0.68 & 0.54 & 0.51 & 0.45 & 0.54 & 0.49 & 0.58 & 0.64 & 0.43 & 0.62 & 0.62 & 0.62 & 0.56 \\
  SimQINCo & 0.98 & 0.95 & 0.90 & 0.91 & 0.94 & 0.88 & 0.94 & 0.94 & 0.91 & 0.98 & 0.98 & 0.98 & 0.94 \\
  RQVAE & 0.87 & 0.84 & 0.82 & 0.82 & 0.84 & 0.81 & 0.85 & 0.84 & 0.81 & 0.87 & 0.86 & 0.86 & 0.84 \\
  DiVeQ & 0.98 & 0.97 & \textbf{0.95} & \textbf{0.95} & \textbf{0.96} & \textbf{0.95} & \textbf{0.96} & 0.96 & 0.97 & 0.98 & 0.98 & 0.98 & 0.96 \\
  SF-DiVeQ & 0.93 & 0.92 & 0.89 & 0.90 & 0.92 & 0.86 & 0.91 & 0.90 & 0.90 & 0.93 & 0.93 & 0.93 & 0.91 \\
  BSQ & \textbf{0.98} & \textbf{0.99} & 0.94 & 0.93 & 0.90 & 0.80 & 0.82 & \textbf{0.96} & \textbf{0.99} & \textbf{0.99} & \textbf{0.99} & \textbf{0.99} & 0.94 \\
  LFQ & 0.90 & 0.85 & 0.85 & 0.89 & 0.85 & 0.71 & 0.79 & 0.84 & 0.91 & 0.91 & 0.90 & 0.91 & 0.86 \\
  \bottomrule
  \end{tabular}}
\end{table}

\begin{table}[h]
  \centering
  \caption{Per-dataset codebook \emph{Utility} ($\uparrow$).}
  \label{tab:utility}
  \setlength{\tabcolsep}{3pt}
  \adjustbox{max width=0.48\textwidth,max totalheight=0.43\textheight,center}{%
  \begin{tabular}{lcccccccccccc|c}
  \toprule
  Method & Path & Chest & Derma & OCT & Pneu & Reti & Breast & Blood & Tiss & OrgA & OrgC & OrgS & mean \\
  \midrule
  \multicolumn{14}{l}{\textit{$f = 4$}} \\
  VQGAN & 0.49 & 0.38 & 0.36 & 0.40 & 0.40 & 0.25 & 0.47 & 0.24 & 0.19 & 0.30 & 0.31 & 0.31 & 0.34 \\
  SimVQ & 0.47 & 0.42 & 0.37 & 0.40 & 0.41 & 0.28 & 0.45 & 0.29 & 0.21 & 0.37 & 0.40 & 0.40 & 0.37 \\
  SimQINCo & 0.57 & 0.45 & 0.40 & 0.44 & 0.48 & 0.26 & 0.55 & 0.27 & 0.18 & 0.33 & 0.35 & 0.35 & 0.39 \\
  RQVAE & 0.11 & 0.10 & 0.10 & 0.11 & 0.11 & 0.11 & 0.11 & 0.10 & 0.09 & 0.10 & 0.10 & 0.10 & 0.10 \\
  DiVeQ & 0.50 & 0.44 & 0.39 & 0.53 & 0.42 & 0.27 & 0.56 & 0.37 & 0.34 & 0.41 & 0.44 & 0.43 & 0.43 \\
  SF-DiVeQ & 0.32 & 0.31 & 0.25 & 0.31 & 0.30 & 0.25 & 0.35 & 0.20 & 0.27 & 0.28 & 0.30 & 0.30 & 0.29 \\
  BSQ & \textbf{0.81} & \textbf{0.77} & 0.72 & \textbf{0.77} & \textbf{0.79} & \textbf{0.68} & 0.80 & \textbf{0.75} & 0.73 & 0.79 & 0.73 & 0.71 & 0.75 \\
  LFQ & 0.80 & 0.74 & \textbf{0.75} & 0.76 & 0.75 & 0.67 & \textbf{0.82} & 0.71 & \textbf{0.78} & \textbf{0.81} & \textbf{0.81} & \textbf{0.81} & 0.77 \\
  \midrule
  \multicolumn{14}{l}{\textit{$f = 8$}} \\
  VQGAN & 0.21 & 0.21 & 0.15 & 0.17 & 0.20 & 0.14 & 0.23 & 0.15 & 0.14 & 0.23 & 0.23 & 0.23 & 0.19 \\
  SimVQ & 0.18 & 0.08 & 0.09 & 0.18 & 0.08 & 0.15 & 0.20 & 0.09 & 0.05 & 0.09 & 0.09 & 0.09 & 0.11 \\
  SimQINCo & 0.55 & 0.50 & 0.39 & 0.41 & 0.48 & 0.30 & 0.53 & 0.36 & 0.28 & 0.52 & 0.53 & 0.52 & 0.45 \\
  RQVAE & 0.13 & 0.12 & 0.13 & 0.13 & 0.12 & 0.14 & 0.13 & 0.12 & 0.10 & 0.12 & 0.12 & 0.12 & 0.12 \\
  DiVeQ & 0.57 & 0.51 & 0.41 & 0.48 & 0.49 & 0.30 & 0.56 & 0.36 & 0.44 & 0.52 & 0.55 & 0.55 & 0.48 \\
  SF-DiVeQ & 0.74 & 0.61 & 0.53 & 0.70 & 0.61 & 0.48 & 0.64 & 0.47 & 0.40 & 0.56 & 0.56 & 0.55 & 0.57 \\
  BSQ & \textbf{0.82} & \textbf{0.78} & \textbf{0.71} & \textbf{0.75} & \textbf{0.75} & \textbf{0.51} & \textbf{0.69} & \textbf{0.71} & \textbf{0.81} & \textbf{0.84} & \textbf{0.82} & \textbf{0.81} & 0.75 \\
  LFQ & 0.79 & 0.68 & 0.70 & 0.70 & 0.69 & 0.51 & 0.66 & 0.66 & 0.78 & 0.79 & 0.77 & 0.76 & 0.71 \\
  \midrule
  \multicolumn{14}{l}{\textit{$f = 16$}} \\
  VQGAN & 0.09 & 0.08 & 0.06 & 0.06 & 0.08 & 0.06 & 0.09 & 0.07 & 0.06 & 0.09 & 0.09 & 0.09 & 0.08 \\
  SimVQ & 0.04 & 0.02 & 0.02 & 0.02 & 0.03 & 0.02 & 0.03 & 0.03 & 0.02 & 0.03 & 0.03 & 0.03 & 0.03 \\
  SimQINCo & \textbf{0.62} & 0.49 & 0.36 & 0.38 & 0.47 & 0.34 & 0.50 & 0.41 & 0.39 & 0.65 & 0.61 & 0.61 & 0.49 \\
  RQVAE & 0.27 & 0.23 & 0.24 & 0.22 & 0.23 & 0.24 & 0.25 & 0.23 & 0.20 & 0.26 & 0.25 & 0.25 & 0.24 \\
  DiVeQ & 0.59 & 0.54 & \textbf{0.48} & \textbf{0.50} & \textbf{0.51} & \textbf{0.42} & \textbf{0.53} & \textbf{0.47} & 0.56 & 0.61 & 0.60 & 0.61 & 0.54 \\
  SF-DiVeQ & 0.48 & 0.45 & 0.38 & 0.39 & 0.44 & 0.30 & 0.43 & 0.37 & 0.37 & 0.47 & 0.47 & 0.47 & 0.42 \\
  BSQ & 0.61 & \textbf{0.60} & 0.42 & 0.43 & 0.37 & 0.15 & 0.25 & 0.46 & \textbf{0.72} & \textbf{0.67} & \textbf{0.63} & \textbf{0.63} & 0.50 \\
  LFQ & 0.32 & 0.18 & 0.22 & 0.30 & 0.25 & 0.09 & 0.19 & 0.18 & 0.31 & 0.31 & 0.32 & 0.32 & 0.25 \\
  \bottomrule
  \end{tabular}}
\end{table}

\section{Linear Probing}
\label{sec:supp-probing}
We fit a linear classifier on the frozen pre-quantization latent $\mathbf{z}_e$
(\cref{tab:probe}) and repeat it on the post-quantization latent $z_q$
(\cref{tab:probe-zq}), each under the per-dataset task-appropriate primary metric
(top-1 accuracy for multi-class; macro-AUROC for multi-label ChestMNIST;
quadratic-weighted $\kappa$ for ordinal RetinaMNIST). The near-identical means
between $z_e$ and $z_q$ confirm that discretization is largely
information-preserving, except for the lookup-free families (LFQ, BSQ); the
continuous AEKL and SoftVQ autoencoders have no discrete bottleneck and are
omitted from \cref{tab:probe-zq}.

\begin{table}[h]
  \centering
  \caption{Per-dataset linear-probing primary metric ($\uparrow$, on $z_e$).}
  \label{tab:probe}
  \setlength{\tabcolsep}{3pt}
  \adjustbox{max width=0.48\textwidth,max totalheight=0.43\textheight,center}{%
  \begin{tabular}{lcccccccccccc|c}
  \toprule
  Method & Path & Chest & Derma & OCT & Pneu & Reti & Breast & Blood & Tiss & OrgA & OrgC & OrgS & mean \\
  \midrule
  \multicolumn{14}{l}{\textit{$f = 4$}} \\
  AEKL & .789 & .584 & .682 & .721 & .979 & .581 & .774 & .934 & .686 & .989 & .984 & .934 & .803 \\
  SoftVQ & \textbf{.921} & \textbf{.610} & .781 & .828 & \textbf{.989} & .587 & .863 & \textbf{.974} & .780 & .996 & .993 & \textbf{.979} & .858 \\
  VQGAN & .865 & .609 & .775 & .820 & .984 & .606 & \textbf{.880} & .961 & \textbf{.807} & .990 & .987 & .963 & .854 \\
  SimVQ & .799 & .603 & .718 & .786 & .984 & .596 & .848 & .963 & .802 & .989 & .983 & .955 & .836 \\
  SimQINCo & .858 & .606 & \textbf{.782} & .791 & .986 & .608 & .867 & .961 & .803 & .988 & .985 & .961 & .850 \\
  RQVAE & .865 & .597 & .755 & .780 & .985 & .635 & .845 & .963 & .788 & .991 & .985 & .968 & .847 \\
  DiVeQ & .856 & .608 & .737 & .794 & .987 & .638 & .867 & .969 & .802 & .992 & .987 & .965 & .850 \\
  SF-DiVeQ & .847 & .603 & .703 & .787 & .983 & .526 & .871 & .959 & .806 & .991 & .986 & .966 & .836 \\
  BSQ & .911 & .599 & .728 & \textbf{.853} & .985 & .516 & .868 & .967 & .764 & .995 & .993 & .971 & .846 \\
  LFQ & .902 & .573 & .682 & .840 & .988 & \textbf{.651} & .819 & .961 & .752 & \textbf{.997} & \textbf{.995} & .973 & .844 \\
  \midrule
  \multicolumn{14}{l}{\textit{$f = 8$}} \\
  AEKL & .854 & .612 & .688 & .776 & .978 & .614 & .723 & .933 & .764 & .988 & .976 & .949 & .821 \\
  SoftVQ & \textbf{.938} & .649 & .776 & \textbf{.849} & .978 & .621 & .695 & .968 & \textbf{.815} & .995 & .993 & .977 & .854 \\
  VQGAN & .896 & \textbf{.656} & .770 & .801 & .983 & .596 & .881 & .946 & .799 & .990 & .982 & .956 & .855 \\
  SimVQ & .835 & .633 & .764 & .835 & .984 & .652 & .831 & .967 & .801 & .993 & .988 & .967 & .854 \\
  SimQINCo & .880 & .639 & .761 & .823 & .983 & \textbf{.667} & .851 & .954 & .800 & .988 & .979 & .954 & .857 \\
  RQVAE & .875 & .626 & .734 & .782 & .978 & .615 & .859 & .952 & .802 & .993 & .991 & .964 & .848 \\
  DiVeQ & .856 & .640 & .715 & .829 & .982 & .580 & .847 & .963 & .795 & .994 & .988 & .971 & .847 \\
  SF-DiVeQ & .906 & .642 & \textbf{.785} & .849 & \textbf{.988} & .653 & \textbf{.904} & \textbf{.977} & .811 & \textbf{.997} & \textbf{.994} & \textbf{.978} & .874 \\
  BSQ & .874 & .622 & .710 & .837 & .986 & .596 & .799 & .939 & .793 & .993 & .994 & .968 & .843 \\
  LFQ & .887 & .616 & .709 & .803 & .981 & .632 & .851 & .935 & .781 & .992 & .991 & .969 & .846 \\
  \midrule
  \multicolumn{14}{l}{\textit{$f = 16$}} \\
  AEKL & .877 & .643 & .750 & .796 & .973 & \textbf{.669} & .836 & .939 & .762 & .987 & .982 & .948 & .847 \\
  SoftVQ & .921 & .663 & \textbf{.816} & \textbf{.876} & .979 & .662 & .778 & \textbf{.975} & .806 & .997 & .994 & .979 & .870 \\
  VQGAN & .904 & .678 & .778 & .834 & .983 & .661 & .876 & .955 & .781 & .986 & .977 & .946 & .863 \\
  SimVQ & .896 & .664 & .816 & .860 & \textbf{.990} & .636 & \textbf{.917} & .961 & .797 & .992 & .989 & .966 & .874 \\
  SimQINCo & .901 & .672 & .792 & .812 & .982 & .639 & .895 & .959 & .775 & .990 & .979 & .954 & .862 \\
  RQVAE & .902 & .657 & .790 & .805 & .982 & .625 & .851 & .957 & .780 & .990 & .987 & .966 & .858 \\
  DiVeQ & \textbf{.927} & .673 & .778 & .800 & .987 & .658 & .822 & .958 & .782 & .990 & .983 & .964 & .860 \\
  SF-DiVeQ & .913 & \textbf{.679} & .804 & .840 & .989 & .638 & .819 & .968 & .800 & .993 & .990 & .971 & .867 \\
  BSQ & .925 & .659 & .756 & .876 & .989 & .572 & .779 & .954 & \textbf{.830} & .994 & .994 & .970 & .858 \\
  LFQ & .877 & .657 & .745 & .817 & .980 & .654 & .855 & .959 & .826 & \textbf{.998} & \textbf{.997} & \textbf{.982} & .862 \\
  \bottomrule
  \end{tabular}}
\end{table}

\begin{table}[h]
  \centering
  \caption{Per-dataset linear-probing primary metric ($\uparrow$, on $z_q$).}
  \label{tab:probe-zq}
  \setlength{\tabcolsep}{3pt}
  \adjustbox{max width=0.48\textwidth,max totalheight=0.43\textheight,center}{%
  \begin{tabular}{lcccccccccccc|c}
  \toprule
  Method & Path & Chest & Derma & OCT & Pneu & Reti & Breast & Blood & Tiss & OrgA & OrgC & OrgS & mean \\
  \midrule
  \multicolumn{14}{l}{\textit{$f = 4$}} \\
  AEKL & -- & -- & -- & -- & -- & -- & -- & -- & -- & -- & -- & -- & -- \\
  SoftVQ & -- & -- & -- & -- & -- & -- & -- & -- & -- & -- & -- & -- & -- \\
  VQGAN & .864 & .607 & .775 & .820 & .984 & .610 & \textbf{.882} & .960 & .804 & .990 & .987 & .963 & .854 \\
  SimVQ & .797 & .603 & .718 & .785 & .984 & .596 & .850 & .963 & .800 & .989 & .983 & .955 & .835 \\
  SimQINCo & .858 & .605 & \textbf{.785} & .791 & .986 & .609 & .865 & .961 & .800 & .988 & .985 & .960 & .849 \\
  RQVAE & .865 & .598 & .755 & .780 & .985 & .635 & .845 & .963 & .788 & .991 & .985 & \textbf{.968} & .847 \\
  DiVeQ & .856 & \textbf{.608} & .738 & .794 & \textbf{.987} & \textbf{.663} & .866 & \textbf{.969} & .802 & .992 & .987 & .965 & .852 \\
  SF-DiVeQ & .846 & .603 & .705 & .785 & .983 & .534 & .870 & .959 & \textbf{.805} & .991 & .986 & .966 & .836 \\
  BSQ & \textbf{.875} & .595 & .700 & .793 & .952 & .497 & .815 & .935 & .747 & .988 & .982 & .949 & .819 \\
  LFQ & .851 & .571 & .660 & \textbf{.821} & .985 & .571 & .792 & .950 & .745 & \textbf{.995} & \textbf{.990} & .966 & .825 \\
  \midrule
  \multicolumn{14}{l}{\textit{$f = 8$}} \\
  AEKL & -- & -- & -- & -- & -- & -- & -- & -- & -- & -- & -- & -- & -- \\
  SoftVQ & -- & -- & -- & -- & -- & -- & -- & -- & -- & -- & -- & -- & -- \\
  VQGAN & .891 & \textbf{.656} & .769 & .799 & .983 & .581 & .883 & .946 & .798 & .990 & .982 & .956 & .853 \\
  SimVQ & .835 & .632 & .764 & .831 & .983 & .664 & .820 & .966 & .799 & .993 & .988 & .966 & .853 \\
  SimQINCo & .878 & .639 & .759 & .821 & .983 & \textbf{.672} & .852 & .953 & .798 & .988 & .979 & .954 & .856 \\
  RQVAE & .875 & .626 & .735 & .782 & .978 & .615 & .858 & .952 & .802 & .993 & .991 & .964 & .848 \\
  DiVeQ & .858 & .640 & .713 & .828 & .982 & .543 & .843 & .962 & .793 & .994 & .988 & .971 & .843 \\
  SF-DiVeQ & \textbf{.906} & .643 & \textbf{.782} & \textbf{.847} & \textbf{.988} & .648 & \textbf{.896} & \textbf{.977} & \textbf{.808} & \textbf{.997} & \textbf{.994} & \textbf{.978} & .872 \\
  BSQ & .839 & .604 & .690 & .757 & .985 & .628 & .784 & .936 & .748 & .991 & .991 & .965 & .827 \\
  LFQ & .850 & .611 & .683 & .772 & .978 & .615 & .851 & .920 & .760 & .989 & .987 & .959 & .831 \\
  \midrule
  \multicolumn{14}{l}{\textit{$f = 16$}} \\
  AEKL & -- & -- & -- & -- & -- & -- & -- & -- & -- & -- & -- & -- & -- \\
  SoftVQ & -- & -- & -- & -- & -- & -- & -- & -- & -- & -- & -- & -- & -- \\
  VQGAN & .889 & .675 & .772 & .823 & .982 & \textbf{.680} & .875 & .954 & .776 & .986 & .977 & .946 & .861 \\
  SimVQ & .890 & .660 & .807 & \textbf{.839} & \textbf{.990} & .584 & \textbf{.914} & .957 & .788 & .992 & .988 & .964 & .864 \\
  SimQINCo & .898 & .670 & .792 & .808 & .983 & .647 & .901 & .958 & .773 & .990 & .978 & .955 & .863 \\
  RQVAE & .902 & .657 & .790 & .804 & .982 & .623 & .849 & .957 & .779 & .990 & .987 & .966 & .857 \\
  DiVeQ & \textbf{.926} & .671 & .774 & .797 & .986 & .664 & .827 & .958 & .781 & .990 & .983 & .964 & .860 \\
  SF-DiVeQ & .912 & \textbf{.678} & \textbf{.807} & .838 & .989 & .625 & .830 & \textbf{.967} & .798 & .993 & .990 & .971 & .866 \\
  BSQ & .882 & .623 & .722 & .788 & .989 & .665 & .811 & .948 & .789 & \textbf{.994} & \textbf{.993} & \textbf{.972} & .848 \\
  LFQ & .832 & .660 & .724 & .765 & .978 & .569 & .806 & .937 & \textbf{.804} & .994 & .992 & .971 & .836 \\
  \bottomrule
  \end{tabular}}
\end{table}

\begin{table}[t]
  \centering
  \caption{Downstream probing gap $\Delta=\mathrm{M}(z_e)-\mathrm{M}(z_q)$ (\cref{sec:exp-downstream}), $\text{mean}_{\pm\sigma}$ over the twelve datasets (each under its primary metric), per compression factor. Larger $\Delta$ means discretization erodes more class-discriminative structure; \textbf{bold} marks the smallest gap per column.}
  \label{tab:probe-gap}
  \setlength{\tabcolsep}{4pt}
  \adjustbox{max width=0.48\textwidth,max totalheight=0.43\textheight,center}{%
  \begin{tabular}{lccc}
  \toprule
  Method & $f{=}4$ & $f{=}8$ & $f{=}16$ \\
  \midrule
  VQGAN~\cite{vqgan} & 0.000$_{\pm 0.009}$ & 0.008$_{\pm 0.025}$ & 0.001$_{\pm 0.014}$ \\
  SimVQ~\cite{simvq} & \textbf{-0.002$_{\pm 0.006}$} & 0.001$_{\pm 0.017}$ & 0.008$_{\pm 0.006}$ \\
  SimQINCo~\cite{qinco} & 0.002$_{\pm 0.006}$ & -0.001$_{\pm 0.006}$ & -0.001$_{\pm 0.010}$ \\
  RQVAE~\cite{rqvae} & 0.000$_{\pm 0.000}$ & 0.000$_{\pm 0.000}$ & -0.001$_{\pm 0.006}$ \\
  DiVeQ~\cite{diveq} & 0.001$_{\pm 0.006}$ & \textbf{-0.002$_{\pm 0.011}$} & -0.003$_{\pm 0.009}$ \\
  SF-DiVeQ~\cite{diveq} & -0.001$_{\pm 0.005}$ & 0.003$_{\pm 0.009}$ & \textbf{-0.004$_{\pm 0.017}$} \\
  BSQ~\cite{bsq} & 0.046$_{\pm 0.033}$ & 0.021$_{\pm 0.013}$ & 0.018$_{\pm 0.029}$ \\
  LFQ~\cite{lfq} & 0.031$_{\pm 0.029}$ & 0.027$_{\pm 0.021}$ & 0.037$_{\pm 0.024}$ \\
  \bottomrule
  \end{tabular}}
\end{table}

\section{Memorization}
\label{sec:supp-mem}
\begin{figure}[h]
    \centering
    \includegraphics[width=\linewidth]{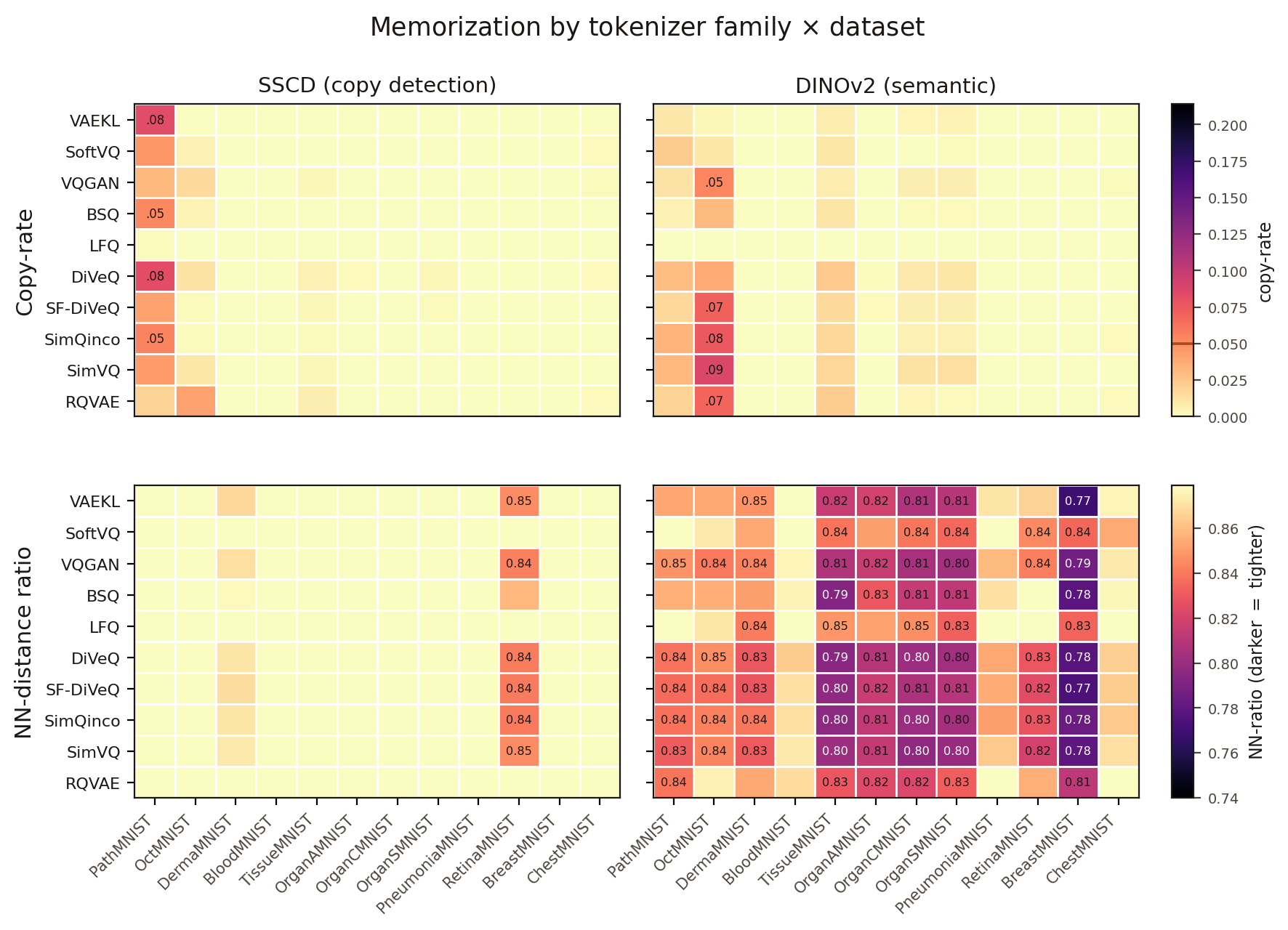}
    \caption{\textbf{Memorization concentrates by dataset, not architecture.}
    Per-(family, dataset) copy-rate at $f{=}8$ under SSCD (left) and DINOv2
    (right) features; rows are quantizer families, columns datasets. Elevated
    values cluster in the lowest-variance modalities (OCT, Path) and are largely
    constant across families, indicating the dataset regime governs replication.}
    \label{fig:supp-memorization_heatmap}
\end{figure}

We test training-set replication with a nearest-neighbour copy detector under two
feature spaces: SSCD, purpose-built for copy detection
(\cref{tab:mem-copy-rate-sscd,tab:mem-mem-ratio-sscd}), and DINOv2, which captures
higher-level semantic similarity and validates the SSCD analysis in a richer space
(\cref{tab:mem-copy-rate-dinov2,tab:mem-mem-ratio-dinov2}). Copy-rates are near
zero almost everywhere, concentrating only in the lowest-variance modalities
(OCT, Path); \cref{fig:supp-memorization_heatmap} shows this is driven by the dataset,
not the quantizer.

\begin{figure}[h]
    \centering
    \includegraphics[width=\linewidth]{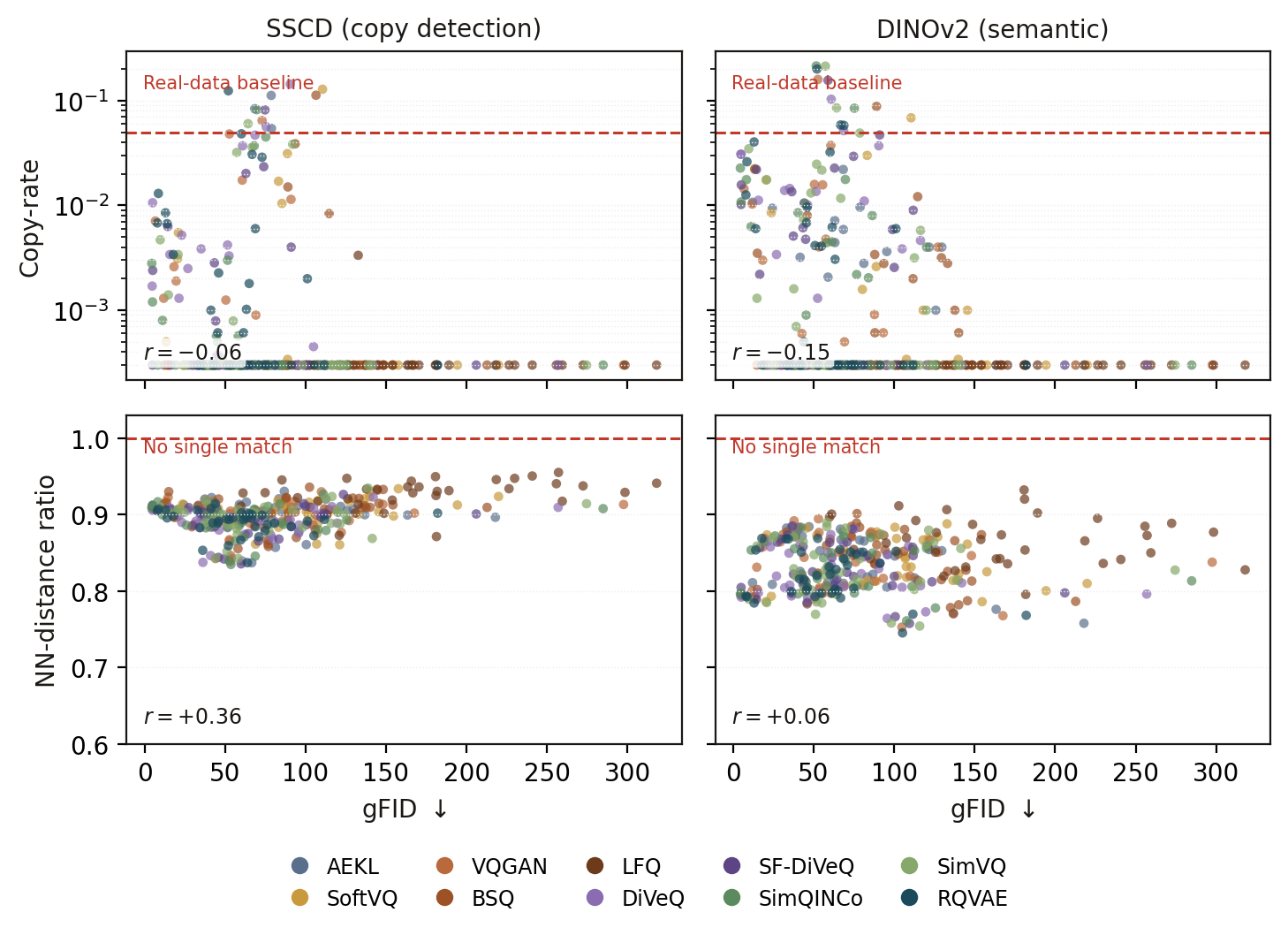}
    \caption{\textbf{Memorization is not a generator-underfitting artifact.} Copy-rate and NN-distance ratio vs.\ gFID (SSCD, DINOv2), colored by family. Copy-rate is decoupled from gFID ($r{=}{-}0.05/{-}0.15$) and elevated copies come from the best (low-gFID) generators; NN-distance ratios stay near one throughout.}
\label{fig:mem-vs-gfid}
\end{figure}

\begin{table}[t]
  \centering
  \caption{Per-dataset variance/diversity of the reference images (grayscale,
  up to 1{,}200 samples each), sorted from least to most diverse by effective
  rank. \emph{Eff.\ rank}: effective number of PCA directions (trace$/\lambda_{\max}$
  of the sample Gram matrix); \emph{NN sim}: mean nearest-neighbour cosine
  similarity (higher $=$ more near-duplicates); \emph{Pairwise dist}: mean
  pairwise cosine distance (lower $=$ more homogeneous).}
  \label{tab:dataset-variance}
  \adjustbox{max width=0.48\textwidth,max totalheight=0.43\textheight,center}{%
  \begin{tabular}{lcccc}
  \toprule
  Dataset & Pixel std $\downarrow$ & Eff.\ rank $\downarrow$ & NN sim $\uparrow$ & Pairwise dist $\downarrow$ \\
  \midrule
  Reti   & 0.214 & 1.7 & 0.986 & 0.086 \\
  Path   & 0.194 & 2.3 & 0.976 & 0.047 \\
  Derma  & 0.146 & 2.5 & 0.993 & 0.029 \\
  OrgS   & 0.267 & 2.8 & 0.938 & 0.215 \\
  OCT    & 0.233 & 3.0 & 0.909 & 0.421 \\
  OrgA   & 0.278 & 3.0 & 0.945 & 0.251 \\
  Pneu   & 0.178 & 3.5 & 0.984 & 0.044 \\
  OrgC   & 0.262 & 3.4 & 0.934 & 0.216 \\
  Chest  & 0.247 & 3.6 & 0.974 & 0.092 \\
  Breast & 0.218 & 3.6 & 0.922 & 0.164 \\
  Tiss   & 0.100 & 3.8 & 0.891 & 0.285 \\
  Blood  & 0.191 & 3.9 & 0.987 & 0.033 \\
  \bottomrule
  \end{tabular}}
\end{table}

\begin{table}[h]
  \centering
  \caption{Per-dataset copy rate ($\downarrow$), SSCD features.}
  \label{tab:mem-copy-rate-sscd}
  \setlength{\tabcolsep}{3pt}
  \adjustbox{max width=0.48\textwidth,max totalheight=0.43\textheight,center}{%
  \begin{tabular}{lcccccccccccc|c}
  \toprule
  Method & Path & Chest & Derma & OCT & Pneu & Reti & Breast & Blood & Tiss & OrgA & OrgC & OrgS & mean \\
  \midrule
  \multicolumn{14}{l}{\textit{$f = 4$}} \\
  AEKL & .084 & \textbf{.000} & \textbf{.000} & \textbf{.000} & \textbf{.000} & \textbf{.000} & \textbf{.000} & \textbf{.000} & \textbf{.000} & \textbf{.000} & \textbf{.000} & \textbf{.000} & .007 \\
  SoftVQ & .128 & .005 & \textbf{.000} & .017 & \textbf{.000} & \textbf{.000} & \textbf{.000} & \textbf{.000} & .001 & \textbf{.000} & \textbf{.000} & .000 & .013 \\
  VQGAN & .017 & .003 & \textbf{.000} & .048 & \textbf{.000} & \textbf{.000} & \textbf{.000} & .000 & .001 & .000 & \textbf{.000} & \textbf{.000} & .006 \\
  SimVQ & .060 & .001 & \textbf{.000} & .032 & \textbf{.000} & \textbf{.000} & \textbf{.000} & \textbf{.000} & .005 & .000 & \textbf{.000} & .000 & .008 \\
  SimQINCo & .045 & .001 & \textbf{.000} & .003 & \textbf{.000} & \textbf{.000} & \textbf{.000} & .000 & .000 & .000 & \textbf{.000} & .001 & .004 \\
  RQVAE & .031 & .007 & \textbf{.000} & .124 & \textbf{.000} & \textbf{.000} & \textbf{.000} & \textbf{.000} & .013 & .000 & \textbf{.000} & .000 & .015 \\
  DiVeQ & .047 & .005 & \textbf{.000} & .037 & \textbf{.000} & \textbf{.000} & \textbf{.000} & \textbf{.000} & .011 & .003 & .000 & .004 & .009 \\
  SF-DiVeQ & .020 & \textbf{.000} & \textbf{.000} & \textbf{.000} & \textbf{.000} & \textbf{.000} & \textbf{.000} & \textbf{.000} & .002 & \textbf{.000} & \textbf{.000} & .001 & .002 \\
  BSQ & .008 & \textbf{.000} & \textbf{.000} & .015 & \textbf{.000} & \textbf{.000} & \textbf{.000} & \textbf{.000} & \textbf{.000} & \textbf{.000} & \textbf{.000} & \textbf{.000} & .002 \\
  LFQ & \textbf{.003} & \textbf{.000} & \textbf{.000} & \textbf{.000} & \textbf{.000} & \textbf{.000} & \textbf{.000} & \textbf{.000} & \textbf{.000} & \textbf{.000} & \textbf{.000} & \textbf{.000} & .000 \\
  \midrule
  \multicolumn{14}{l}{\textit{$f = 8$}} \\
  AEKL & .054 & \textbf{.000} & \textbf{.000} & \textbf{.000} & \textbf{.000} & \textbf{.000} & \textbf{.000} & \textbf{.000} & \textbf{.000} & \textbf{.000} & \textbf{.000} & \textbf{.000} & .005 \\
  SoftVQ & .010 & \textbf{.000} & \textbf{.000} & \textbf{.000} & \textbf{.000} & \textbf{.000} & \textbf{.000} & \textbf{.000} & .000 & \textbf{.000} & \textbf{.000} & \textbf{.000} & .001 \\
  VQGAN & .064 & .002 & \textbf{.000} & \textbf{.000} & \textbf{.000} & \textbf{.000} & \textbf{.000} & .000 & .007 & .001 & \textbf{.000} & .001 & .006 \\
  SimVQ & .036 & .000 & \textbf{.000} & \textbf{.000} & \textbf{.000} & \textbf{.000} & \textbf{.000} & \textbf{.000} & .003 & .000 & .000 & .001 & .003 \\
  SimQINCo & .082 & \textbf{.000} & \textbf{.000} & \textbf{.000} & \textbf{.000} & \textbf{.000} & \textbf{.000} & \textbf{.000} & .001 & .001 & .000 & .001 & .007 \\
  RQVAE & \textbf{.000} & \textbf{.000} & \textbf{.000} & \textbf{.000} & \textbf{.000} & \textbf{.000} & \textbf{.000} & \textbf{.000} & \textbf{.000} & \textbf{.000} & \textbf{.000} & \textbf{.000} & .000 \\
  DiVeQ & .144 & .001 & \textbf{.000} & \textbf{.000} & \textbf{.000} & \textbf{.000} & \textbf{.000} & \textbf{.000} & .003 & .003 & .000 & .004 & .013 \\
  SF-DiVeQ & .081 & \textbf{.000} & \textbf{.000} & .004 & \textbf{.000} & \textbf{.000} & \textbf{.000} & \textbf{.000} & .006 & .000 & .000 & .003 & .008 \\
  BSQ & .039 & \textbf{.000} & \textbf{.000} & \textbf{.000} & \textbf{.000} & \textbf{.000} & \textbf{.000} & \textbf{.000} & \textbf{.000} & \textbf{.000} & \textbf{.000} & \textbf{.000} & .003 \\
  LFQ & \textbf{.000} & \textbf{.000} & \textbf{.000} & \textbf{.000} & \textbf{.000} & \textbf{.000} & \textbf{.000} & \textbf{.000} & \textbf{.000} & \textbf{.000} & \textbf{.000} & \textbf{.000} & .000 \\
  \midrule
  \multicolumn{14}{l}{\textit{$f = 16$}} \\
  AEKL & .112 & \textbf{.000} & \textbf{.000} & \textbf{.000} & \textbf{.000} & \textbf{.000} & \textbf{.000} & \textbf{.000} & \textbf{.000} & \textbf{.000} & \textbf{.000} & \textbf{.000} & .009 \\
  SoftVQ & \textbf{.000} & \textbf{.000} & \textbf{.000} & \textbf{.000} & \textbf{.000} & \textbf{.000} & \textbf{.000} & \textbf{.000} & \textbf{.000} & \textbf{.000} & \textbf{.000} & \textbf{.000} & .000 \\
  VQGAN & .011 & \textbf{.000} & \textbf{.000} & \textbf{.000} & \textbf{.000} & \textbf{.000} & \textbf{.000} & \textbf{.000} & \textbf{.000} & \textbf{.000} & \textbf{.000} & \textbf{.000} & .001 \\
  SimVQ & .038 & \textbf{.000} & \textbf{.000} & \textbf{.000} & \textbf{.000} & \textbf{.000} & \textbf{.000} & \textbf{.000} & \textbf{.000} & \textbf{.000} & \textbf{.000} & \textbf{.000} & .003 \\
  SimQINCo & .037 & .000 & \textbf{.000} & \textbf{.000} & \textbf{.000} & \textbf{.000} & \textbf{.000} & \textbf{.000} & .003 & \textbf{.000} & \textbf{.000} & .000 & .003 \\
  RQVAE & .029 & \textbf{.000} & \textbf{.000} & .002 & \textbf{.000} & \textbf{.000} & \textbf{.000} & \textbf{.000} & .009 & .002 & .001 & .001 & .004 \\
  DiVeQ & .056 & .000 & \textbf{.000} & \textbf{.000} & \textbf{.000} & \textbf{.000} & \textbf{.000} & \textbf{.000} & .002 & \textbf{.000} & \textbf{.000} & .000 & .005 \\
  SF-DiVeQ & .023 & \textbf{.000} & \textbf{.000} & \textbf{.000} & \textbf{.000} & \textbf{.000} & \textbf{.000} & \textbf{.000} & \textbf{.000} & \textbf{.000} & \textbf{.000} & \textbf{.000} & .002 \\
  BSQ & .112 & .000 & \textbf{.000} & \textbf{.000} & \textbf{.000} & \textbf{.000} & \textbf{.000} & \textbf{.000} & .000 & \textbf{.000} & \textbf{.000} & \textbf{.000} & .009 \\
  LFQ & .000 & \textbf{.000} & \textbf{.000} & \textbf{.000} & \textbf{.000} & \textbf{.000} & \textbf{.000} & \textbf{.000} & \textbf{.000} & \textbf{.000} & \textbf{.000} & \textbf{.000} & .000 \\
  \bottomrule
  \end{tabular}}
\end{table}

\begin{table}[h]
  \centering
  \caption{Per-dataset memorization ratio ($\uparrow$), SSCD features.}
  \label{tab:mem-mem-ratio-sscd}
  \setlength{\tabcolsep}{3pt}
  \adjustbox{max width=0.48\textwidth,max totalheight=0.43\textheight,center}{%
  \begin{tabular}{lcccccccccccc|c}
  \toprule
  Method & Path & Chest & Derma & OCT & Pneu & Reti & Breast & Blood & Tiss & OrgA & OrgC & OrgS & mean \\
  \midrule
  \multicolumn{14}{l}{\textit{$f = 4$}} \\
  AEKL & .898 & .911 & .861 & .911 & .904 & .857 & .900 & .896 & .922 & .930 & .913 & .916 & .902 \\
  SoftVQ & .900 & .907 & .891 & .883 & .904 & .861 & .898 & .885 & .918 & .918 & .904 & .907 & .898 \\
  VQGAN & .898 & .896 & .867 & .865 & .885 & .846 & .887 & .884 & .914 & .917 & .905 & .906 & .889 \\
  SimVQ & .900 & .904 & .878 & .881 & .897 & .852 & .883 & .890 & .910 & .914 & .896 & .901 & .892 \\
  SimQINCo & .900 & .900 & .871 & .884 & .885 & .840 & .880 & .883 & .917 & .915 & .896 & .899 & .889 \\
  RQVAE & .900 & .896 & .879 & .859 & .887 & .853 & .872 & .889 & .907 & .914 & .895 & .896 & .887 \\
  DiVeQ & .895 & .890 & .874 & .872 & .886 & .838 & .876 & .883 & .906 & .906 & .891 & .893 & .884 \\
  SF-DiVeQ & .899 & .907 & .865 & .888 & .899 & .844 & .887 & .884 & .911 & .920 & .903 & .905 & .893 \\
  BSQ & .906 & .909 & .883 & .892 & .913 & .862 & .912 & .902 & .920 & .933 & .910 & .915 & .905 \\
  LFQ & \textbf{.913} & \textbf{.928} & \textbf{.902} & \textbf{.922} & \textbf{.928} & \textbf{.902} & \textbf{.930} & \textbf{.921} & \textbf{.945} & \textbf{.950} & \textbf{.936} & \textbf{.936} & .926 \\
  \midrule
  \multicolumn{14}{l}{\textit{$f = 8$}} \\
  AEKL & .896 & .905 & .866 & .907 & .896 & .843 & .899 & .888 & .913 & .921 & .902 & .906 & .895 \\
  SoftVQ & .912 & .908 & .875 & .904 & .901 & .861 & .913 & .900 & .914 & .930 & .916 & .918 & .904 \\
  VQGAN & .900 & .896 & .867 & .892 & .889 & .843 & .902 & .887 & .906 & .906 & .892 & .889 & .889 \\
  SimVQ & .903 & .903 & .874 & .888 & .900 & .843 & .889 & .890 & .907 & .910 & .886 & .887 & .890 \\
  SimQINCo & .900 & .904 & .865 & .887 & .891 & .835 & .893 & .886 & .913 & .905 & .885 & .887 & .888 \\
  RQVAE & \textbf{.942} & \textbf{.960} & \textbf{.950} & \textbf{.967} & .881 & \textbf{.958} & \textbf{.950} & \textbf{.963} & \textbf{.974} & \textbf{.963} & \textbf{.955} & .946 & .951 \\
  DiVeQ & .893 & .899 & .868 & .890 & .891 & .846 & .889 & .883 & .909 & .898 & .880 & .883 & .886 \\
  SF-DiVeQ & .896 & .909 & .875 & .884 & .902 & .842 & .897 & .886 & .904 & .908 & .893 & .895 & .891 \\
  BSQ & .903 & .912 & .870 & .908 & .900 & .858 & .909 & .902 & .930 & .936 & .922 & .922 & .906 \\
  LFQ & .931 & .950 & .917 & .940 & \textbf{.938} & .925 & .941 & .934 & .947 & .955 & .946 & \textbf{.947} & .939 \\
  \midrule
  \multicolumn{14}{l}{\textit{$f = 16$}} \\
  AEKL & .903 & .900 & .878 & .905 & .894 & .837 & .896 & .900 & .913 & .920 & .909 & .911 & .897 \\
  SoftVQ & \textbf{.962} & \textbf{.967} & \textbf{.951} & \textbf{.962} & \textbf{.942} & \textbf{.947} & .922 & \textbf{.965} & \textbf{.959} & \textbf{.961} & \textbf{.954} & \textbf{.953} & .954 \\
  VQGAN & .910 & .901 & .877 & .913 & .892 & .839 & .913 & .909 & .923 & .930 & .917 & .919 & .904 \\
  SimVQ & .906 & .902 & .869 & .905 & .888 & .843 & .914 & .907 & .920 & .935 & .920 & .924 & .903 \\
  SimQINCo & .907 & .900 & .879 & .903 & .889 & .846 & .908 & .899 & .909 & .927 & .911 & .914 & .899 \\
  RQVAE & .899 & .906 & .871 & .891 & .895 & .837 & .902 & .894 & .906 & .901 & .893 & .895 & .891 \\
  DiVeQ & .904 & .901 & .873 & .901 & .895 & .840 & .909 & .900 & .908 & .923 & .910 & .909 & .898 \\
  SF-DiVeQ & .904 & .902 & .868 & .899 & .891 & .836 & .901 & .899 & .912 & .926 & .911 & .914 & .897 \\
  BSQ & .899 & .909 & .880 & .909 & .901 & .855 & .909 & .905 & .916 & .932 & .921 & .922 & .905 \\
  LFQ & .919 & .910 & .871 & .917 & .900 & .873 & \textbf{.929} & .911 & .929 & .944 & .933 & .933 & .914 \\
  \bottomrule
  \end{tabular}}
\end{table}

\begin{table}[h]
  \centering
  \caption{Per-dataset copy rate ($\downarrow$), DINOv2 features.}
  \label{tab:mem-copy-rate-dinov2}
  \setlength{\tabcolsep}{3pt}
  \adjustbox{max width=0.48\textwidth,max totalheight=0.43\textheight,center}{%
  \begin{tabular}{lcccccccccccc|c}
  \toprule
  Method & Path & Chest & Derma & OCT & Pneu & Reti & Breast & Blood & Tiss & OrgA & OrgC & OrgS & mean \\
  \midrule
  \multicolumn{14}{l}{\textit{$f = 4$}} \\
  AEKL & .022 & \textbf{.000} & \textbf{.000} & .004 & \textbf{.000} & \textbf{.000} & \textbf{.000} & \textbf{.000} & .003 & \textbf{.000} & .002 & .004 & .003 \\
  SoftVQ & .069 & \textbf{.000} & \textbf{.000} & .030 & \textbf{.000} & \textbf{.000} & \textbf{.000} & \textbf{.000} & .022 & .000 & .002 & .003 & .010 \\
  VQGAN & .037 & .003 & \textbf{.000} & .159 & \textbf{.000} & \textbf{.000} & \textbf{.000} & \textbf{.000} & .010 & .001 & .004 & .005 & .018 \\
  SimVQ & .085 & .001 & \textbf{.000} & .213 & \textbf{.000} & \textbf{.000} & \textbf{.000} & \textbf{.000} & .035 & .001 & .007 & .013 & .030 \\
  SimQINCo & .085 & .006 & \textbf{.000} & .214 & \textbf{.000} & \textbf{.000} & \textbf{.000} & \textbf{.000} & .018 & .000 & .004 & .004 & .028 \\
  RQVAE & .059 & .006 & \textbf{.000} & .200 & \textbf{.000} & \textbf{.000} & \textbf{.000} & \textbf{.000} & .026 & .001 & .004 & .004 & .025 \\
  DiVeQ & .052 & .000 & \textbf{.000} & .103 & \textbf{.000} & \textbf{.000} & \textbf{.000} & \textbf{.000} & .031 & .003 & .014 & .015 & .018 \\
  SF-DiVeQ & .023 & .002 & \textbf{.000} & .156 & \textbf{.000} & \textbf{.000} & \textbf{.000} & \textbf{.000} & .016 & .005 & .014 & .011 & .019 \\
  BSQ & .012 & \textbf{.000} & \textbf{.000} & .088 & \textbf{.000} & \textbf{.000} & \textbf{.000} & \textbf{.000} & .008 & \textbf{.000} & .001 & .000 & .009 \\
  LFQ & \textbf{.000} & \textbf{.000} & \textbf{.000} & \textbf{.000} & \textbf{.000} & \textbf{.000} & \textbf{.000} & \textbf{.000} & \textbf{.000} & \textbf{.000} & \textbf{.000} & \textbf{.000} & .000 \\
  \midrule
  \multicolumn{14}{l}{\textit{$f = 8$}} \\
  AEKL & .010 & \textbf{.000} & \textbf{.000} & .004 & \textbf{.000} & \textbf{.000} & \textbf{.000} & \textbf{.000} & .009 & \textbf{.000} & .007 & .006 & .003 \\
  SoftVQ & \textbf{.000} & \textbf{.000} & \textbf{.000} & .001 & \textbf{.000} & \textbf{.000} & \textbf{.000} & \textbf{.000} & .009 & \textbf{.000} & .000 & .000 & .001 \\
  VQGAN & \textbf{.000} & \textbf{.000} & \textbf{.000} & .004 & \textbf{.000} & \textbf{.000} & \textbf{.000} & \textbf{.000} & .014 & .001 & .016 & .016 & .004 \\
  SimVQ & .012 & \textbf{.000} & \textbf{.000} & .049 & \textbf{.000} & \textbf{.000} & \textbf{.000} & \textbf{.000} & .018 & .000 & .025 & .022 & .010 \\
  SimQINCo & .018 & .000 & \textbf{.000} & .008 & \textbf{.000} & \textbf{.000} & \textbf{.000} & \textbf{.000} & .011 & .001 & .009 & .010 & .005 \\
  RQVAE & \textbf{.000} & \textbf{.000} & \textbf{.000} & \textbf{.000} & \textbf{.000} & \textbf{.000} & \textbf{.000} & \textbf{.000} & \textbf{.000} & \textbf{.000} & \textbf{.000} & \textbf{.000} & .000 \\
  DiVeQ & .037 & \textbf{.000} & \textbf{.000} & .011 & \textbf{.000} & \textbf{.000} & \textbf{.000} & \textbf{.000} & .011 & .001 & .010 & .014 & .007 \\
  SF-DiVeQ & .029 & \textbf{.000} & \textbf{.000} & .047 & \textbf{.000} & \textbf{.000} & \textbf{.000} & \textbf{.000} & .022 & .000 & .005 & .006 & .009 \\
  BSQ & .003 & \textbf{.000} & \textbf{.000} & .002 & \textbf{.000} & \textbf{.000} & \textbf{.000} & \textbf{.000} & .004 & \textbf{.000} & .001 & .003 & .001 \\
  LFQ & \textbf{.000} & \textbf{.000} & \textbf{.000} & \textbf{.000} & \textbf{.000} & \textbf{.000} & \textbf{.000} & \textbf{.000} & \textbf{.000} & \textbf{.000} & \textbf{.000} & \textbf{.000} & .000 \\
  \midrule
  \multicolumn{14}{l}{\textit{$f = 16$}} \\
  AEKL & .000 & \textbf{.000} & \textbf{.000} & .001 & \textbf{.000} & \textbf{.000} & \textbf{.000} & \textbf{.000} & .011 & \textbf{.000} & .003 & .004 & .002 \\
  SoftVQ & \textbf{.000} & \textbf{.000} & \textbf{.000} & \textbf{.000} & \textbf{.000} & \textbf{.000} & \textbf{.000} & \textbf{.000} & \textbf{.000} & \textbf{.000} & \textbf{.000} & \textbf{.000} & .000 \\
  VQGAN & \textbf{.000} & \textbf{.000} & \textbf{.000} & \textbf{.000} & \textbf{.000} & \textbf{.000} & \textbf{.000} & \textbf{.000} & .000 & \textbf{.000} & .001 & .001 & .000 \\
  SimVQ & \textbf{.000} & \textbf{.000} & \textbf{.000} & .001 & \textbf{.000} & \textbf{.000} & \textbf{.000} & \textbf{.000} & .002 & \textbf{.000} & .003 & .006 & .001 \\
  SimQINCo & \textbf{.000} & \textbf{.000} & \textbf{.000} & .004 & \textbf{.000} & \textbf{.000} & \textbf{.000} & \textbf{.000} & .023 & \textbf{.000} & .002 & .002 & .003 \\
  RQVAE & \textbf{.000} & \textbf{.000} & \textbf{.000} & .006 & \textbf{.000} & \textbf{.000} & \textbf{.000} & \textbf{.000} & .040 & .000 & .006 & .003 & .005 \\
  DiVeQ & \textbf{.000} & \textbf{.000} & \textbf{.000} & \textbf{.000} & \textbf{.000} & \textbf{.000} & \textbf{.000} & \textbf{.000} & .031 & .000 & .005 & .004 & .003 \\
  SF-DiVeQ & \textbf{.000} & \textbf{.000} & \textbf{.000} & .009 & \textbf{.000} & \textbf{.000} & \textbf{.000} & \textbf{.000} & .010 & \textbf{.000} & .003 & .006 & .002 \\
  BSQ & .000 & \textbf{.000} & \textbf{.000} & .001 & \textbf{.000} & \textbf{.000} & \textbf{.000} & \textbf{.000} & .022 & \textbf{.000} & .003 & .003 & .002 \\
  LFQ & \textbf{.000} & \textbf{.000} & \textbf{.000} & \textbf{.000} & \textbf{.000} & \textbf{.000} & \textbf{.000} & \textbf{.000} & \textbf{.000} & \textbf{.000} & .000 & .000 & .000 \\
  \bottomrule
  \end{tabular}}
\end{table}

\begin{table*}[h]
  \centering
  \caption{Per-dataset memorization ratio ($\uparrow$), DINOv2 features.}
  \label{tab:mem-mem-ratio-dinov2}
  \setlength{\tabcolsep}{3pt}
  \adjustbox{max width=0.48\textwidth,max totalheight=0.43\textheight,center}{%
  \begin{tabular}{lcccccccccccc|c}
  \toprule
  Method & Path & Chest & Derma & OCT & Pneu & Reti & Breast & Blood & Tiss & OrgA & OrgC & OrgS & mean \\
  \midrule
  \multicolumn{14}{l}{\textit{$f = 4$}} \\
  AEKL & .843 & .882 & .846 & .854 & .875 & .865 & .771 & \textbf{.884} & .826 & .822 & .812 & .814 & .841 \\
  SoftVQ & .832 & .873 & \textbf{.852} & .846 & \textbf{.883} & .854 & .786 & .873 & .790 & .814 & .809 & .813 & .835 \\
  VQGAN & .797 & .858 & .823 & .823 & .850 & .804 & .753 & .864 & .800 & .821 & .802 & .807 & .817 \\
  SimVQ & .796 & .859 & .816 & .828 & .847 & .770 & .758 & .871 & .789 & .819 & .802 & .802 & .813 \\
  SimQINCo & .798 & .854 & .823 & .822 & .832 & .816 & .761 & .855 & .793 & .822 & .797 & .806 & .815 \\
  RQVAE & .791 & .854 & .804 & .820 & .848 & .800 & .745 & .851 & .793 & .815 & .793 & .797 & .809 \\
  DiVeQ & .806 & .858 & .821 & .835 & .847 & .794 & .764 & .850 & .792 & .820 & .804 & .808 & .817 \\
  SF-DiVeQ & .808 & .857 & .825 & .828 & .844 & .785 & .767 & .858 & .796 & .822 & .797 & .811 & .816 \\
  BSQ & .814 & .871 & .847 & .848 & .874 & \textbf{.889} & .770 & .860 & .783 & .813 & .782 & .787 & .828 \\
  LFQ & \textbf{.906} & \textbf{.901} & .820 & \textbf{.863} & .874 & .887 & \textbf{.796} & .876 & \textbf{.842} & \textbf{.841} & \textbf{.842} & \textbf{.836} & .857 \\
  \midrule
  \multicolumn{14}{l}{\textit{$f = 8$}} \\
  AEKL & .851 & .875 & .847 & .853 & .867 & .853 & .776 & .879 & .809 & .819 & .804 & .810 & .837 \\
  SoftVQ & .875 & .882 & .845 & .861 & .888 & .869 & .800 & .877 & .793 & .823 & .820 & .818 & .846 \\
  VQGAN & .872 & .869 & .838 & .851 & .851 & .844 & .768 & .874 & .794 & .807 & .797 & .790 & .830 \\
  SimVQ & .824 & .870 & .821 & .842 & .864 & .799 & .754 & .865 & .785 & .809 & .786 & .784 & .817 \\
  SimQINCo & .839 & .864 & .844 & .850 & .850 & .797 & .778 & .872 & .797 & .812 & .796 & .799 & .825 \\
  RQVAE & .875 & \textbf{.953} & \textbf{.925} & \textbf{.954} & \textbf{.989} & \textbf{.941} & \textbf{.922} & .887 & \textbf{.911} & .855 & \textbf{.878} & \textbf{.897} & .916 \\
  DiVeQ & .829 & .865 & .823 & .845 & .850 & .820 & .773 & .861 & .798 & .798 & .792 & .795 & .821 \\
  SF-DiVeQ & .819 & .864 & .815 & .836 & .857 & .825 & .758 & .868 & .789 & .814 & .804 & .809 & .822 \\
  BSQ & .873 & .877 & .854 & .853 & .868 & .901 & .779 & .883 & .802 & .848 & .836 & .835 & .851 \\
  LFQ & \textbf{.902} & .932 & .850 & .885 & .889 & .920 & .828 & \textbf{.895} & .851 & \textbf{.873} & .866 & .836 & .877 \\
  \midrule
  \multicolumn{14}{l}{\textit{$f = 16$}} \\
  AEKL & .867 & .872 & .848 & .854 & .873 & .881 & .758 & .884 & .811 & .817 & .807 & .805 & .840 \\
  SoftVQ & \textbf{.936} & .807 & .864 & \textbf{.911} & .889 & .812 & \textbf{.918} & \textbf{.961} & \textbf{.935} & \textbf{.916} & \textbf{.890} & \textbf{.875} & .893 \\
  VQGAN & .874 & .895 & \textbf{.867} & .846 & .877 & .877 & .838 & .892 & .831 & .820 & .817 & .813 & .854 \\
  SimVQ & .874 & .882 & .856 & .861 & .880 & \textbf{.889} & .827 & .884 & .829 & .814 & .802 & .812 & .851 \\
  SimQINCo & .878 & .873 & .853 & .856 & .870 & .871 & .813 & .883 & .797 & .806 & .807 & .806 & .843 \\
  RQVAE & .851 & .868 & .834 & .852 & .854 & .828 & .768 & .869 & .784 & .801 & .798 & .803 & .826 \\
  DiVeQ & .882 & .873 & .847 & .859 & .863 & .874 & .796 & .882 & .795 & .811 & .807 & .805 & .841 \\
  SF-DiVeQ & .881 & .874 & .846 & .846 & .864 & .864 & .798 & .885 & .805 & .812 & .819 & .807 & .842 \\
  BSQ & .881 & .883 & .852 & .866 & .871 & .860 & .787 & .885 & .793 & .827 & .826 & .815 & .845 \\
  LFQ & .867 & \textbf{.911} & .854 & .870 & \textbf{.892} & .859 & .877 & .897 & .854 & .842 & .832 & .827 & .865 \\
  \bottomrule
  \end{tabular}}
\end{table*}

\FloatBarrier

\end{document}